%% file: main.tex
\documentclass[12pt]{article}

\usepackage{comment}

\usepackage{amsmath, amssymb}
\usepackage{tikz}
\usepackage{amsmath} 
\usepackage[dvipsnames]{xcolor} 
\usepackage{multirow}
\usepackage{soul}

\usetikzlibrary{shapes.geometric, shadows, matrix}

\definecolor{realColor}{RGB}{31, 119, 180}   
\definecolor{synColor}{RGB}{214, 39, 40}     
\usepackage{hyperref}
\usepackage{tikz}
\usetikzlibrary{arrows.meta, positioning, calc, backgrounds, shadows, decorations.pathreplacing}
\definecolor{tensorColor}{RGB}{255, 255, 255}  
\definecolor{lineColor}{RGB}{50, 50, 50}       
\definecolor{realColor}{RGB}{0, 100, 200}    
\definecolor{synColor}{RGB}{220, 50, 50}     
\definecolor{overlapColor}{RGB}{100, 0, 100} 
\definecolor{realColor}{RGB}{0, 100, 200}    
\definecolor{synColor}{RGB}{220, 50, 50}     
\definecolor{transportColor}{RGB}{230, 140, 0} 

\definecolor{encoderColor}{RGB}{230, 240, 250} 
\definecolor{decoderColor}{RGB}{250, 230, 230} 
\definecolor{tensorColor}{RGB}{255, 255, 255}  
\definecolor{lineColor}{RGB}{50, 50, 50}       
\definecolor{highlight}{RGB}{0, 100, 180}      

\usepackage{float}
\usepackage{rotating}
\usepackage{rotfloat}
\usepackage{graphicx}
\usepackage{caption}
\usepackage{todonotes}
\usepackage{marginnote}
\newif\ifreviewnotes\reviewnotestrue

\usepackage{xcolor}
\usepackage{bclogo}

\newif\ifdraft\drafttrue

\usepackage[T1]{fontenc}
\usepackage{sourceserifpro}
\usepackage{sourcecodepro}

\usepackage{MnSymbol}

\usepackage[toc, page]{appendix}
\AfterEndEnvironment{thebibliography}{
\setcounter{theorem}{0}
\setcounter{proposition}{0}
\setcounter{lemma}{0}
\setcounter{corollary}{0}
\setcounter{definition}{0}
\setcounter{assumption}{0}
\setcounter{remark}{0}
\setcounter{table}{0}
\setcounter{figure}{0}
\setcounter{equation}{0}

\renewcommand{\thetable}{A\arabic{table}}
\renewcommand{\thefigure}{A\arabic{figure}}
\renewcommand{\theequation}{A\arabic{equation}}
\titleformat{\section}{\centering\large\bfseries}{Appendix~\thesection.}{1em}{}}

\usepackage{xparse}
\NewDocumentCommand{\note}{o g}{%
\parbox{\textwidth}{\footnotesize\vspace*{10pt}%
\IfValueT{#1}{\textit{#1}:\quad}%
#2}}

\usepackage{booktabs}

\usepackage[
  backend=biber,
  style=numeric,
  maxnames=3,
  minnames=1
]{biblatex}
\AtBeginBibliography{\footnotesize}

\usepackage{authblk}

\usepackage{makecell}
\usepackage{listings}

\usepackage{algorithm}
\usepackage{algpseudocode}

\algrenewcommand\algorithmicrequire{\textbf{Input:}}
\algrenewcommand\algorithmicensure{\textbf{Output:}}
\algblockdefx[DoWhile]{Do}{While}{\textbf{do}}{\textbf{while}}

\usepackage{enumitem}
\usepackage{bbm}

\begin{document}

\lstset{
  basicstyle=\ttfamily\small,
  columns=fullflexible,
  keepspaces=true,
  moredelim=**[is][\color{cyan}]{@}{@},
  moredelim=**[is][\color{blue}]{!}{!},
  moredelim=**[is][\bfseries]{~}{~}
}

\title{
Glioma IDH Classification via CVAE: Synthetic Augmentation, Generative Classification, and MIL for Raman Biopsies
}
\author[1]{Andrei Iușan}
\author[1]{Daria Voiculescu}
\author[1,2]{Ion Petre}
\author[3,4]{Andrei P\u{a}un}
\author[1,5]{Bogdan Oancea}
\author[1,5]{Mihaela P\u{a}un}

\affil[1]{National Institute of Research and Development for Biological Sciences, Romania}
\affil[2]{Department of Mathematics and Statistics, University of Turku, Finland}
\affil[3]{Research Institute for Artificial Intelligence ``Mihai Drăgănescu'', Romanian Academy}
\affil[4]{Faculty of Mathematics and Computer Science, University of Bucharest, Romania}
\affil[5]{Faculty of Business and Administration, University of Bucharest, Romania}

\date{}

\maketitle

\begin{abstract}
Biomedical Raman datasets used in machine learning for glioma diagnostics are characterized by small cohort size and absence of diagnostic tissue annotations. We address both problems using a $\beta$-regularized conditional variational autoencoder and a dataset of 59 glioma biopsies from 46 patients for binary IDH-status classification. The $\beta$-CVAE was used as a: (1) Raman spectra generator; (2) generative classifier of Raman spectra; (3) Multiple Instance Learning (MIL) classifier of biopsy slides. Four classifiers were trained on real spectra, synthetic spectra, or a combined set and evaluated on held-out real slides. Random Forest and XGBoost achieved accuracy $0.83$ with synthetic training and $0.81$ with real training. Augmentation increased RF+SVM accuracy from $0.76$ to $0.83$; changes for the other classifiers were smaller. We show that the $\beta$-CVAE model can successfully be used in training IDH-status classifiers and as a generative classifier, with comparable results to other models; however, the results of using it in MIL are inconclusive and further study is needed.
\end{abstract}

\section{Introduction}
Glioma diagnosis relies on both histology and molecular markers~\cite{louis2021who}. In particular, the mutation status of the isocitrate dehydrogenase genes (IDH1 and IDH2) is a strong predictor of disease prognosis, as it differentiates between IDH-Mutant and IDH-Wildtype tumors~\cite{Ceccarelli2016, cancer2015comprehensive,eckel2015glioma}. Accurate molecular classification is therefore important for diagnosis and treatment planning~\cite{gbm_review_2025}. 

Raman spectroscopy has emerged as a potential tool for tumor grading and molecular characterization, because subtle biochemical differences between tissue types can be reflected in spectral signatures~\cite{krafft2009raman,bocklitz2016raman}. Machine learning can exploit Raman measurements to distinguish glioma molecular groups~\cite{sciortino2021raman, apollonov2024raman}. However, Raman studies often involve only tens of patients, and spectra originating from the same biopsy are strongly correlated~\cite{stupak2025raman}. The number of spectra therefore overstates the number of independent observations available for assessing generalization~\cite{guo2021chemometric}. High spectral dimensionality and acquisition-related variability further complicate learning in small cohorts~\cite{berisha2021overfitting,subramanian2013overfitting}.

Deep generative models for synthetic data augmentation can be used to address these limitations~\cite{jordon2022synthetic,waseem2025review}. Encouraging results have been reported in Raman cancerous tissue classification~\cite{wu2021raman,zhao2024raman}. Nonetheless, the practical utility of generative augmentation in low-data settings characteristic of biomedical Raman spectroscopy remains insufficiently understood. In particular, it is unclear to what extent generative models trained on relatively small cohorts can learn sufficiently meaningful spectral structure to improve classification performance.

In this work, we investigated three roles of a $\beta$-regularized conditional variational autoencoder ($\beta$-CVAE) for IDH-status classification on a Raman dataset of 59 glioma biopsy slides. We first studied whether formulating the task as Multiple Instance Learning (MIL), with each slide as a bag of spectra, improves classification over the usual assumption that every spectrum inherits the label of its slide and designed MIL formulations using the CVAE. Moreover, we analyzed classification performance on synthetic and augmented data, under the \textit{Train on Synthetic, Test on Real} (TS/TR) and\textit{Train on Synthetic+Real, Test on Real} (TSR/TR) frameworks. We tested four standard classifiers (a Random Forest feature selector followed by an SVM, a Random Forest, XGBoost and a 1D-CNN) and the $\beta$-CVAE itself as a generative classifier. All evaluations are done at the patient level, with patient-grouped cross-validation that prevents data leakage between training and test patients.

\section{Background}
\label{sec:background}
Raman spectroscopy yields multiple spectra per slide, yet the diagnostic labels are assigned to the slide, without pixel-level annotations~\cite{stupak2025raman,apollonov2024raman}. However, a portion of the spectra may originate from non-diagnostic tissue regions of the biopsy. The Multiple Instance Learning (MIL) paradigm can be used to formalize this problem~\cite{dietterich1997solving}; it has become a standard in computational pathology~\cite{waqas2024exploring,gadermayr2024multiple} and has been successfully applied to Raman-based breast cancer diagnosis~\cite{yang2023histological}. 

Classical augmentation methods perturb or interpolate available observations, for example through noise injection or SMOTE~\cite{zhou2022deeply,chawla2002smote,song2022rapid}.
In contrast, deep generative models attempt to learn a distribution from which additional samples can be drawn~\cite{jordon2022synthetic}. Variational autoencoders (VAEs) encode spectra into a latent representation and reconstruct them through a decoder~\cite{kingma2013auto, flanagan2025exploring}. A conditional VAE (CVAE) further includes class information to generate spectra for a requested label~\cite{sohn2015learning}. Moreover, a $\beta$-VAE architecture is characterized by a regularization parameter $\beta$ which quantifies how closely the latent space should follow the distribution, at the cost of reconstruction quality~\cite{higgins2017betavae}.

The utility of deep generative models can be evaluated using the \emph{Train on Synthetic, Test on Real} (TS/TR) framework. A performance close to the corresponding \emph{Train on Real, Test on Real} (TR/TR) baseline provides strong evidence that the generator has captured discriminative structure relevant to the classification task~\cite{esteban2017real}. A complementary protocol, \emph{Train on Synthetic+Real, Test on Real} (TSR/TR) has also been proposed for measuring whether the generator adds information beyond the one already provided by the real training set~\cite{kiyasseh2020pleth}. 

We implemented a $\beta$-regularized conditional variational autoencoder ($\beta$-CVAE) and used it in three settings: as a data source for training classifiers under the TS/TR and TSR/TR frameworks, as a generative classifier, and finally as a MIL design. To the best of our knowledge, this combination of use cases for the CVAE model has never been explored in other Raman studies.


\section{Methodology}
\label{sec:methodology}

Our methodology combines preprocessing and partitioning Raman spectra, and implementing a $\beta$-CVAE for both MIL design and synthetic spectra generation. The $\beta$-CVAE architecture was used in Attention MIL and CVAE MIL. For testing predictive utility of synthetic data, four classifiers were evaluated under TS/TR, TSR/TR settings; the $\beta$-CVAE was also used as a generative classifier, through the Classification by Reconstruction rule. 

\subsection{Data collection}
\label{sec:data_col}
We analyzed the dataset from Lita et al.~\cite{apollonov2024raman}, which contains Raman spectroscopic recordings from 59 brain tissue biopsies across 46 patients. Each biopsy sample is structured as an $(H,W,N_\lambda)$ tensor, describing the spatial grid of size $(H,W)$ (the slice dimensions) and the size of the Raman spectrum vector ($N_\lambda = 1738$), stored at each point on the grid (also called \textit{pixel}). In total, the dataset consists of 300,506 Raman spectra. The grid size $(H,W)$ ranges from $(46,46)$ to $(245,61)$. 

Using the methodology of Ceccarelli et al., the biopsies were labeled based on their methylation profiles, denoted \texttt{LGm1} to \texttt{LGm6}~\cite{Ceccarelli2016}. According to the WHO classification framework for central nervous system tumors~\cite{louis2021who}, the mutation status of isocitrate dehydrogenase (IDH) represents an essential biomarker for glioma diagnosis, categorizing gliomas as either IDH-Mutant or IDH-Wildtype. Binary classification of IDH status (IDH-Mutant or IDH-Wildtype) can be performed by grouping the methylation profiles \texttt{LGm1}--\texttt{LGm3} as IDH-Mutant, and \texttt{LGm4}--\texttt{LGm6} as IDH-Wildtype (Table \ref{tab:dataset}).

While the binary classification setting presents a roughly symmetric label distribution, the dataset is imbalanced for the six-profile classification task. 

\begin{table}[H]
\small
\centering
\begin{tabular}{llccccc}
\toprule
\textbf{Class} & \textbf{IDH Group} & \textbf{Slides} & \textbf{Patients} & \textbf{Spectra} & \makecell{\textbf{Slides} \\ \textbf{Total}} & \makecell{\textbf{Spectra} \\ \textbf{Total}} \\
\midrule
\texttt{LGm1} &   & 5 & 5 & 37,319 \\
\texttt{LGm2} & IDH-Mutant   & 12 & 12 & 77,642 & 26 & 167,414\\
\texttt{LGm3} &  & 9 & 4 & 52,453 \\
\midrule
\texttt{LGm4} &  & 13 & 10 & 50,660  \\
\texttt{LGm5} & IDH-Wildtype & 15 & 11 & 62,256 &  33 & 133,092\\
\texttt{LGm6} &  & 5 & 4 & 20,176  \\
\bottomrule
\end{tabular}
\caption{Dataset composition.}
\label{tab:dataset}
\end{table}

To prevent spectra from the same biopsy from appearing in both the training and test sets, data partitioning was performed with slides grouped by patient, following the stratified five-fold evaluation protocol of Lita et al.~\cite{apollonov2024raman}. All slides from a given patient were assigned exclusively to either the training or the test partition in each fold, preventing patient-level data leakage (Appendix \ref{app:cv_schema}). The classifiers were trained independently on the respective training partitions and evaluated on the corresponding test partitions, and performance metrics were aggregated across the five folds.

\subsection{Data preprocessing}
\label{sec:annotation}

\begin{figure}[H]
    \centering
    \includegraphics[width=\linewidth]{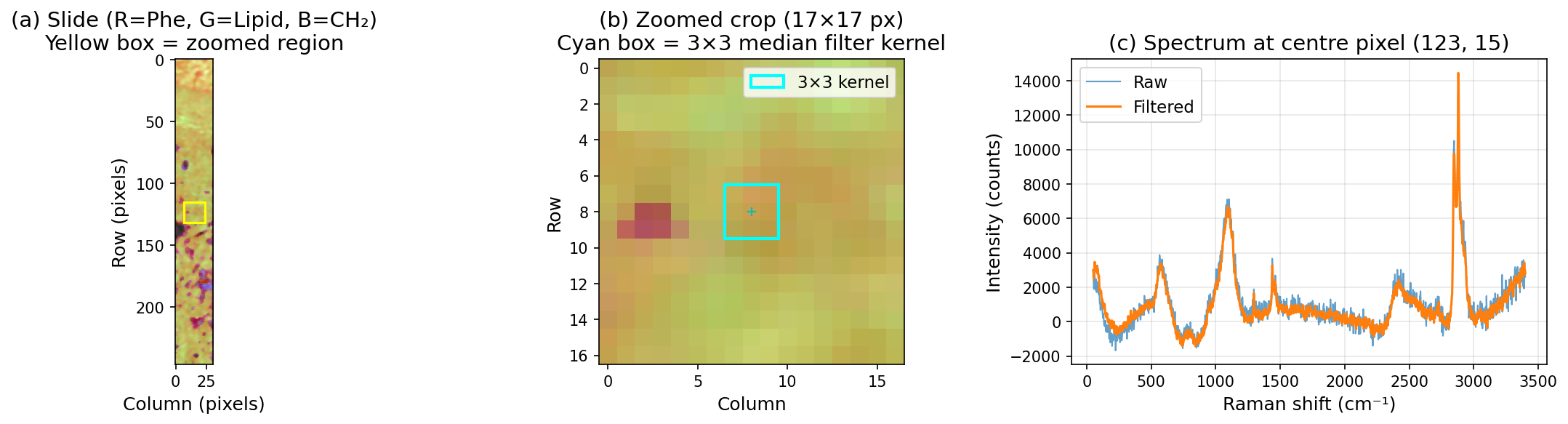}
    \caption{\footnotesize (a) False coloring of a slide, with Red = Phenylalanine marker (990--1020$\,\text{cm}^{-1}$), Green = Lipids (2800--3000$\,\text{cm}^{-1}$), Blue = $\mathrm{CH_2}$ deformations (1430--1460$\,\text{cm}^{-1}$). (b) Zoomed region for 3$\times$3 median kernel visualization. (c) Raw vs. Median filtered spectra.}
    \label{fig:preprocessing_median_filter_illustration}
\end{figure}

An important preprocessing step is removing low-signal spectra, such as those corresponding to non-biological background regions of the slide. Therefore, we developed a preprocessing pipeline that operated on each slide separately, consisting of the following steps:

\begin{enumerate}
    \item \emph{Median filtering.} We applied a 3$\times$3 median kernel on the spatial dimensions $(H,W)$ of each slide for denoising purposes, filtering out cosmic radiation and Gaussian noise (Figure \ref{fig:preprocessing_median_filter_illustration}).

    \item \emph{Slide-wise normalization.} Each data tensor was normalized independently to the $[0,1]$ range to ensure numerical stability.

    \item \emph{Signal filtering.} We removed spectra with low intensity (likely dominated by noise or coming from slide regions without tissue), by discarding those with maximum intensity below 0.60; this threshold was chosen after testing among 0.40, 0.60, and an \emph{adaptive} rule: top 90\%, floor 0.80 (Section \ref{subsec:max_int} in Appendix \ref{app:controlled_experiments}). 
    

    \item \emph{Baseline correction.} For the spectra selected as relevant, we performed two steps on each spectrum: (i) normalization to $[0, 1]$; (ii) baseline correction with the Improved Asymmetrically Reweighted Penalized Least Squares method (IARPLS)~\cite{Kizilov2025, ye2020baseline}, with $\lambda=10^4$ and difference order $p=2$. We selected this method by comparing it to the Adaptive Iteratively Reweighted Penalized Least Squares (airPLS)~\cite{zhang2010baseline} method and tuning their parameters to our data. The experiment suite for this selection can be found in Appendix \ref{app:baseline_correction}.
\end{enumerate}

The pipeline for threshold 0.60 yielded 288,678 retained spectra out of the original 300,506. Retention varied from 75\% to 100\%, reflecting differences in biopsy quality. 

We observed that the initial dataset did not have histopathological annotations for slide regions, and as such the retained spectra of a slide could include non-diagnostic tissue~\cite{apollonov2024raman}. To handle this, we formulated the problem with the Multiple Instance Learning (MIL) paradigm~\cite{dietterich1997solving, waqas2024exploring}. 

\begin{figure}[H]

    \centering
    \includegraphics[width=\linewidth]{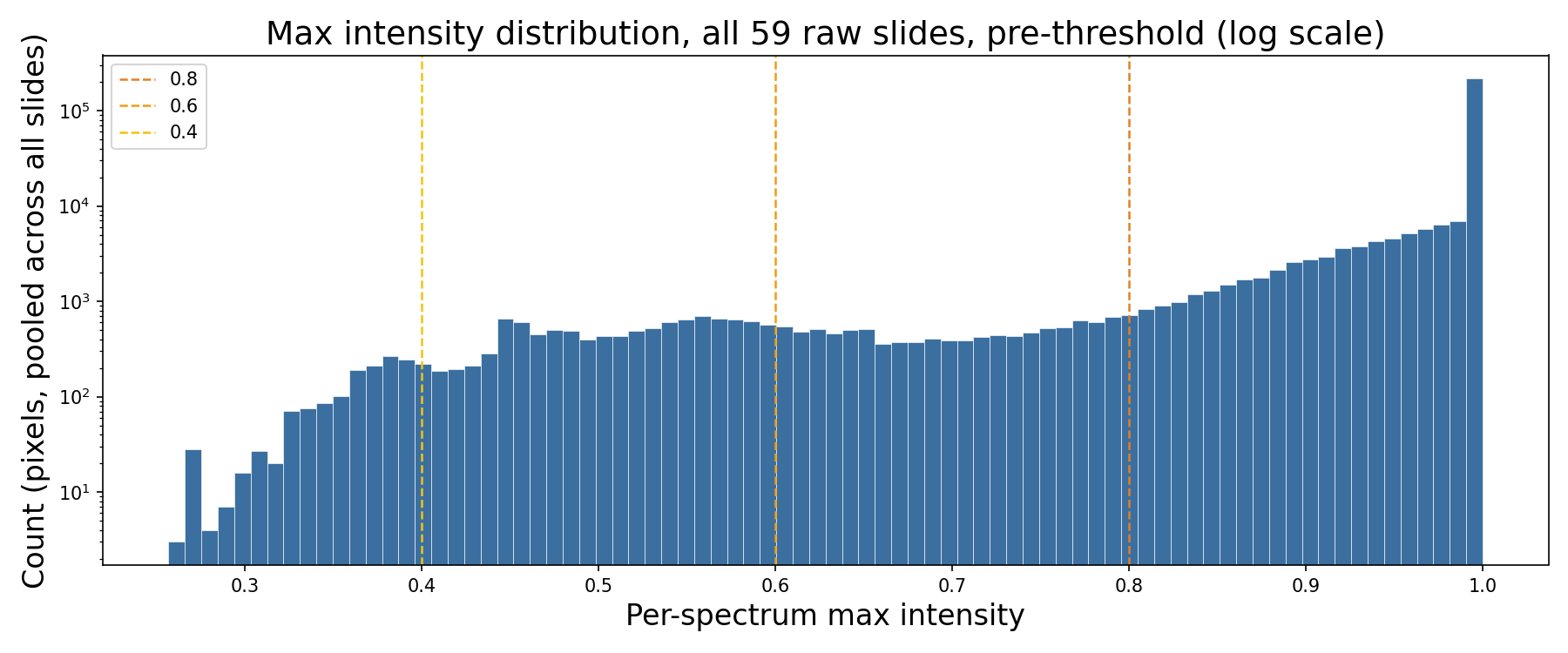} 
   
    \caption{\footnotesize Histogram of maximum intensity of spectra for the entire dataset. Vertical lines indicate threshold values tested.}
    \label{fig:preprocessing_cluster_spectra}
\end{figure}

\subsection{Multiple Instance Learning}\label{subsec:MIL}

To account for the possibility that a slide contains cancerous and healthy tissue regions, we also modeled the problem in a Multiple Instance Learning framework. Each biopsy slide is a \textit{bag} containing multiple spectra (called \textit{instances}) and one class label. We assumed that a subset of spectra were diagnostic and the rest were non-diagnostic; only the diagnostic spectra contributed to the classification decision. We proposed three MIL models (mi-SVM, Attention MIL, CVAE MIL) under these assumptions; more details can be found in Appendix~\ref{app:supporting_methods}.

\subsection{The $\beta$-CVAE architecture}
\label{sec:cvae_architecture}


To model the distribution of retained spectra, we implemented a 1D-con\-vo\-lutional $\beta$-regularized conditional variational autoencoder ($\beta$-CVAE). One-dimensional convolutions were chosen to capture the local spectral patterns between adjacent Raman shift frequencies~\cite{mozaffari2020review}, e.g. the broad lipid and protein bands spanning 1400--1500$\,\text{cm}^{-1}$. 

\begin{figure}[H]
    \centering
    \includegraphics[width=\linewidth]{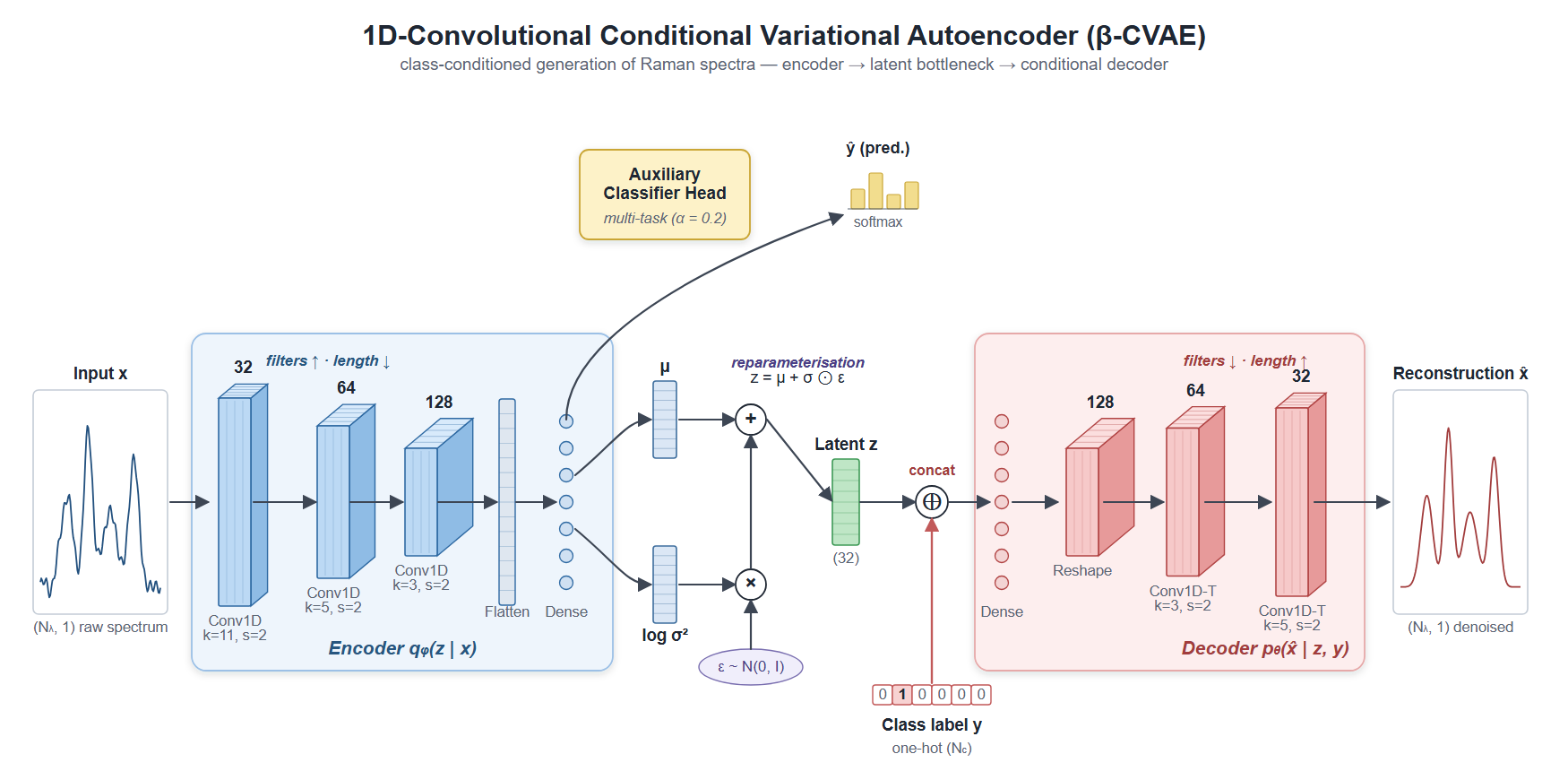}
    \caption{\footnotesize Architecture of the $\beta$-CVAE. The encoder is composed of one-dimensional convolutional layers with 3 blocks: filters $[32, 64, 128]$, kernels $[11, 5, 3]$, and strides $[2, 2, 2]$, compressing the input spectrum to the $d$-dimensional latent space. The decoder reconstructs the spectrum conditioned on both the latent representation and the class label, using transpose convolutions with initial filters of 128, followed by 3 blocks with filters $[128, 64, 32]$, kernels $[3, 5, 11]$, and strides $[2, 2, 2]$. 
    An auxiliary classifier head was attached to the encoder to promote the extraction of class-discriminative features across convolutional layers.}
    \label{fig:cvae_architecture}
\end{figure}

As illustrated in Figure \ref{fig:cvae_architecture}, the overall architecture is composed of an encoder, a $d$-dimensional latent bottleneck, and a decoder. In the model equations, $N_\lambda=571$ denotes the number of input bands after cropping to 700--1800$\,\text{cm}^{-1}$ (Section~\ref{sec:training_details}).

For a given Raman spectrum $x \in \mathbb{R}^{N_\lambda}$, the encoder produced $\mu_\phi(x) \in \mathbb{R}^d$ and $\log \sigma_\phi^2(x) \in \mathbb{R}^d$, parameters that define a Gaussian approximate posterior
\begin{equation}
    q_\phi(z|x) = \mathcal{N}\left(z; \mu_\phi(x), \mathrm{diag}\left(\sigma_\phi^2(x)\right)\right),
\end{equation} where $\phi$ denotes the trainable parameters of the encoder network, i.e. its weights and biases. 

Using the reparameterization trick~\cite{kingma2013auto}, the latent representation $z \in \mathbb{R}^d$ was obtained by sampling from this distribution:
\begin{equation}
    z = \mu_\phi(x) + \sigma_\phi(x) \odot \varepsilon, \quad \varepsilon \sim \mathcal{N}(0, I_d),
\end{equation} where $\odot$ is the element-wise product, and $d \ll N_\lambda$. In our experiments, we tested $d \in \{ 16, 32, 64 \}$, but most of the experiments used $d=32$. To guide the encoder toward capturing class-discriminative features, we framed the encoding process as a Multitask Learning problem~\cite{caruana1997multitask} by appending an auxiliary classification head to the encoder. This classifier $g_\omega$ (with $\omega$ representing its trainable parameters) mapped the latent mean $\mu_\phi(x)$ to class probabilities
\begin{equation}
    \hat{y} = \mathrm{softmax}\left(g_\omega(\mu_\phi(x))\right) \in [0,1]^C, \quad \sum_{c=1}^C \hat{y}_c = 1,
\end{equation} where $C \in \{2, 6\}$ denotes the number of label classes ($C=2$ for binary classification or $C = 6$ for classifying all profiles).

We conditioned the decoder on class label by concatenating $z$ with the one-hot vector $y \in \{0,1\}^C$, giving the input $z \oplus y \in \mathbb{R}^{d + C}$. The decoder network $f_\theta$ produced the reconstructed spectrum 
\begin{equation}
\hat{x} = f_\theta(z \oplus y) \in \mathbb{R}^{N_\lambda},
\end{equation} where $\theta$ denotes the trainable parameters.

Once trained, we used the decoder to generate spectra without an input spectrum by drawing $z$ from the prior $p(z) = \mathcal{N}(0, I_d)$. We formed the class label $y \in \{0,1\}^C$ and used the decoder on the vector $z \oplus y$. 


The model was optimized using a composite loss function:
\begin{equation}
\mathcal{L}_{\text{total}} = \mathcal{L}_{\text{recon}} + \beta \cdot D_{KL}\left(q_\phi(z|x) \| p(z)\right) + \alpha \cdot \mathcal{L}_{\text{class}}. 
\end{equation} The reconstruction loss $\mathcal{L}_{\text{recon}}$ is defined as
\begin{align}
\mathcal{L}_{\text{recon}} &= \gamma \cdot N_{\lambda} \cdot \text{MSE}(x, \hat{x}) + (1 - \gamma) \cdot \text{CosineLoss}(x, \hat{x}) \\
\nonumber &= \gamma \cdot \| x - \hat{x} \|_2^2 + (1 - \gamma) \cdot \left(1 - \frac{x \cdot \hat{x}}{\|x\|_2 \|\hat{x}\|_2}\right). 
 \end{align} We set $\gamma = 1$, except in the reconstruction-loss comparison in Appendix \ref{app:controlled_experiments}.

The Kullback-Leibler divergence ($D_{KL}$) between the encoder distribution $q_\phi(z|x)$ and the prior $p(z)$ has a closed form, since both distributions are Gaussian with diagonal covariance~\cite{higgins2017betavae, Asperti2020}:
\begin{equation}
D_{KL}(q_\phi(z|x) \| p(z)) = \frac{1}{2} \sum_{i=1}^{d} \Big( \mu_{\phi, i}(x)^2 + \sigma_{\phi, i}^2(x) - \log\sigma_{\phi, i}^2(x) - 1 \Big),
\end{equation} where $\mu_{\phi, i}(x)$ and $\sigma_{\phi, i}^2(x)$ are the $i$-th components of the mean and variance vectors. Each term is minimized when $\mu_{\phi, i}(x) = 0$ and $\sigma_{\phi, i}^2(x) = 1$. Thus, the divergence term $D_{KL}$ draws the approximate posterior of each spectrum toward the standard Gaussian prior. The reconstruction term $\mathcal{L}_{\text{recon}}$ counteracts this by requiring $z$ to retain enough information about $x$. The coefficient $\beta$ controls the degree of regularization~\cite{higgins2017betavae}, with smaller values prioritizing spectrum reconstruction fidelity. We evaluated $\beta \in \{0.25, 0.50, 0.75\}$ and used $\beta = 0.50$ as the reference configuration; the alternatives characterize sensitivity within this cohort (Appendix \ref{app:beta_sensitivity}).

The loss of the auxiliary classifier $\mathcal{L}_{\text{class}}$ is the categorical cross-entropy loss, defined as 
\begin{equation}
\mathcal{L}_{\text{class}} = - \sum_{c=1}^{C} y_c \log \hat{y}_c,
\end{equation} where $y_c$ is the $c$-th component of $y$, and $\hat{y}_c$ the predicted probability associated with class $c$. The coefficient $\alpha$ controls how strongly the encoder is driven to separate classes in latent space. As the classification task was complementary and included only for training the encoder, we empirically set $\alpha \le 0.50$ in our experiments. Since $\alpha = 0.50$ led to overfitting the encoder on the classification, we chose a smaller value, $\alpha = 0.20$, resulting in a similar rate of convergence across the classification and the reconstruction tasks.

\subsection{Classification by reconstruction}
\label{sec:cbr}

Standard discriminative models such as Random Forest and XGBoost learn class-separating decision boundaries directly from labeled training data. In Raman studies with few patients, spectra from the same patient are strongly correlated and cannot be treated as independent observations~\cite{guo2021chemometric}; therefore, the effective sample size for generalizing to new patients is closer to the number of patients than to the number of spectra. Instrument drift and fluorescence background introduce additional variability between measurement sessions, and boundaries learned under these conditions can fit cohort-specific features and generalize poorly to unseen patients. 

Generative classifiers, however, model the distribution of the whole spectrum for each class and assign a new spectrum to the class that explains it best. These classifiers can be more sample-efficient when labeled data is scarce~\cite{ng2001discriminative, zheng2023revisiting}. We therefore investigated Classification by Reconstruction (CbR), an inference rule that turns a trained conditional generative model into a classifier~\cite{li2023your}: the model reconstructs each spectrum under every candidate class label, and the class with the smallest reconstruction error is selected. 

We implemented CbR with our $\beta$-CVAE and applied the following steps to each preprocessed, unseen test spectrum $x_{\text{test}}$:

\begin{enumerate}
    \item \emph{Encoding.} The encoder mapped $x_{\text{test}}$ to its posterior mean $\mu_\phi(x_{\text{test}})$ which we used as the latent variable without sampling Gaussian noise. Differences in the reconstruction error between classes were therefore only due to the class label.
    
    \item \emph{Conditional decoding.} The decoder reconstructed the spectrum from $\mu_\phi(x_{\text{test}})$ and the label $y_k$ for each candidate class $k \in \{1, \dots, C\}$: 
    \begin{equation}
        \hat{x}_k = f_\theta(\mu_\phi(x_{\text{test}}) \oplus y_k).
    \end{equation}
    
    \item \emph{Reconstruction error.} The final predicted class $y_{_\text{CbR}}$ is the one that yielded the lowest reconstruction error:
    \begin{equation}
    y_{_\text{CbR}} = \underset{k \in \{1, \dots, C\}}{\arg \min} \left( \| x_{\text{test}} - \hat{x}_k \|_2^2 \right).
    \end{equation}
\end{enumerate}

This approach was based on the assumption that the decoder would reconstruct a spectrum with lower error under its true label than under the other labels.

\section{Experimental design}
\label{sec:experimental_design}


All experiments used five patient-grouped train/test splits (Appendix \ref{app:cv_schema}), such that spectra originating from the same biopsy sample never appeared simultaneously in the training and test partitions of a fold, while taking class balance into account. 

We tested four classifiers (RF+SVM, RF, XGBoost, and a 1D-CNN) on real data, under three training settings: trained on real data (TR/TR), trained on synthetic data (TS/TR), and trained on real data augmented with synthetic data (TSR/TR). The synthetic data was generated with the $\beta$-CVAE. In addition, we evaluated the proposed Classification by Reconstruction (CbR) strategy using the $\beta$-CVAE as the generative classifier. All models were evaluated on the binary classification task (IDH-Mutant versus IDH-Wildtype).

\subsection{Evaluation protocols}

We used separate protocols to assess the performance of the classifiers and the behavior of the MIL selection rules.

\subsubsection{Classifier evaluation on synthetic and augmented data}
We employed the \emph{Train on Synthetic, Test on Real} (TS/TR) framework to evaluate whether the $\beta$-CVAE captured discriminative structure from the training data~\cite{esteban2017real}. For each fold, the $\beta$-CVAE was trained exclusively on the corresponding real training split. We then sampled from the latent prior, $z \sim \mathcal{N}(0, I_d)$, and used the decoder with a class label to generate synthetic spectra per class in each fold. We established the real-data baseline using the \emph{Train on Real, Test on Real} (TR/TR) paradigm. We analyzed four classifiers:
\begin{enumerate}
    \item \emph{RF+SVM:} a Random Forest first selected the 20 most important spectral bands, followed by a Support Vector Machine with RBF kernel and balanced class weights trained on them~\cite{apollonov2024raman}.
    
    \item \emph{Random Forest:} a standard ensemble of 100 trees with balanced class weights.
    
    \item \emph{XGBoost:} a gradient-boosted tree ensemble utilizing histogram-based tree methods optimized for high-dimensional data.
    
    \item \emph{1D-CNN:} a deep convolutional classifier trained with early stopping.
\end{enumerate}

We also evaluated a \emph{Train on Synthetic+Real, Test on Real} (TSR/TR) framework to analyze whether synthetic augmentation improved classification performance relative to the TR/TR baseline. The TSR/TR experiments mixed real and synthetic spectra at a synthetic-to-real ratio; this ratio $r_{\text{S/R}}$ was set to 1 by default, except for a suite of experiments testing the sensitivity of the results for $r_{\text{S/R}} \in \{0.50, 1, 2, 3\}$ (Section \ref{app:synth-ratio} in Appendix \ref{app:controlled_experiments}). Moreover, we evaluated the Classification by Reconstruction (CbR) strategy on the test partition of each fold, and compared its predictions directly against the TR/TR and TS/TR classifiers. These comparisons used the majority-vote (Naive MIL) benchmark for classifier training and slide prediction. 

\subsubsection{MIL evaluation}

We evaluated the MIL models for classifying diagnostic spectra by the following three criteria: performance metrics (bag-level accuracy, sensitivity, specificity, AUC), spatial coherence and key instance utility tests. We detail the MIL evaluation protocol in Appendix~\ref{app:supporting_methods}.

\subsection{Training details}\label{sec:training_details}
To test the feasibility of training classifiers on synthetic data, we measured the performance difference between identical model configurations in TR/TR and TS/TR settings.

We note that the classifiers' hyperparameters were not tuned. Most experiments varied one factor at a time (ablations and sensitivity analyses) and measured the effect on classification performance. All models received spectra cropped to the fingerprint region 700--1800$\,\text{cm}^{-1}$~\cite{Harris2023}, retaining 571 spectral points from the 1738 acquired bands.

To reduce the variability in the number of spectra per slide, we randomly drew a maximum of 500 spectra from each slide, to limit differences in slide contributions to the training set. For the classifiers, we drew one fixed sample per fold. For $\beta$-CVAE training and the 1D-CNN classification model, the sample was redrawn at every epoch and we applied dynamic augmentation to it, using the following operations: 
\begin{enumerate}[label=(\alph*)]
    \item \textit{Within-class mix-up.} We used linear combinations of spectra within the same class to expand the input space: $x = \rho x_1 + (1-\rho) x_2$, with $\rho \sim U(0,0.50)$~\cite{zhang2017mixup}.
    \item \textit{Spectral shift.} We shifted each spectrum along the spectral axis by a random integer drawn uniformly from $[-3,3]$.
    \item \textit{Baseline perturbation.} We added to each spectrum a smooth random baseline, consisting of a weighted sum of two sinusoids with frequencies from $[1,3)$ and $[3,6)$, respectively, with an amplitude of 5\% of the range of the spectrum.
\end{enumerate}

For CVAE training, the optimizer used was Adam, the learning rate was 0.001, the batch size was 256, the configured seed was 42, and the number of epochs varied from 120 to 200, as we observed that the loss plateaus after 100 epochs.

\subsection{Statistical evaluation}
\label{sec:statistical_eval}

Each biopsy slide was treated as a bag with a label, and the spectra within it as instances with no individual ground-truth label. The results of the four classifiers and CbR are reported at bag level in the Naive MIL setting: for each slide, the per-spectrum predictions of a classifier are combined by majority vote into a single decision. Standard accuracy, balanced accuracy, sensitivity, specificity and AUC were computed on the 59 slide-level decisions against the true slide labels, under the five-fold cross-validation schema. Four slides were assigned to test partitions of two folds (Appendix \ref{app:cv_schema}), since two classes (\texttt{LGm3} and \texttt{LGm6}) had four patients each (Table \ref{tab:dataset}); for these slides, the predictions from both folds were averaged before making the decision, ensuring that they are not counted twice. 

To obtain more robust uncertainty estimates, we implemented a stratified cluster bootstrap procedure at the patient level. Across 1,000 bootstrap iterations, we sampled patients with replacement while preserving class proportions; for each patient, we drew all associated slides. For pairwise comparisons between evaluation settings (TS/TR vs. TR/TR or TSR/TR vs. TR/TR), we computed bootstrap distributions of the corresponding accuracy differences:
\[
\Delta = \mathrm{Accuracy}_{A} - \mathrm{Accuracy}_{B}.
\]

Statistical significance was evaluated using bootstrap $p$-value estimates. Confidence intervals (CI) were determined using the bootstrap distribution, and a result is considered significant (and marked with *) when the 95\% CI around $\Delta$ excludes zero. The two-sided $p$-value is defined as \[p = 2 \cdot \min\!\left(\frac{\sum_{i = 1}^N \mathbbm{1}_{[\Delta_i<0]}}{N},\frac{\sum_{i = 1}^N \mathbbm{1}_{[\Delta_i>0]}}{N}\right),\] where $N = 1{,}000$. We note that no statistical correction is applied for multiple tests across the suite of experiments or classifiers; this is an important limitation and inflates the Type~I error rate.

\begin{sidewaysfigure}[H]
    \centering
    \includegraphics[scale=0.9]{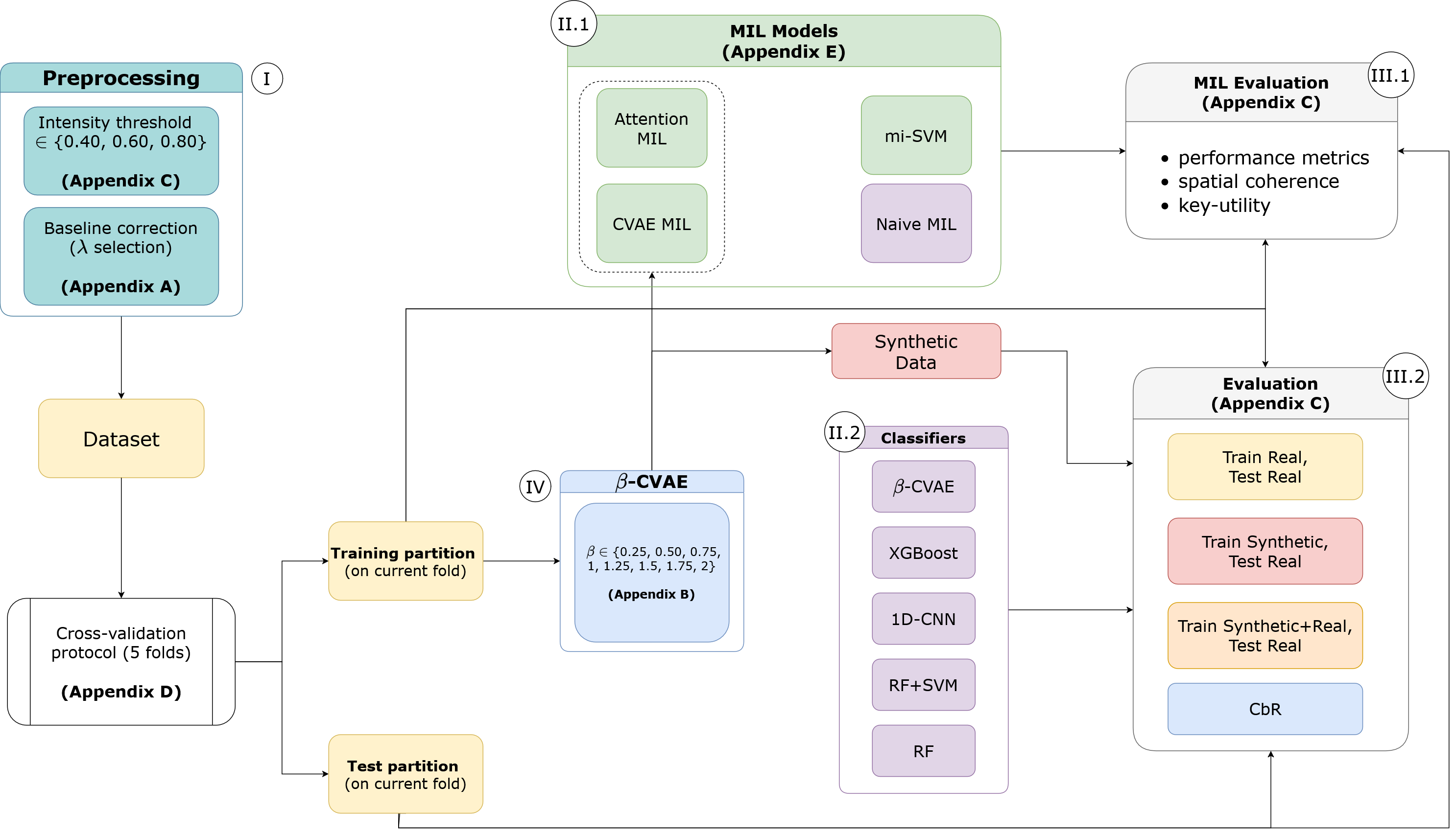}
    \caption{\scriptsize Project workflow: (I) the preprocessing pipeline (with baseline correction as in Appendix~\ref{app:baseline_correction} and with different intensity thresholds as in Appendix~\ref{app:controlled_experiments}) yielded a dataset of filtered Raman spectra. Using a patient-isolated cross-validation protocol (Appendix~\ref{app:cv_schema}) with five folds, we obtained training and test partitions of the real data on each fold. (II) Models: (1) MIL models (Appendix~\ref{app:supporting_methods}); (2) instance-level models. (III) Evaluation: assessing the (1) classification performance and model behavior for different MIL formulations used in identifying relevant training spectra and (2) the performance of instance-level classifiers under TR/TR, TS/TR and TSR/TR regimes trained under Naive MIL setting. (IV) $\beta$-CVAE: The generative model used for (II.1) by defining the Attention MIL and the CVAE MIL, for (II.2) as a generative classifier (with CbR), and as a synthetic data generator for the evaluation protocol (III.2).}
\end{sidewaysfigure}

\section{Results}
\label{sec:results}

We tested the Naive MIL models under the TR/TR, TS/TR and TSR/TR regimes to measure the utility of the synthetic data for training classifiers.
We performed controlled experiments over five aspects: (1) the $\beta$ weight of the $D_{KL}$ regularizing parameter, (2) the latent dimension $d$, (3) the type of the reconstruction loss $\mathcal{L}_{\text{recon}}$, (4) the synthetic to real ratio for TSR/TR; (5) the maximum-intensity threshold \texttt{thr} used as a signal filter during preprocessing. We also tested several MIL formulations against the Naive MIL benchmark.

\subsection{Weight $\beta$ analysis}
The $\beta$ coefficient of the $D_{KL}$ regularizer acts as a tradeoff between the reconstruction capability of the autoencoder and the degree of smoothness of the latent space~\cite{higgins2017betavae}. This property was reflected by the inverse proportionality between the reconstruction loss and the $D_{KL}$ loss (Figure~\ref{fig:beta_sensitivity_combined} in Appendix~\ref{app:beta_sensitivity}). We also studied the variance of the components of the latent vector as given by the CVAE trained with different values of $\beta$ (Table~\ref{tab:beta_kl_dims} in Appendix~\ref{app:beta_sensitivity}).

The results generally reported a better accuracy for the TS/TR and TSR/TR regimes with $\beta=0.50$ (Table \ref{tab:beta_results}). Moreover, for this choice of $\beta$, the models that reported a TS/TR accuracy lower than the TR/TR baseline (RF+SVM, 1D-CNN) had a TSR/TR accuracy greater than the baseline, indicating classification improvement when the dataset was augmented with synthetic data.


\begin{table}[H]
\centering
\scriptsize
\begin{tabular*}{\textwidth}{@{\extracolsep{\fill}}llllll|l}
\toprule
\emph{Regimen} & \emph{$\beta$} & \emph{RF + SVM} & \emph{RF} & \emph{XGBoost} & \emph{1D-CNN} & \emph{CbR}\\
\midrule
TR/TR & -- & $0.763$ & $0.814$ & $0.814$ & $0.712$ & \\
\midrule
\multirow{3}{*}{TS/TR} & $0.25$ & $0.576^{*}$ & $0.797$ & $0.814$ & $0.746$ & $0.746$\\
 & $0.50$ & $0.695$ & $0.831$ & $0.831$ & $0.678$ & $0.780$\\
 & $0.75$ & $0.576^{*}$ & $0.763$ & $0.763$ & $0.712$ & $0.780$\\
\midrule
\multirow{3}{*}{TSR/TR} & $0.25$ & $0.831^{*}$ & $0.831$ & $0.814$ & $0.814^{*}$ & \\
 & $0.50$ & $0.831^{*}$ & $0.831$ & $0.814$ & $0.746$ & \\
 & $0.75$ & $0.814$ & $0.831$ & $0.814$ & $0.729$ & \\
\bottomrule
\end{tabular*}
\caption{\small Effect of the CVAE $D_{KL}$ weight $\beta$ on bag-level binary IDH classification ($d=32$, reconstruction = MSE, synthetic ratio: 1, maximum intensity threshold = 0.60).}
\label{tab:beta_results}
\note{Bag-level accuracies ($n=59$ slides, majority vote). Regimens: TR/TR (real-only), TS/TR (synthetic-only), TSR/TR (synthetic + real). TR/TR is independent of $\beta$. CbR (Classification by Reconstruction) depends on $\beta$ but not on the training regimen. (*) = 95\% CI of the accuracy difference versus TR/TR (same classifier) excluding zero, from a class-stratified bootstrap over patients (1,000 iterations; 46 patients).}
\end{table}


\subsection{Dimension $d$ analysis}
We trained the $\beta$-CVAE model with the latent space dimension $d \in \{16, 32, 64\}$ and analyzed the TS/TR and TSR/TR performance. While the accuracy was generally greater for $d=32$, the effect was not significant and fell within the 95\% CI (Table~\ref{tab:ctrl_boot_latent} in Appendix~\ref{app:controlled_experiments}), with the sole exception of XGBoost in the TS/TR regime, showing a significant increase in accuracy from  $d=16$ to $d=32$. This was in accordance with the finding that only about 3 to 5 latent dimensions are considered active based on the $D_{KL}$ (Table~\ref{tab:beta_kl_dims} in Appendix~\ref{app:beta_sensitivity}).

\begin{table}[H]
\centering
\small
\begin{tabular*}{\textwidth}{@{\extracolsep{\fill}}lcccccc}
\toprule
\multirow{2}{*}{\emph{Model}} & \multicolumn{3}{c}{\emph{TS/TR}} & \multicolumn{3}{c}{\emph{TSR/TR}} \\
\cmidrule(lr){2-4} \cmidrule(lr){5-7}
 & $d=16$ & $d=32$ & $d=64$ & $d=16$ & $d=32$ & $d=64$ \\
\midrule
RF + SVM & $0.627$ & $0.695$ & $0.610$ & $0.831$ & $0.831$ & $0.831$ \\
RF & $0.797$ & $0.831$ & $0.797$ & $0.831$ & $0.831$ & $0.831$ \\
XGBoost & $0.763$ & $0.831$ & $0.780$ & $0.814$ & $0.814$ & $0.814$ \\
1D-CNN & $0.729$ & $0.678$ & $0.746$ & $0.780$ & $0.746$ & $0.797$ \\
\bottomrule
\end{tabular*}
\caption{\small Bag-level accuracy ($n=59$ slides) of the classifiers trained on synthetic data only (TS/TR) and on synthetic and real data (TSR/TR), for CVAE latent dimensions $d \in \{16, 32, 64\}$ ($\beta=0.50$, reconstruction = MSE, synthetic ratio: 1, maximum intensity threshold = 0.60).}
\label{tab:latent_dim_results}
\end{table}

\subsection{Loss $\mathcal{L}_{\text{recon}}$ analysis}

The reported accuracy when using only the (scaled) Mean Squared Error (MSE) was generally greater in both TS/TR and TSR/TR regimens than using only the CosineLoss (Cos) or the average between MSE and Cos, with one exception for the 1D-CNN. Furthermore, the TSR/TR regime is more robust to changes, since the training set includes the real data set.

\begin{table}[H]
\centering
\small
\begin{tabular*}{\textwidth}{@{\extracolsep{\fill}}lcccccc}
\toprule
\multirow{2}{*}{\emph{Model}} & \multicolumn{3}{c}{\emph{TS/TR}} & \multicolumn{3}{c}{\emph{TSR/TR}} \\
\cmidrule(lr){2-4} \cmidrule(lr){5-7}
 & MSE & MSE + Cos & Cos & MSE & MSE + Cos & Cos \\
\midrule
RF + SVM & $0.695$ & $0.576$ & $0.407$ & $0.831$ & $0.814$ & $0.831$ \\
RF & $0.831$ & $0.746$ & $0.644$ & $0.831$ & $0.831$ & $0.814$ \\
XGBoost & $0.831$ & $0.797$ & $0.695$ & $0.814$ & $0.814$ & $0.814$ \\
1D-CNN & $0.678$ & $0.695$ & $0.458$ & $0.746$ & $0.729$ & $0.746$ \\
\bottomrule
\end{tabular*}
\caption{\small Bag-level accuracy ($n=59$ slides) of the classifiers trained on synthetic data only (TS/TR) and on synthetic and real data (TSR/TR), for three CVAE reconstruction losses: mean squared error (MSE), an equal mixture of MSE and cosine loss ($0.50/0.50$, MSE + Cos), and cosine loss (Cos) ($\beta=0.50$, $d=32$, synthetic ratio: 1, maximum intensity threshold = 0.60).}
\label{tab:recon_loss_results}
\end{table}

\subsection{Synthetic/Real ratio analysis}

We analyzed the classification accuracy in the TSR/TR regimen for different numbers of synthetic spectra added to the initial dataset, such that the ratio between the number of synthetic spectra and the number of real spectra (also called \textit{synthetic/real ratio} or \textit{synthetic ratio}) would be in $\{0.50, 1, 2, 3\}$.

A notable difference was observed for the 1D-CNN model when augmenting the real dataset with a number of spectra greater or equal to the size of the dataset: in all cases it performed at least as well as the TR/TR baseline of $0.712$.

\begin{table}[H]
\centering
\small
\begin{tabular*}{\textwidth}{@{\extracolsep{\fill}}lcccccc}
\toprule
\multirow{2}{*}{\emph{Model}} & \multicolumn{2}{c}{\emph{TS/TR}} & \multicolumn{4}{c}{\emph{TSR/TR}} \\
\cmidrule(lr){2-3} \cmidrule(lr){4-7}
 & $40$k & $80$k & $0.5{:}1$ & $1{:}1$ & $2{:}1$ & $3{:}1$ \\
\midrule
RF + SVM & $0.695$ & $0.712$ & $0.831$ & $0.831$ & $0.814$ & $0.831$ \\
RF & $0.831$ & $0.814$ & $0.847$ & $0.831$ & $0.847$ & $0.847$ \\
XGBoost & $0.831$ & $0.831$ & $0.814$ & $0.814$ & $0.814$ & $0.814$ \\
1D-CNN & $0.678$ & $0.780$ & $0.678$ & $0.746$ & $0.712$ & $0.780$ \\
\bottomrule
\end{tabular*}
\caption{\small Bag-level accuracy ($n=59$ slides) of the classifiers trained on synthetic data only (TS/TR; synthetic pools of 40,000 and 80,000 spectra) and on synthetic and real data (TSR/TR) at synthetic-to-real ratios 0.5:1, 1:1, 2:1 and 3:1 ($\beta=0.50$, $d=32$, reconstruction = MSE, maximum intensity threshold = 0.60).}
\label{tab:synth_ratio_results}
\end{table}

\subsection{Intensity threshold analysis}
We observed that by setting the spectra intensity threshold too high in the preprocessing pipeline, the accuracy for the TS/TR regime decreased. We noticed a trend of the models performing better when more spectra were in the dataset, with RF and 1D-CNN slightly better when spectra were selected at a maximum intensity threshold of 0.40. We reasoned that the 0.60 threshold was still an adequate tradeoff.

\begin{table}[H]
\centering
\small
\begin{tabular*}{\textwidth}{@{\extracolsep{\fill}}lcccccc}
\toprule
\multirow{2}{*}{\emph{Model}} & \multicolumn{3}{c}{\emph{TS/TR}} & \multicolumn{3}{c}{\emph{TSR/TR}} \\
\cmidrule(lr){2-4} \cmidrule(lr){5-7}
 & $0.40$ & $0.60$ & $0.8$-adaptive & $0.40$ & $0.60$ & $0.8$-adaptive \\
\midrule
RF + SVM & $0.661$ & $0.695$ & $0.559$ & $0.814$ & $0.831$ & $0.814$ \\
RF & $0.847$ & $0.831$ & $0.780$ & $0.831$ & $0.831$ & $0.831$ \\
XGBoost & $0.831$ & $0.831$ & $0.763$ & $0.814$ & $0.814$ & $0.814$ \\
1D-CNN & $0.712$ & $0.678$ & $0.610$ & $0.712$ & $0.746$ & $0.746$ \\
\bottomrule
\end{tabular*}
\caption{\small Bag-level accuracy ($n=59$ slides) of the classifiers trained on synthetic data only (TS/TR) and on synthetic and real data (TSR/TR), for the maximum-intensity threshold of the strong-signal mask: fixed thresholds of $0.40$ and $0.60$, and an adaptive rule that keeps each slide's top $90\%$ of pixels by maximum intensity subject to an absolute floor of $0.80$ ($\beta=0.50$, $d=32$, reconstruction = MSE, synthetic ratio: 1).}
\label{tab:threshold_results}
\end{table}

\subsection{MIL evaluation}

We compared the best performing Naive MIL implementation (RF) with the other MIL algorithms. While no other MIL formulation achieved a better accuracy than the Naive MIL, the accuracy reported for the CVAE MIL (encoder head) was the closest. This observation remained valid even for other experimental settings, e.g. different values of $\beta$ (Section~\ref{app:mil-results} in Appendix~\ref{app:controlled_experiments}).

\begin{table}[H]
\centering
\scriptsize
\begin{tabular*}{\textwidth}{@{\extracolsep{\fill}}lccccc}
\toprule
\emph{Model} & \emph{Accuracy} & \emph{Bal.\ Acc} & \emph{Sensitivity} & \emph{Specificity} & \emph{AUC} \\
\midrule
Naive MIL (RF) & $0.814$ & $0.797$ & $0.939$ & $0.654$ & $0.790$ \\
\midrule
mi-SVM & $0.763$ & $0.772$ & $0.697$ & $0.846$ & $0.774$ \\
Attention MIL & $0.559$ & $0.529$ & $0.788$ & $0.269$ & $0.521$ \\
CVAE MIL (reconstruction) & $0.729$ & $0.709$ & $0.879$ & $0.538$ & $0.758$ \\
CVAE MIL (encoder head) & $0.797$ & $0.786$ & $0.879$ & $0.692$ & $0.784$ \\
\bottomrule
\end{tabular*}
\caption{\small Bag-level results ($n=59$ slides) of the Multiple Instance Learning (MIL) formulations against the Naive MIL benchmark, a random forest trained on real data in which every instance inherits its bag's label, with one majority-vote decision per bag ($\beta=0.50$, $d=32$, reconstruction = MSE, maximum intensity threshold = 0.60). Attention MIL and CVAE MIL use the $\beta=0.50$ CVAE; mi-SVM is independent of $\beta$.}
\label{tab:mil_results}
\end{table}

\begin{table}[H]
\centering
\scriptsize
\setlength{\tabcolsep}{1.5pt}
\begin{tabular*}{\textwidth}{@{\extracolsep{\fill}}lccccccc}
\toprule
 & \multicolumn{3}{c}{\emph{Moran\textquotesingle{}s $I$}} & \multicolumn{3}{c}{\emph{Var-stand. semivariogram}} & \multirow{2}{*}{\shortstack{\emph{Real $>$}\\\emph{shuf.?}}} \\
\cmidrule(lr){2-4} \cmidrule(lr){5-7}
\emph{Model} & \emph{real} & \emph{shuf.} & \emph{$\Delta$ [95\% CI]} & \emph{real} & \emph{shuf.} & \emph{$\Delta$ [95\% CI]} & \\
\midrule
mi-SVM & $0.005$ & $0.017$ & $-0.013$ $[-0.025,-0.004]$ & $0.010$ & $0.052$ & $-0.042$ $[-0.061,-0.024]$ & no \\
Attention MIL & $0.054$ & $0.066$ & $-0.012$ $[-0.026,+0.001]$ & $0.069$ & $0.077$ & $-0.008$ $[-0.026,+0.008]$ & no \\
CVAE MIL (reconstruction) & $0.157$ & $0.171$ & $-0.014$ $[-0.042,+0.013]$ & $0.279$ & $0.285$ & $-0.006$ $[-0.039,+0.026]$ & no \\
CVAE MIL (encoder head) & $0.059$ & $0.057$ & $+0.003$ $[-0.027,+0.027]$ & $0.076$ & $0.066$ & $+0.010$ $[-0.022,+0.038]$ & no \\
\bottomrule
\end{tabular*}
\caption{\small Spatial coherence of the MIL posteriors for the default configuration ($\beta=0.50$, $d=32$, reconstruction = MSE, maximum intensity threshold = 0.60): Moran's $I$ and the variance-standardized semivariogram, reported as $1-\hat\gamma(3)/\hat\sigma^2$, at Chebyshev distance $\delta=3$, per slide, for the real fit and for a refit under shuffled bag labels. Means are over the slides where both fits are defined. $\Delta$ is the mean paired per-slide difference (real $-$ shuffled) with its 95\% CI from a class-stratified bootstrap over patients (10,000 resamples). The verdict is yes only if both statistics have CIs above 0.}\label{tab:mil_tests_status}
\end{table}

\begin{table}[H]
\centering
\scriptsize
\begin{tabular*}{\textwidth}{@{\extracolsep{\fill}}lccccc}
\toprule
 & \multicolumn{3}{c}{\emph{RF accuracy}} & \multirow{2}{*}{\shortstack{\emph{Sel.\ $>$}\\\emph{naive?}}} & \multirow{2}{*}{\shortstack{\emph{Sel.\ $>$}\\\emph{rand.?}}} \\
\cmidrule(lr){2-4}
\emph{Model} & \emph{Selected} & \emph{Naive} & \emph{Random} & & \\
\midrule
mi-SVM & $0.762$ & $0.813$ & $0.779$ & no & no \\
Attention MIL & $0.796$ & $0.813$ & $0.813$ & no & no \\
CVAE MIL (reconstruction) & $0.762$ & $0.813$ & $0.762$ & no & tie \\
CVAE MIL (encoder head) & $0.813$ & $0.813$ & $0.813$ & tie & tie \\
\bottomrule
\end{tabular*}
\caption{\small Key-instance utility for the default configuration ($\beta=0.50$, $d=32$, reconstruction = MSE, maximum intensity threshold = 0.60): bag accuracy of a random forest trained on the instances selected by a model, against (i) the Naive MIL benchmark, the same forest trained on all instances with all of them voting, and (ii) a forest trained on a size-matched random subset of the training instances.}\label{tab:mil_tests_keyinst}
\end{table}

\section{Discussion}
We studied the use of $\beta$-CVAE, a generative deep learning model, in the context of a small cohort of Raman spectral measurements of glioma biopsies. We employed the model as a:  
(1) data generator for training classifiers;
(2) generative instance classifier;
(3) MIL classifier.

At the reference configuration, RF and XGBoost reached accuracy $0.83$ under TS/TR and $0.81$ under TR/TR (Table~\ref{tab:beta_results}), a net difference of one correctly classified slide. Their reported accuracy-difference intervals included zero (Appendix \ref{tab:ctrl_exp_608}). These close point estimates support retained classification information but establish neither equivalence of training sources nor a reliable improvement. RF+SVM and 1D-CNN performed less well, showing that utility depended on the classifier.

The TSR/TR results also show that the synthetic data generated with the $\beta$-CVAE may be useful as an augmentation technique, offering modest performance improvements relative to the TR/TR baseline. 
Augmentation changed RF+SVM accuracy from $0.76$ to $0.83$, a net gain of four slides, while RF gained one slide and XGBoost had no net change. The 1D-CNN increased from $0.71$ to $0.75$; its sensitivity increased from $0.76$ to $0.91$ while specificity decreased from $0.65$ to $0.54$ (Table~\ref{tab:beta_results}). Overall accuracy therefore needs to be read alongside class-specific performance. The reported bootstrap flags remain exploratory for the reasons given in Section~\ref{sec:statistical_eval}.

The performance of the proposed $\beta$-CVAE model as a generative classifier through CbR is comparable to the top discriminative classifiers (RF and XGBoost), even though it does not outperform them. At the reference configuration, the tested MIL models did not improve on the naive RF benchmark. No selection method exceeded its shuffled-label control on both spatial statistics, and selected-instance RF did not exceed the all-instance benchmark (Tables~\ref{tab:mil_tests_status}--~\ref{tab:mil_tests_keyinst}).

Nonetheless, the study has several limitations which should be acknowledged. The most important limitation is the small size of the dataset, specific to biomedical applications. Currently, the dataset of Lita et al.~\cite{apollonov2024raman} is, to the best of our knowledge, the largest Raman dataset with biopsies labeled according to the methodology of Ceccarelli et al.\cite{Ceccarelli2016}. The performance of the $\beta$-CVAE in both MIL and spectra generation for glioma classification should be inspected with a Raman dataset of greater size. Additionally, the analyzed MIL formulations failed to provide better results than the Naive MIL. Future work should examine whether these MIL approaches offer improvements on larger Raman datasets or whether the MIL approaches tested in this study are not suited for this particular type of data. 

\section*{Data availability}
This study relies on the Raman dataset provided by Lita et al.~\cite{apollonov2024raman}. The raw data can be accessed at 

\href{https://drive.google.com/drive/folders/1CNAwZa3H22FQtprSRXwHXsJ8FdazIUBd}{https://drive.google.com/drive/folders/1CNAwZa3H22FQtprSRXwHXsJ8FdazIUBd}.

\section*{Code availability}

The code for reproducing this work can be found at 

\href{https://github.com/NIRDBS/glioma-synthetic-data}{https://github.com/NIRDBS/glioma-synthetic-data}.

\section*{Author contributions}
Andrei Iu\c san proposed the CVAE architecture, methodology and evaluation protocol, developed and conducted the experiments, and contributed to writing the manuscript.
Daria Voiculescu contributed to the literature review, writing and editing the manuscript, and experimental suggestions.
Ion Petre and Bogdan Oancea formulated the research direction, supervised the project and contributed to writing the manuscript.
Andrei P\u aun and Mihaela P\u aun supervised the project.
All authors read and approved the final manuscript.

\begin{samepage}
\section*{Note on AI usage}
Claude was used for code implementation and manuscript consistency checks, formatting and transcribing results into manuscript tables.
\end{samepage}

\section*{Funding}
This work was supported by the National Institute of Research and Development for Biological Sciences, Romania, and funded under project SIA-PRO no. 23020101, contract no. 7N/2022.

\printbibliography

\newpage
\input{appendix}

\end{document}

%% file: appendix.tex
\appendix

\section{Baseline correction: method selection}
\label{app:baseline_correction}
\setcounter{table}{0}
\renewcommand{\thetable}{\Alph{section}.\arabic{table}}
\setcounter{equation}{0}
\renewcommand{\theequation}{\Alph{section}.\arabic{equation}}

In Raman spectroscopy, baseline correction is required because the acquired signal combines different phenomena, such as the molecular fingerprint we want to capture, background fluorescence, and noise. The most common algorithm for baseline correction is Adaptive Iteratively Re\-weighted Penalized Least Squares (airPLS), with parameters $\lambda$ and $p$-order, which are used to control the shape of the baseline~\cite{zhang2010baseline}. Generalized Cross Validation (GCV)~\cite{craven1978smoothing} and L-curve~\cite{hansen1992analysis} are two methods that can be used to select the best $\lambda$. An alternative algorithm for baseline correction is Improved Asymmetrically Reweighted Penalized Least Squares (IARPLS)~\cite{ye2020baseline}. We designed a suite of experiments to compare airPLS (with $\lambda$ fitted by GCV and L-curve) and IARPLS. Three experiments were run in sequence: (1) fitting $\lambda$ for IARPLS, (2) comparing the three correctors against known ground truth, (3) an invariance test on real spectra.

\subsection*{Synthetic data generator}
We defined a synthetic spectra generator to fit the parameters of the baseline correction to data that looks approximately like our dataset, without fitting to our dataset specifically. We chose this approach to avoid fitting parameters to the full dataset, parameters that would be specific to this dataset and probably not representative for the data population.

The generator defines the signal as a sum of peaks, baseline, and noise:
\begin{equation}
   s(\nu) = p(\nu) + b(\nu) + \varepsilon(\nu), 
\end{equation} where $\nu$ is the wavenumber (1738 points in the interval 50--3400$\,\text{cm}^{-1}$). The peaks $p(\cdot)$ signal is a sum of 14 Gaussians chosen at biologically relevant positions (Table \ref{tab:baseline_generator_bands}). 

Relative amplitudes and width sizes are drawn randomly from uniform distributions around each band's center values to simulate a wide range of spectra.

\begin{table}[H]
\centering
\small
\begin{tabular*}{\textwidth}{@{\extracolsep{\fill}}lccc}
\toprule
Region & Bands ($\text{cm}^{-1}$) & Relative amplitude & Width ($\text{cm}^{-1}$) \\
\midrule
Protein / nucleic acid & \makecell{1004, 1032, 1128,\\ 1230, 1260, 1300, 1340} & $0.30$--$1.00$ & $8$--$18$ \\
\midrule
Lipid / protein & 1440, 1550, 1660 & $0.35$--$1.20$ & $14$--$20$ \\
\midrule
Lipid CH stretch & 2850, 2885, 2930 & $1.45$--$1.80$ & $18$--$22$ \\
\midrule
Aromatic & 3060 & $0.30$ & $20$ \\
\bottomrule
\end{tabular*}
\caption{\small Synthetic band positions, relative amplitudes, and width.}
\label{tab:baseline_generator_bands}
\end{table}


\subsection*{IARPLS $\lambda$ selection}
\begin{table}[H]
\centering
\small
\begin{tabular*}{\textwidth}{@{\extracolsep{\fill}}lccc}
\toprule
$\lambda$ & Low & Medium & High \\
\midrule
$10^2$ & $0.0052$ & $0.0052$ & $0.0052$ \\
$10^3$ & $0.0048$ & $0.0048$ & $0.0048$ \\
$10^4$ & $0.0046$ & $0.0046$ & $0.0046$ \\
$10^5$ & $0.0046$ & $0.0047$ & $0.0269$ \\
$10^6$ & $0.0047$ & $0.1433$ & $4.6800$ \\
$10^7$ & $0.0351$ & $2.1871$ & $4.6795$ \\
\bottomrule
\end{tabular*}
\caption{\small Fingerprint-region (700--1800$\,\text{cm}^{-1}$) Normalized Root MSE (NRMSE) against $\lambda$, nominal generator, median of 6 spectra per level.}
\label{tab:baseline_lambda_fit}
\end{table}

\begin{table}[H]
\centering
\small
\begin{tabular*}{\textwidth}{@{\extracolsep{\fill}}lccc}
\toprule
$\lambda$ & Smooth & Structured & Sharp \\
\midrule
$10^2$$10^2$ & $0.0167$ & $0.0141$ & $0.0315$ \\
$3.2\times10^2$ & $0.0144$ & $0.0094$ & $0.0325$ \\
$10^3$ & $0.0107$ & $0.0100$ & $0.5404$ \\
$3.2\times10^3$ & $0.0122$ & $0.0099$ & $0.6914$ \\
$10^4$ & $0.0119$ & $0.0120$ & $0.7369$ \\
$10^5$ & $0.0116$ & $0.3036$ & $1.0711$ \\
$10^6$ & $0.6160$ & $2.0387$ & $2.3219$ \\
\bottomrule
\end{tabular*}
\caption{\small Fingerprint-region NRMSE vs.\ $\lambda$ across three synthetic baseline curvature families (smooth, structured, sharp), median over lineshape and noise conditions, $n=12$ per cell.}
\label{tab:baseline_curvature_family}
\end{table}

\begin{table}[H]
\centering
\small

\begin{tabular*}{\textwidth}{@{\extracolsep{\fill}}lcccc}
\toprule
Family & Best $\lambda$ & Best NRMSE & Window & Decades \\
\midrule
Smooth & $10^3$ & $0.0107$ & $10^3$--$10^5$ & $2.00$ \\
Structured & $3.2\times10^2$ & $0.0094$ & $3.2\times10^2$--$3.2\times10^3$ & $1.00$ \\
Sharp & $10^2$ & $0.0315$ & $10^2$--$3.2\times10^2$ & $0.50$ \\
\bottomrule
\end{tabular*}
\caption{\small Width of the near-optimal $\lambda$ window (within 25\% of the family's minimum NRMSE).}
\label{tab:baseline_curvature_window}
\end{table}

\subsection*{Comparison against ground truth}
The baseline correctors were tested on synthetic data with known ground truth signal: $n=40$ per fluorescence level, $\lambda=10^4$ for IARPLS.

\begin{table}[H]
\centering
\small
\begin{tabular*}{\textwidth}{@{\extracolsep{\fill}}lcccc}
\toprule
Method & 1004 & 1440 & 1660 & 2930 \\
\midrule
airPLS+GCV & $0.132$ & $0.025$ & $0.017$ & $0.011$ \\
airPLS+L-curve & $0.998$ & $0.995$ & $0.991$ & $0.927$ \\
IARPLS & $1.002$ & $1.003$ & $1.003$ & $1.003$ \\
\bottomrule
\end{tabular*}
\caption{\small Recovered/true peak height (median over fluorescence levels).}
\label{tab:baseline_peak_fidelity}
\end{table}

\begin{table}[H]
\centering
\small
\begin{tabular*}{\textwidth}{@{\extracolsep{\fill}}lccc}
\toprule
Method & Low & Medium & High \\
\midrule
airPLS+GCV & $0.0105$ & $0.1196$ & $0.1135$ \\
airPLS+L-curve & $0.0115$ & $0.0151$ & $0.0287$ \\
IARPLS & $0.0047$ & $0.0048$ & $0.0047$ \\
\bottomrule
\end{tabular*}
\caption{\small Fingerprint-region NRMSE by fluorescence level. GCV pinned $\lambda$ at the grid floor in $83.3\%$ of spectra; L-curve pinned in $0\%$.}
\label{tab:baseline_nrmse_level}
\end{table}

\subsection*{Invariance on real spectra}
Synthetic fluorescence is added to a real spectrum. Each corrector is applied to both versions and the peak-height drift is measured. The peak-height drift is 0 if the corrector is invariant.

\begin{table}[H]
\centering
\small
\begin{tabular*}{\textwidth}{@{\extracolsep{\fill}}lccc}
\toprule
Method & Low & Medium & High \\
\midrule
airPLS+GCV & $0.493$ & $0.552$ & $0.860$ \\
airPLS+L-curve & $0.471$ & $0.545$ & $0.603$ \\
IARPLS & $0.0005$ & $0.0009$ & $0.0043$ \\
\bottomrule
\end{tabular*}
\caption{\small Peak-height drift under added fluorescence, real spectra.}
\label{tab:baseline_invariance}
\end{table}

We chose IARPLS ($\lambda=10^4$) as the baseline corrector as a near-optimal value for the tested generator. Although this parameter is not fitted on the entire dataset, given those results we consider it to be a close to optimal solution. Kizilov et al.~\cite{Kizilov2026} report a similar $\lambda$-sensitivity for IARPLS on non-biological samples.

\newpage
\section{$\beta$ sensitivity analysis}
\label{app:beta_sensitivity}
\setcounter{table}{0}
\renewcommand{\thetable}{\Alph{section}.\arabic{table}}
\setcounter{figure}{0}
\renewcommand{\thefigure}{\Alph{section}.\arabic{figure}}

We characterize how $\beta$, the regularization coefficient of the CVAE and the weight of $D_{KL}$, affects training dynamics and latent-space usage with a flat-$\beta$ sweep over $\beta\in\{0.25,0.50,0.75,1.00,1.25,1.50,1.75,2.00\}$ (\texttt{exp\_607}, \texttt{exp\_608}, \texttt{exp\_609}, \texttt{exp\_601}, \texttt{exp\_602}, \texttt{exp\_603}, \texttt{exp\_604}, \texttt{exp\_501} respectively; the dimension of the latent space is $d=32$ throughout). We also tried a windowed approach, alternating $\beta$ $1.25/1.00$ every 10 epochs in \texttt{exp\_605}, respectively $0.75/1.00$ in \texttt{exp\_606}; both are alternative continuations of \texttt{exp\_601} from epoch 100 until epoch 200. The experiment \texttt{exp\_601} also continues until epoch 200 with constant $\beta=1.00$.

\begin{figure}[H]
\centering
\includegraphics[width=0.82\linewidth]{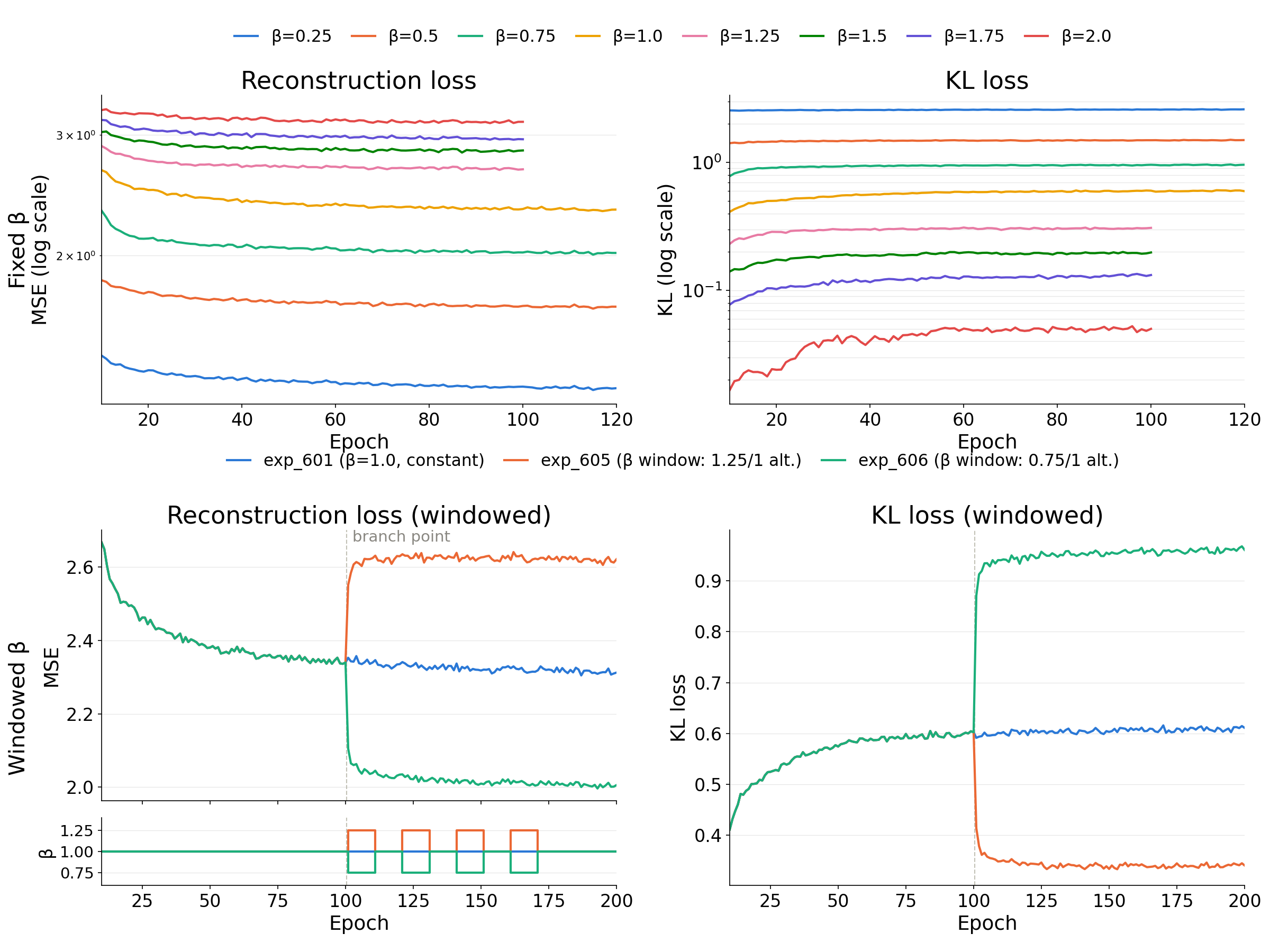}
\caption{\small Top row: reconstruction loss (left) and $D_{KL}$ (right) across the flat-$\beta$ sweep, mean over cross-validations folds, log scale, epochs 10--120 or 10--100 (each series ends at its own experiment's trained epoch count). Higher $\beta$ consistently trades reconstruction fidelity for KL compression, and $D_{KL}$ loss decreases monotonically with $\beta$. Bottom row: the windowed $\beta$-alternative continuation, epochs 10--200: reconstruction loss and the realized $\beta$ trace (left) and $D_{KL}$ (right) for the constant-$\beta$ control arm (\texttt{exp\_601}) vs.\ both windowed schedules (\texttt{exp\_605}, \texttt{exp\_606}), diverging at the epoch-100/101 branch point.}
\label{fig:beta_sensitivity_combined}
\end{figure}

For each configuration, on Fold 0 we compute $\mathrm{KL}_j$ defined as 
\[
\mathrm{KL}_j = \frac{1}{2}\left(\mathbb{E}[\mu_j^2] + \mathbb{E}[\sigma_j^2] - \mathbb{E}[\log\sigma_j^2] - 1\right),
\] where $\mu_j$ and $\sigma_j^2$ are the $j$-th dimensional component of the initial vectors $\mu,\sigma^2$. 

A dimension is counted \emph{active} if $\mathrm{KL}_j > 0.01$; all inactive dimensions carry negligible information (the encoder's posterior for that dimension has collapsed onto the prior).

\begin{table}[H]
\centering
\small
\begin{tabular*}{\textwidth}{@{\extracolsep{\fill}}llcc}
\toprule
$\beta$ & Config & Active, folds 0--4 & Total $D_{KL}$, mean (SD) \\
\midrule
0.25 & \texttt{exp\_607} & 5, 5, 13, 5, 4 / 32 & 2.684 (0.521) \\
0.50 & \texttt{exp\_608} & 3, 3, 5, 4, 4 / 32 & 1.416 (0.279) \\
0.75 & \texttt{exp\_609} & 3, 3, 5, 3, 3 / 32 & 0.861 (0.187) \\
1.00 & \texttt{exp\_601} & 3, 2, 4, 3, 2 / 32 & 0.533 (0.139) \\
1.25 & \texttt{exp\_602} & 2, 1, 1, 1, 1 / 32 & 0.328 (0.113) \\
1.50 & \texttt{exp\_603} & 1, 1, 1, 1, 1 / 32 & 0.199 (0.072) \\
1.75 & \texttt{exp\_604} & 1, 1, 1, 1, 1 / 32 & 0.110 (0.034) \\
Windowed A & \texttt{exp\_605} & 3, 1, 4, 2, 1 / 32 & 0.350 (0.129) \\
Windowed B & \texttt{exp\_606} & 3, 4, 4, 3, 3 / 32 & 0.828 (0.185) \\
\bottomrule
\end{tabular*}
\caption{\small Active latent dimensions ($\mathrm{KL}_j > 0.01$, out of 32) per fold, and total $D_{KL}$ as mean (SD) over the five folds, by $\beta$; each fold's encoder is evaluated on its own test spectra. Windowed A/B (\texttt{exp\_605}/\texttt{exp\_606}) are added for comparison; their $D_{KL}$ reflects a different training history.
}
\label{tab:beta_kl_dims}
\end{table}

The pattern is monotonic in $\beta$, as expected: a stronger $D_{KL}$ penalty produces more aggressive posterior collapse. An important finding is that even at $\beta=0.25$ (the least-collapsed point in the entire sweep) only 5 of 32 latent dimensions carry any information. Strong posterior collapse is present at every $\beta$ tested, with a possible exception at $\beta=0.25$ and only for one fold where the number of active dimensions is 13. In light of the results of the TS/TR and TSR/TR experiments, we conclude that the number of required latent dimensions of the spectra is indeed very low. For fixed $\beta = 0.50$, the effective latent capacity converges to the same $\sim$3--5 active dimensions regardless of $d$ (16, 32, or 64).

\begin{table}[H]
\centering
\small
\begin{tabular*}{\textwidth}{@{\extracolsep{\fill}}llcc}
\toprule
Latent dimension $d$ & Config & Active, folds 0--4 & Total $D_{KL}$, mean (SD) \\
\midrule
16 & \texttt{exp\_701} & 3, 3, 4, 4, 4 / 16 & 1.469 (0.265) \\
32 & \texttt{exp\_608} & 3, 3, 5, 4, 4 / 32 & 1.416 (0.279) \\
64 & \texttt{exp\_702} & 4, 3, 3, 3, 3 / 64 & 1.425 (0.208) \\
\bottomrule
\end{tabular*}
\caption{\small Active latent dimensions per fold and total $D_{KL}$ as mean (SD) over the five folds vs.\ nominal $d$, at fixed $\beta=0.50$.}
\label{tab:beta_dim_crosscheck}
\end{table}

\newpage

\input{appendix_controlled_experiments}

\newpage
\section{Stratified evaluation protocol}
\label{app:cv_schema}
\setcounter{table}{0}
\renewcommand{\thetable}{\Alph{section}.\arabic{table}}

This appendix details the five-fold evaluation protocol adopted from Lita et al.~\cite{apollonov2024raman}, which we applied to all 59 slides of the dataset. Due to the class imbalance (Table \ref{tab:dataset}), a strictly patient-stratified partition would leave classes \texttt{LGm3} or \texttt{LGm6} absent from the test set of at least one fold. Therefore, the fold assignments were specified manually, such that every class is represented in the test partition of every fold. As a consequence, the slides from two patients (color coded with \textcolor{cyan}{cyan} and \textcolor{blue}{blue}, respectively) in the underrepresented classes \texttt{LGm3} and \texttt{LGm6} (\textcolor{cyan}{\texttt{HF-2104\_\#5\_1.h5\_0}}, \textcolor{cyan}{\texttt{HF-2104\_\#9\_1.h5\_1}}, \textcolor{cyan}{\texttt{HF-2104\_V1T\_1.h5\_2}}, \textcolor{blue}{\texttt{HF-592\_V3T\_1.h5\_4}}) appear in the test partitions of two folds (folds \texttt{0} and \texttt{4}) and in the training partitions of the remaining three folds; all other slides appear in exactly one test partition. Accordingly, the sum of test-set sizes across all five folds (63) exceeds the number of unique slides in the dataset (59).

Within each fold, the training partition is the exact complement of the test partition, thus the training sets can be recovered from Listing \ref{lst:test-list}. To prevent patient-level data leakage, all slides acquired from the same patient are kept together and assigned exclusively to either the training or the test partition of each fold.

\vspace*{1cm}

\begin{table}[H]
\centering
\begin{tabular}{lrr}
\toprule
Fold & Train slides & Test slides \\
\midrule
0 & 45 & 14 \\
1 & 47 & 12 \\
2 & 45 & 14 \\
3 & 48 & 11 \\
4 & 47 & 12 \\
\bottomrule
\end{tabular}
\caption{Stratified fold sizes.}
\label{tab:cv_folds_summary}
\end{table}



\newpage

\begin{lstlisting}[label=lst:test-list, caption=Test assignments.]
~Fold 0 (14 slides):~
HF-448_V5B_1.h5_3        HF-305_v4b_1_1.h5_6      HF-615_V5BB_1.h5_9
@HF-2104_#5_1.h5_0@        @HF-2104_#9_1.h5_1@        @HF-2104_V1T_1.h5_2@
HF-442_V4BB_1.h5_12      HF-1002_V1AT_1.h5_0      HF-1002_V2AT_1.h5_1
HF-682_V3AT_1.h5_9       HF-682_V3BB_1.h5_10      HF-894_9_1.h5_11
HF-894_V1BB_1.h5_12      !HF-592_V3T_1.h5_4!

~Fold 1 (12 slides):~
HF-868_1_2.h5_4          HF-901_V2T_2.h5_10        HF-960_VIAT_2.h5_11
HF-2614_V1B_1.h5_3       HF-1825_V2B_1.h5_2        HF-2102_V2BB_1.h5_3
HF-2102_V3AM_1.h5_4      HF-2102_V3AM_2.h5_5       HF-988_V1-T_1.h5_13
HF-988_V1B_1.h5_14       HF-1043_V1AM_1.h5_0       HF-2106_V3AM_1.h5_0

~Fold 2 (14 slides):~
HF-1293_13_1.h5_0        HF-1010_V1T_1.h5_0         HF-1016_IAT_2.h5_1
HF-1334_V58-B_2_1.h5_2   HF-2849_VIT2_1.h5_4        HF-2849_VIT2_1.h5_5
HF-2849_VIT2_2.h5_6      HF-2849_VIT_2_new2021.h5_7 HF-2454_V1AT_1.h5_6
HF-2548_V1T_1.h5_7       HF-1086_#1_1.h5_1          HF-2355_V2AM_1.h5_2
HF-2493_V1T_1.h5_1       HF-2493_V1T_2.h5_2

~Fold 3 (11 slides):~
HF-1295_V3AM_2.h5_1      HF-2070_V1T_1.h5_4        HF-2776_V2B_2.h5_5
HF-2852_VIT_2_2.h5_8     HF-2715_VIL_1.h5_8        HF-2802_V3T_1.h5_9
HF-2485_V1B_1.h5_3       HF-2600_V1B_1.h5_4        HF-2608_V1T_1.h5_5
HF-2544_V1B_1.h5_3       HF-1887_via-t_2_1.h5_3

~Fold 4 (12 slides):~
HF-2534_V2B_1.h5_2       HF-3271_VIB_2.h5_7        HF-3337_V3T_1.h5_8
@HF-2104_#5_1.h5_0@        @HF-2104_#9_1.h5_1@         @HF-2104_V1T_1.h5_2@
HF-2876_V1T_1.h5_10      HF-2898_V1T_1.h5_11       HF-2619_V1T_1.h5_6
HF-2619_V4T_1.h5_7       HF-2666_V2B_1.h5_8        !HF-592_V3T_1.h5_4!

\end{lstlisting}

\newpage

\input{appendix_mil}

%% file: appendix_controlled_experiments.tex
\section{Controlled experiments: downstream classifier results}
\label{app:controlled_experiments}
\setcounter{table}{0}
\renewcommand{\thetable}{\Alph{section}.\arabic{table}}
\setcounter{figure}{0}
\renewcommand{\thefigure}{\Alph{section}.\arabic{figure}}

We report complete results of the downstream classifiers for all of the experiments in this project. We select five aspects, each varying by one factor: (1) the CVAE's $D_{KL}$ loss weight $\beta$, (2) the nominal latent dimension $d$, (3) the reconstruction-loss weighting $\gamma$, (4) the synthetic/real ratio used by TSR/TR, and (5) the strong-signal mask's max-intensity threshold \texttt{thr}. We consider the experimental configuration with $\beta=0.50$, $d=32$, $\gamma=1$, synthetic/real ratio = 1, \texttt{thr}=0.60 to be the \textit{reference configuration} and the results of the other configurations are compared against it.

We report the bag-level accuracy for each training regimen (TR/TR: real only; TS/TR: synthetic only; TSR/TR: synthetic+real) under the Naive MIL approach (one majority-vote decision per bag, i.e. per biopsy slide), and the bootstrap difference $\Delta$ relative to the TR/TR baseline with its 95\% bootstrap confidence interval; (*) marks a 95\% CI excluding zero, and four additional bag-level metrics: balanced accuracy, sensitivity, specificity, and AUC.

The bootstrap $\Delta$/CI/significance columns are computed for raw accuracy. We report the balanced accuracy as well; given the fairly balanced nature of this dataset for the binary IDH classification task, the two metrics are close.
The CbR row reports the bag-level accuracy and 95\% CI for it, and it does not represent a difference vs.\ TR/TR.

Confidence intervals of $[+0.000, +0.000]$ can appear when two compared models agree on all decisions for every sample. Since we test the difference using bootstrap samples, the difference is null for each sample when the models agree perfectly. 

For all configuration files for all models, the random seed was set to 42.
\newpage
\input{appendix_controlled_experiments_tables}

\subsection*{MIL results}\label{app:mil-results}

We ran MIL experiments for the most relevant configurations. Below are all the results with the same metrics as the downstream classifiers: accuracy, balanced accuracy, sensitivity, specificity, and AUC. We note that mi-SVM is independent of $\beta$, and the others are built on top of the CVAE. For each experiment we report three tables: the bag-level classification metrics against the Naive MIL benchmark, the spatial coherence of the posterior maps, and the key-instance utility test.

The spatial-coherence tables compare each tier's real fit with a refit under shuffled bag labels using Moran's $I$ and the variance-standardized semivariogram. We test the difference (real$-$shuffled) slide by slide. 

We compute a 95\% percentile CI from a class-stratified bootstrap over patients of the per-slide differences between real and shuffled metrics (10,000 resamples, seed 42); the real fit is considered more coherent than its control on a statistic only if the CI lies above zero, and a tier passes only if it does so on both statistics. 

In the key-instance utility test, a random forest (RF) is trained only on the training instances that a tier selects as key (non-diagnostic posterior mass below the median of the training fold), and classifies each held-out slide by majority vote over its selected instances. It is compared with the Naive MIL benchmark and with a forest trained on a size-matched random subset of the training instances.

\input{results_mil_appendix_tables}


\subsection*{Per-fold downstream-classifier accuracy: \texttt{exp\_608}}

TR/TR bag-level accuracy for each fold's held-out bags. The models display high variability in performance between folds, due to the small dataset size.

\begin{table}[H]
\centering
\scriptsize
\begin{tabular*}{\textwidth}{@{\extracolsep{\fill}}lccccc}
\toprule
Model & Fold 0 ($n{=}14$) & Fold 1 ($n{=}12$) & Fold 2 ($n{=}14$) & Fold 3 ($n{=}11$) & Fold 4 ($n{=}12$) \\
\midrule
RF + SVM & $0.643$ & $0.750$ & $0.857$ & $0.818$ & $0.583$ \\
RF & $0.643$ & $0.833$ & $0.929$ & $0.909$ & $0.583$ \\
XGBoost & $0.643$ & $0.833$ & $0.929$ & $0.909$ & $0.583$ \\
1D-CNN & $0.571$ & $0.833$ & $0.571$ & $0.818$ & $0.583$ \\
\bottomrule
\end{tabular*}
\caption{\small Per-fold TR/TR bag-level accuracy, \texttt{exp\_608}.}
\label{tab:ctrl_fold_accuracy}

\end{table}

%% file: appendix_controlled_experiments_tables.tex
\subsection{CVAE $\beta$}
The effect of $\beta$ on the classifiers.
\begin{table}[H]
\centering
\scriptsize

\begin{tabular*}{\textwidth}{@{\extracolsep{\fill}}llccccccc}
\toprule
\emph{Model} & \emph{Regimen} & \emph{Accuracy} & \emph{$\Delta$} & \emph{95\% CI} & \emph{Bal.\ Acc} & \emph{Sensitivity} & \emph{Specificity} & \emph{AUC}\\
\midrule
\multirow{3}{*}{RF + SVM} & TR/TR & $0.763$ & -- & -- & $0.751$ & $0.848$ & $0.654$ & $0.745$\\
 & TS/TR & $0.559$ & $-0.203$ & $[-0.421, +0.018]$ & $0.586$ & $0.364$ & $0.808$ & $0.565$\\
 & TSR/TR & $0.831$ & $+0.068$* & $[+0.016, +0.138]$ & $0.816$ & $0.939$ & $0.692$ & $0.761$\\
\midrule
\multirow{3}{*}{RF} & TR/TR & $0.814$ & -- & -- & $0.797$ & $0.939$ & $0.654$ & $0.790$\\
 & TS/TR & $0.729$ & $-0.085$ & $[-0.183, +0.015]$ & $0.705$ & $0.909$ & $0.500$ & $0.758$\\
 & TSR/TR & $0.831$ & $+0.017$ & $[+0.000, +0.055]$ & $0.812$ & $0.970$ & $0.654$ & $0.786$\\
\midrule
\multirow{3}{*}{XGBoost} & TR/TR & $0.814$ & -- & -- & $0.797$ & $0.939$ & $0.654$ & $0.780$\\
 & TS/TR & $0.763$ & $-0.051$ & $[-0.127, +0.017]$ & $0.751$ & $0.848$ & $0.654$ & $0.774$\\
 & TSR/TR & $0.814$ & $+0.000$ & $[+0.000, +0.000]$ & $0.797$ & $0.939$ & $0.654$ & $0.789$\\
\midrule
\multirow{3}{*}{1D-CNN} & TR/TR & $0.712$ & -- & -- & $0.706$ & $0.758$ & $0.654$ & $0.790$\\
 & TS/TR & $0.542$ & $-0.169$ & $[-0.390, +0.054]$ & $0.534$ & $0.606$ & $0.462$ & $0.544$\\
 & TSR/TR & $0.729$ & $+0.017$ & $[-0.108, +0.158]$ & $0.713$ & $0.848$ & $0.577$ & $0.812$\\
\midrule
CbR & -- & $0.678$ & -- & $[+0.534, +0.812]$ & $0.684$ & $0.636$ & $0.731$ & $0.732$\\
\bottomrule
\end{tabular*}
\caption{\small \texttt{exp\_501}: IDH binary classification, $\beta = 2$, $d=32$, reconstruction = MSE, $r_{S/R}=1$, $\texttt{thr} = 0.60$.}
\label{tab:ctrl_exp_501}
\end{table}

\begin{table}[H]
\centering
\scriptsize

\begin{tabular*}{\textwidth}{@{\extracolsep{\fill}}llccccccc}
\toprule
\emph{Model} & \emph{Regimen} & \emph{Accuracy} & \emph{$\Delta$} & \emph{95\% CI} & \emph{Bal.\ Acc} & \emph{Sensitivity} & \emph{Specificity} & \emph{AUC}\\
\midrule
\multirow{3}{*}{RF + SVM} & TR/TR & $0.763$ & -- & -- & $0.751$ & $0.848$ & $0.654$ & $0.745$\\
 & TS/TR & $0.593$ & $-0.169$* & $[-0.309, -0.036]$ & $0.587$ & $0.636$ & $0.538$ & $0.605$\\
 & TSR/TR & $0.831$ & $+0.068$* & $[+0.016, +0.138]$ & $0.816$ & $0.939$ & $0.692$ & $0.809$\\
\midrule
\multirow{3}{*}{RF} & TR/TR & $0.814$ & -- & -- & $0.797$ & $0.939$ & $0.654$ & $0.790$\\
 & TS/TR & $0.729$ & $-0.085$ & $[-0.180, +0.000]$ & $0.700$ & $0.939$ & $0.462$ & $0.685$\\
 & TSR/TR & $0.831$ & $+0.017$ & $[+0.000, +0.055]$ & $0.812$ & $0.970$ & $0.654$ & $0.783$\\
\midrule
\multirow{3}{*}{XGBoost} & TR/TR & $0.814$ & -- & -- & $0.797$ & $0.939$ & $0.654$ & $0.780$\\
 & TS/TR & $0.746$ & $-0.068$ & $[-0.154, +0.018]$ & $0.720$ & $0.939$ & $0.500$ & $0.762$\\
 & TSR/TR & $0.814$ & $+0.000$ & $[+0.000, +0.000]$ & $0.797$ & $0.939$ & $0.654$ & $0.787$\\
\midrule
\multirow{3}{*}{1D-CNN} & TR/TR & $0.712$ & -- & -- & $0.706$ & $0.758$ & $0.654$ & $0.790$\\
 & TS/TR & $0.661$ & $-0.051$ & $[-0.210, +0.109]$ & $0.644$ & $0.788$ & $0.500$ & $0.702$\\
 & TSR/TR & $0.712$ & $+0.000$ & $[-0.161, +0.160]$ & $0.685$ & $0.909$ & $0.462$ & $0.782$\\
\midrule
CbR & -- & $0.729$ & -- & $[+0.597, +0.855]$ & $0.733$ & $0.697$ & $0.769$ & $0.681$\\
\bottomrule
\end{tabular*}
\caption{\small \texttt{exp\_601}: IDH binary classification, $\beta = 1$, $d=32$, reconstruction = MSE, $r_{S/R}=1$, $\texttt{thr} = 0.60$.}
\label{tab:ctrl_exp_601}
\end{table}

\begin{table}[H]
\centering
\scriptsize
\begin{tabular*}{\textwidth}{@{\extracolsep{\fill}}llccccccc}
\toprule
\emph{Model} & \emph{Regimen} & \emph{Accuracy} & \emph{$\Delta$} & \emph{95\% CI} & \emph{Bal.\ Acc} & \emph{Sensitivity} & \emph{Specificity} & \emph{AUC}\\
\midrule
\multirow{3}{*}{RF + SVM} & TR/TR & $0.763$ & -- & -- & $0.751$ & $0.848$ & $0.654$ & $0.745$\\
 & TS/TR & $0.576$ & $-0.186$* & $[-0.338, -0.033]$ & $0.519$ & $1.000$ & $0.038$ & $0.744$\\
 & TSR/TR & $0.831$ & $+0.068$* & $[+0.016, +0.138]$ & $0.816$ & $0.939$ & $0.692$ & $0.786$\\
\midrule
\multirow{3}{*}{RF} & TR/TR & $0.814$ & -- & -- & $0.797$ & $0.939$ & $0.654$ & $0.790$\\
 & TS/TR & $0.797$ & $-0.017$ & $[-0.102, +0.066]$ & $0.769$ & $1.000$ & $0.538$ & $0.787$\\
 & TSR/TR & $0.831$ & $+0.017$ & $[+0.000, +0.055]$ & $0.812$ & $0.970$ & $0.654$ & $0.786$\\
\midrule
\multirow{3}{*}{XGBoost} & TR/TR & $0.814$ & -- & -- & $0.797$ & $0.939$ & $0.654$ & $0.780$\\
 & TS/TR & $0.814$ & $+0.000$ & $[-0.051, +0.050]$ & $0.793$ & $0.970$ & $0.615$ & $0.819$\\
 & TSR/TR & $0.814$ & $+0.000$ & $[+0.000, +0.000]$ & $0.797$ & $0.939$ & $0.654$ & $0.791$\\
\midrule
\multirow{3}{*}{1D-CNN} & TR/TR & $0.712$ & -- & -- & $0.706$ & $0.758$ & $0.654$ & $0.790$\\
 & TS/TR & $0.746$ & $+0.034$ & $[-0.050, +0.129]$ & $0.732$ & $0.848$ & $0.615$ & $0.803$\\
 & TSR/TR & $0.814$ & $+0.102$* & $[+0.016, +0.208]$ & $0.797$ & $0.939$ & $0.654$ & $0.828$\\
\midrule
CbR & -- & $0.746$ & -- & $[+0.621, +0.865]$ & $0.736$ & $0.818$ & $0.654$ & $0.824$\\
\bottomrule
\end{tabular*}
\caption{\small \texttt{exp\_607}: IDH binary classification, $\beta$ = 0.25, $d$=32, reconstruction = MSE, $r_{S/R}=1$, $\texttt{thr} = 0.60$.}
\label{tab:ctrl_exp_607}
\end{table}

\begin{table}[H]
\centering
\scriptsize
\begin{tabular*}{\textwidth}{@{\extracolsep{\fill}}llccccccc}
\toprule
\emph{Model} & \emph{Regimen} & \emph{Accuracy} & \emph{$\Delta$} & \emph{95\% CI} & \emph{Bal.\ Acc} & \emph{Sensitivity} & \emph{Specificity} & \emph{AUC}\\
\midrule
\multirow{3}{*}{RF + SVM} & TR/TR & $0.763$ & -- & -- & $0.751$ & $0.848$ & $0.654$ & $0.745$\\
 & TS/TR & $0.695$ & $-0.068$ & $[-0.193, +0.045]$ & $0.682$ & $0.788$ & $0.577$ & $0.667$\\
 & TSR/TR & $0.831$ & $+0.068$* & $[+0.016, +0.138]$ & $0.816$ & $0.939$ & $0.692$ & $0.803$\\
\midrule
\multirow{3}{*}{RF} & TR/TR & $0.814$ & -- & -- & $0.797$ & $0.939$ & $0.654$ & $0.790$\\
 & TS/TR & $0.831$ & $+0.017$ & $[-0.038, +0.077]$ & $0.812$ & $0.970$ & $0.654$ & $0.755$\\
 & TSR/TR & $0.831$ & $+0.017$ & $[+0.000, +0.055]$ & $0.812$ & $0.970$ & $0.654$ & $0.783$\\
\midrule
\multirow{3}{*}{XGBoost} & TR/TR & $0.814$ & -- & -- & $0.797$ & $0.939$ & $0.654$ & $0.780$\\
 & TS/TR & $0.831$ & $+0.017$ & $[-0.037, +0.074]$ & $0.812$ & $0.970$ & $0.654$ & $0.798$\\
 & TSR/TR & $0.814$ & $+0.000$ & $[+0.000, +0.000]$ & $0.797$ & $0.939$ & $0.654$ & $0.782$\\
\midrule
\multirow{3}{*}{1D-CNN} & TR/TR & $0.712$ & -- & -- & $0.706$ & $0.758$ & $0.654$ & $0.790$\\
 & TS/TR & $0.678$ & $-0.034$ & $[-0.167, +0.086]$ & $0.675$ & $0.697$ & $0.654$ & $0.768$\\
 & TSR/TR & $0.746$ & $+0.034$ & $[-0.113, +0.186]$ & $0.724$ & $0.909$ & $0.538$ & $0.786$\\
\midrule
CbR & -- & $0.780$ & -- & $[+0.655, +0.891]$ & $0.774$ & $0.818$ & $0.731$ & $0.760$\\
\bottomrule
\end{tabular*}
\caption{\small \texttt{exp\_608}: IDH binary classification, $\beta$ = 0.50, $d$=32, reconstruction = MSE, $r_{S/R}=1$, $\texttt{thr} = 0.60$.}
\label{tab:ctrl_exp_608}
\end{table}

\begin{table}[H]
\centering
\scriptsize
\begin{tabular*}{\textwidth}{@{\extracolsep{\fill}}llccccccc}
\toprule
\emph{Model} & \emph{Regimen} & \emph{Accuracy} & \emph{$\Delta$} & \emph{95\% CI} & \emph{Bal.\ Acc} & \emph{Sensitivity} & \emph{Specificity} & \emph{AUC}\\
\midrule
\multirow{3}{*}{RF + SVM} & TR/TR & $0.763$ & -- & -- & $0.751$ & $0.848$ & $0.654$ & $0.745$\\
 & TS/TR & $0.576$ & $-0.186$* & $[-0.357, -0.019]$ & $0.556$ & $0.727$ & $0.385$ & $0.599$\\
 & TSR/TR & $0.814$ & $+0.051$ & $[+0.000, +0.115]$ & $0.801$ & $0.909$ & $0.692$ & $0.808$\\
\midrule
\multirow{3}{*}{RF} & TR/TR & $0.814$ & -- & -- & $0.797$ & $0.939$ & $0.654$ & $0.790$\\
 & TS/TR & $0.763$ & $-0.051$ & $[-0.140, +0.033]$ & $0.739$ & $0.939$ & $0.538$ & $0.739$\\
 & TSR/TR & $0.831$ & $+0.017$ & $[+0.000, +0.055]$ & $0.812$ & $0.970$ & $0.654$ & $0.781$\\
\midrule
\multirow{3}{*}{XGBoost} & TR/TR & $0.814$ & -- & -- & $0.797$ & $0.939$ & $0.654$ & $0.780$\\
 & TS/TR & $0.763$ & $-0.051$ & $[-0.133, +0.034]$ & $0.739$ & $0.939$ & $0.538$ & $0.763$\\
 & TSR/TR & $0.814$ & $+0.000$ & $[+0.000, +0.000]$ & $0.797$ & $0.939$ & $0.654$ & $0.786$\\
\midrule
\multirow{3}{*}{1D-CNN} & TR/TR & $0.712$ & -- & -- & $0.706$ & $0.758$ & $0.654$ & $0.790$\\
 & TS/TR & $0.712$ & $+0.000$ & $[-0.109, +0.125]$ & $0.722$ & $0.636$ & $0.808$ & $0.716$\\
 & TSR/TR & $0.729$ & $+0.017$ & $[-0.077, +0.122]$ & $0.705$ & $0.909$ & $0.500$ & $0.807$\\
\midrule
CbR & -- & $0.780$ & -- & $[+0.655, +0.895]$ & $0.779$ & $0.788$ & $0.769$ & $0.688$\\
\bottomrule
\end{tabular*}
\caption{\small \texttt{exp\_609}: IDH binary classification, $\beta = 0.75$, $d=32$, reconstruction = MSE, $r_{S/R}=1$, $\texttt{thr} = 0.60$.}
\label{tab:ctrl_exp_609}
\end{table}

\subsection{Latent dimension}
The effect of the CVAE's latent vector size on the classifiers. The number of active components is much smaller (see Appendix \ref{app:beta_sensitivity})
\begin{table}[H]
\centering
\scriptsize

\begin{tabular*}{\textwidth}{@{\extracolsep{\fill}}llccccccc}
\toprule
\emph{Model} & \emph{Regimen} & \emph{Accuracy} & \emph{$\Delta$} & \emph{95\% CI} & \emph{Bal.\ Acc} & \emph{Sensitivity} & \emph{Specificity} & \emph{AUC}\\
\midrule
\multirow{3}{*}{RF + SVM} & TR/TR & $0.763$ & -- & -- & $0.751$ & $0.848$ & $0.654$ & $0.745$\\
 & TS/TR & $0.627$ & $-0.136$* & $[-0.261, -0.016]$ & $0.626$ & $0.636$ & $0.615$ & $0.638$\\
 & TSR/TR & $0.831$ & $+0.068$* & $[+0.016, +0.138]$ & $0.816$ & $0.939$ & $0.692$ & $0.810$\\
\midrule
\multirow{3}{*}{RF} & TR/TR & $0.814$ & -- & -- & $0.797$ & $0.939$ & $0.654$ & $0.790$\\
 & TS/TR & $0.797$ & $-0.017$ & $[-0.081, +0.035]$ & $0.773$ & $0.970$ & $0.577$ & $0.775$\\
 & TSR/TR & $0.831$ & $+0.017$ & $[+0.000, +0.055]$ & $0.812$ & $0.970$ & $0.654$ & $0.783$\\
\midrule
\multirow{3}{*}{XGBoost} & TR/TR & $0.814$ & -- & -- & $0.797$ & $0.939$ & $0.654$ & $0.780$\\
 & TS/TR & $0.763$ & $-0.051$ & $[-0.123, +0.017]$ & $0.739$ & $0.939$ & $0.538$ & $0.787$\\
 & TSR/TR & $0.814$ & $+0.000$ & $[+0.000, +0.000]$ & $0.797$ & $0.939$ & $0.654$ & $0.791$\\
\midrule
\multirow{3}{*}{1D-CNN} & TR/TR & $0.712$ & -- & -- & $0.706$ & $0.758$ & $0.654$ & $0.790$\\
 & TS/TR & $0.729$ & $+0.017$ & $[-0.055, +0.093]$ & $0.721$ & $0.788$ & $0.654$ & $0.780$\\
 & TSR/TR & $0.780$ & $+0.068$ & $[-0.034, +0.180]$ & $0.762$ & $0.909$ & $0.615$ & $0.772$\\
\midrule
CbR & -- & $0.797$ & -- & $[+0.667, +0.917]$ & $0.798$ & $0.788$ & $0.808$ & $0.761$\\
\bottomrule
\end{tabular*}
\caption{\small \texttt{exp\_701}: IDH binary classification, $\beta = 0.50$, $d=16$, reconstruction = MSE, $r_{S/R}=1$, $\texttt{thr} = 0.60$.}
\label{tab:ctrl_exp_701}
\end{table}

\texttt{exp\_608}: shared configuration, see its table above.

\begin{table}[H]
\centering
\scriptsize

\begin{tabular*}{\textwidth}{@{\extracolsep{\fill}}llccccccc}
\toprule
\emph{Model} & \emph{Regimen} & \emph{Accuracy} & \emph{$\Delta$} & \emph{95\% CI} & \emph{Bal.\ Acc} & \emph{Sensitivity} & \emph{Specificity} & \emph{AUC}\\
\midrule
\multirow{3}{*}{RF + SVM} & TR/TR & $0.763$ & -- & -- & $0.751$ & $0.848$ & $0.654$ & $0.745$\\
 & TS/TR & $0.610$ & $-0.153$* & $[-0.304, -0.018]$ & $0.570$ & $0.909$ & $0.231$ & $0.641$\\
 & TSR/TR & $0.831$ & $+0.068$* & $[+0.016, +0.138]$ & $0.816$ & $0.939$ & $0.692$ & $0.803$\\
\midrule
\multirow{3}{*}{RF} & TR/TR & $0.814$ & -- & -- & $0.797$ & $0.939$ & $0.654$ & $0.790$\\
 & TS/TR & $0.797$ & $-0.017$ & $[-0.075, +0.036]$ & $0.773$ & $0.970$ & $0.577$ & $0.793$\\
 & TSR/TR & $0.831$ & $+0.017$ & $[+0.000, +0.055]$ & $0.812$ & $0.970$ & $0.654$ & $0.781$\\
\midrule
\multirow{3}{*}{XGBoost} & TR/TR & $0.814$ & -- & -- & $0.797$ & $0.939$ & $0.654$ & $0.780$\\
 & TS/TR & $0.780$ & $-0.034$ & $[-0.109, +0.046]$ & $0.758$ & $0.939$ & $0.577$ & $0.809$\\
 & TSR/TR & $0.814$ & $+0.000$ & $[+0.000, +0.000]$ & $0.797$ & $0.939$ & $0.654$ & $0.784$\\
\midrule
\multirow{3}{*}{1D-CNN} & TR/TR & $0.712$ & -- & -- & $0.706$ & $0.758$ & $0.654$ & $0.790$\\
 & TS/TR & $0.746$ & $+0.034$ & $[-0.065, +0.138]$ & $0.740$ & $0.788$ & $0.692$ & $0.791$\\
 & TSR/TR & $0.797$ & $+0.085$* & $[+0.015, +0.173]$ & $0.781$ & $0.909$ & $0.654$ & $0.815$\\
\midrule
CbR & -- & $0.780$ & -- & $[+0.658, +0.893]$ & $0.779$ & $0.788$ & $0.769$ & $0.772$\\
\bottomrule
\end{tabular*}
\caption{\small \texttt{exp\_702}: IDH binary classification, $\beta = 0.50$, $d=64$, reconstruction = MSE, $r_{S/R}=1$, $\texttt{thr} = 0.60$.}
\label{tab:ctrl_exp_702}
\end{table}


Paired bag-level bootstrap tests between experiments that differ in one factor: the CVAE latent dimension (Table~\ref{tab:ctrl_boot_latent}) and the size of the synthetic pool used for TS/TR (Table~\ref{tab:ctrl_boot_pool}). The procedure is the one used for the classifier tables above: one majority-vote decision per bag from each experiment's own fold models, class-stratified resampling of the 46 patients, all slides of a drawn patient kept together ($n=59$ paired bag decisions), 1,000 iterations, percentile 95\% CI.

\input{appendix_bootstrap_tables}

\subsection{Reconstruction loss}
\label{app:recon-loss}
We experiment with reconstruction loss as MSE, cosine, or 0.50 MSE + 0.50 cosine.
\begin{table}[H]
\centering
\scriptsize

\begin{tabular*}{\textwidth}{@{\extracolsep{\fill}}llccccccc}
\toprule
\emph{Model} & \emph{Regimen} & \emph{Accuracy} & \emph{$\Delta$} & \emph{95\% CI} & \emph{Bal.\ Acc} & \emph{Sensitivity} & \emph{Specificity} & \emph{AUC}\\
\midrule
\multirow{3}{*}{RF + SVM} & TR/TR & $0.763$ & -- & -- & $0.751$ & $0.848$ & $0.654$ & $0.745$\\
 & TS/TR & $0.576$ & $-0.186$* & $[-0.351, -0.034]$ & $0.556$ & $0.727$ & $0.385$ & $0.483$\\
 & TSR/TR & $0.814$ & $+0.051$ & $[+0.000, +0.115]$ & $0.801$ & $0.909$ & $0.692$ & $0.803$\\
\midrule
\multirow{3}{*}{RF} & TR/TR & $0.814$ & -- & -- & $0.797$ & $0.939$ & $0.654$ & $0.790$\\
 & TS/TR & $0.746$ & $-0.068$ & $[-0.164, +0.018]$ & $0.724$ & $0.909$ & $0.538$ & $0.709$\\
 & TSR/TR & $0.831$ & $+0.017$ & $[+0.000, +0.055]$ & $0.812$ & $0.970$ & $0.654$ & $0.781$\\
\midrule
\multirow{3}{*}{XGBoost} & TR/TR & $0.814$ & -- & -- & $0.797$ & $0.939$ & $0.654$ & $0.780$\\
 & TS/TR & $0.797$ & $-0.017$ & $[-0.091, +0.054]$ & $0.777$ & $0.939$ & $0.615$ & $0.802$\\
 & TSR/TR & $0.814$ & $+0.000$ & $[+0.000, +0.000]$ & $0.797$ & $0.939$ & $0.654$ & $0.790$\\
\midrule
\multirow{3}{*}{1D-CNN} & TR/TR & $0.712$ & -- & -- & $0.706$ & $0.758$ & $0.654$ & $0.790$\\
 & TS/TR & $0.695$ & $-0.017$ & $[-0.161, +0.119]$ & $0.699$ & $0.667$ & $0.731$ & $0.740$\\
 & TSR/TR & $0.729$ & $+0.017$ & $[-0.100, +0.145]$ & $0.725$ & $0.758$ & $0.692$ & $0.816$\\
\midrule
CbR & -- & $0.746$ & -- & $[+0.613, +0.873]$ & $0.748$ & $0.727$ & $0.769$ & $0.679$\\
\bottomrule
\end{tabular*}
\caption{\small \texttt{exp\_703}: IDH binary classification, $\beta$ = 0.50, $d$=32, reconstruction = MSE + Cosine (0.50/0.50), $r_{S/R}=1$, $\texttt{thr} = 0.60$.}
\label{tab:ctrl_exp_703}
\end{table}

\begin{table}[H]
\centering
\scriptsize

\begin{tabular*}{\textwidth}{@{\extracolsep{\fill}}llccccccc}
\toprule
\emph{Model} & \emph{Regimen} & \emph{Accuracy} & \emph{$\Delta$} & \emph{95\% CI} & \emph{Bal.\ Acc} & \emph{Sensitivity} & \emph{Specificity} & \emph{AUC}\\
\midrule
\multirow{3}{*}{RF + SVM} & TR/TR & $0.763$ & -- & -- & $0.751$ & $0.848$ & $0.654$ & $0.745$\\
 & TS/TR & $0.407$ & $-0.356$* & $[-0.549, -0.138]$ & $0.437$ & $0.182$ & $0.692$ & $0.478$\\
 & TSR/TR & $0.831$ & $+0.068$ & $[+0.000, +0.152]$ & $0.812$ & $0.970$ & $0.654$ & $0.749$\\
\midrule
\multirow{3}{*}{RF} & TR/TR & $0.814$ & -- & -- & $0.797$ & $0.939$ & $0.654$ & $0.790$\\
 & TS/TR & $0.644$ & $-0.169$* & $[-0.288, -0.053]$ & $0.600$ & $0.970$ & $0.231$ & $0.751$\\
 & TSR/TR & $0.814$ & $+0.000$ & $[+0.000, +0.000]$ & $0.797$ & $0.939$ & $0.654$ & $0.782$\\
\midrule
\multirow{3}{*}{XGBoost} & TR/TR & $0.814$ & -- & -- & $0.797$ & $0.939$ & $0.654$ & $0.780$\\
 & TS/TR & $0.695$ & $-0.119$* & $[-0.228, -0.018]$ & $0.662$ & $0.939$ & $0.385$ & $0.718$\\
 & TSR/TR & $0.814$ & $+0.000$ & $[+0.000, +0.000]$ & $0.797$ & $0.939$ & $0.654$ & $0.789$\\
\midrule
\multirow{3}{*}{1D-CNN} & TR/TR & $0.712$ & -- & -- & $0.706$ & $0.758$ & $0.654$ & $0.790$\\
 & TS/TR & $0.458$ & $-0.254$* & $[-0.426, -0.071]$ & $0.511$ & $0.061$ & $0.962$ & $0.493$\\
 & TSR/TR & $0.746$ & $+0.034$ & $[-0.109, +0.193]$ & $0.744$ & $0.758$ & $0.731$ & $0.818$\\
\midrule
CbR & -- & $0.576$ & -- & $[+0.443, +0.725]$ & $0.564$ & $0.667$ & $0.462$ & $0.706$\\
\bottomrule
\end{tabular*}
\caption{\small \texttt{exp\_704}: IDH binary classification, $\beta$ = 0.50, $d$=32, reconstruction = Cosine, $r_{S/R}=1$, $\texttt{thr} = 0.60$.}
\label{tab:ctrl_exp_704}
\end{table}

\subsection{Synthetic/real ratio for TSR/TR}
\label{app:synth-ratio}
We change the ratio $r_{S/R}$ of synthetic/real data. For all other experiments, this ratio is fixed at 1.

\begin{table}[H]
\centering
\scriptsize

\begin{tabular*}{\textwidth}{@{\extracolsep{\fill}}llccccccc}
\toprule
\emph{Model} & \emph{Regimen} & \emph{Accuracy} & \emph{$\Delta$} & \emph{95\% CI} & \emph{Bal.\ Acc} & \emph{Sensitivity} & \emph{Specificity} & \emph{AUC}\\
\midrule
\multirow{3}{*}{RF + SVM} & TR/TR & $0.763$ & -- & -- & $0.751$ & $0.848$ & $0.654$ & $0.745$\\
 & TS/TR & $0.695$ & $-0.068$ & $[-0.193, +0.045]$ & $0.682$ & $0.788$ & $0.577$ & $0.667$\\
 & TSR/TR & $0.831$ & $+0.068$* & $[+0.016, +0.138]$ & $0.816$ & $0.939$ & $0.692$ & $0.795$\\
\midrule
\multirow{3}{*}{RF} & TR/TR & $0.814$ & -- & -- & $0.797$ & $0.939$ & $0.654$ & $0.790$\\
 & TS/TR & $0.831$ & $+0.017$ & $[-0.038, +0.077]$ & $0.812$ & $0.970$ & $0.654$ & $0.755$\\
 & TSR/TR & $0.847$ & $+0.034$ & $[+0.000, +0.088]$ & $0.831$ & $0.970$ & $0.692$ & $0.782$\\
\midrule
\multirow{3}{*}{XGBoost} & TR/TR & $0.814$ & -- & -- & $0.797$ & $0.939$ & $0.654$ & $0.780$\\
 & TS/TR & $0.831$ & $+0.017$ & $[-0.037, +0.074]$ & $0.812$ & $0.970$ & $0.654$ & $0.798$\\
 & TSR/TR & $0.814$ & $+0.000$ & $[+0.000, +0.000]$ & $0.797$ & $0.939$ & $0.654$ & $0.790$\\
\midrule
\multirow{3}{*}{1D-CNN} & TR/TR & $0.712$ & -- & -- & $0.706$ & $0.758$ & $0.654$ & $0.790$\\
 & TS/TR & $0.678$ & $-0.034$ & $[-0.167, +0.086]$ & $0.675$ & $0.697$ & $0.654$ & $0.768$\\
 & TSR/TR & $0.678$ & $-0.034$ & $[-0.182, +0.121]$ & $0.700$ & $0.515$ & $0.885$ & $0.728$\\
\midrule
CbR & -- & $0.780$ & -- & $[+0.655, +0.891]$ & $0.774$ & $0.818$ & $0.731$ & $0.760$\\
\bottomrule
\end{tabular*}
\caption{\small \texttt{exp\_705}: IDH binary classification, $\beta$ = 0.50, $d$=32, reconstruction = MSE, $r_{S/R}=0.5$, $\texttt{thr} = 0.60$.}
\label{tab:ctrl_exp_705}
\end{table}

\begin{table}[H]
\centering
\scriptsize

\begin{tabular*}{\textwidth}{@{\extracolsep{\fill}}llccccccc}
\toprule
\emph{Model} & \emph{Regimen} & \emph{Accuracy} & \emph{$\Delta$} & \emph{95\% CI} & \emph{Bal.\ Acc} & \emph{Sensitivity} & \emph{Specificity} & \emph{AUC}\\
\midrule
\multirow{3}{*}{RF + SVM} & TR/TR & $0.763$ & -- & -- & $0.751$ & $0.848$ & $0.654$ & $0.745$\\
 & TS/TR & $0.695$ & $-0.068$ & $[-0.193, +0.045]$ & $0.682$ & $0.788$ & $0.577$ & $0.667$\\
 & TSR/TR & $0.814$ & $+0.051$ & $[+0.000, +0.113]$ & $0.797$ & $0.939$ & $0.654$ & $0.798$\\
\midrule
\multirow{3}{*}{RF} & TR/TR & $0.814$ & -- & -- & $0.797$ & $0.939$ & $0.654$ & $0.790$\\
 & TS/TR & $0.831$ & $+0.017$ & $[-0.038, +0.077]$ & $0.812$ & $0.970$ & $0.654$ & $0.755$\\
 & TSR/TR & $0.847$ & $+0.034$ & $[+0.000, +0.088]$ & $0.831$ & $0.970$ & $0.692$ & $0.787$\\
\midrule
\multirow{3}{*}{XGBoost} & TR/TR & $0.814$ & -- & -- & $0.797$ & $0.939$ & $0.654$ & $0.780$\\
 & TS/TR & $0.831$ & $+0.017$ & $[-0.037, +0.074]$ & $0.812$ & $0.970$ & $0.654$ & $0.798$\\
 & TSR/TR & $0.814$ & $+0.000$ & $[+0.000, +0.000]$ & $0.797$ & $0.939$ & $0.654$ & $0.779$\\
\midrule
\multirow{3}{*}{1D-CNN} & TR/TR & $0.712$ & -- & -- & $0.706$ & $0.758$ & $0.654$ & $0.790$\\
 & TS/TR & $0.678$ & $-0.034$ & $[-0.167, +0.086]$ & $0.675$ & $0.697$ & $0.654$ & $0.768$\\
 & TSR/TR & $0.712$ & $+0.000$ & $[-0.117, +0.132]$ & $0.689$ & $0.879$ & $0.500$ & $0.808$\\
\midrule
CbR & -- & $0.780$ & -- & $[+0.655, +0.891]$ & $0.774$ & $0.818$ & $0.731$ & $0.760$\\
\bottomrule
\end{tabular*}
\caption{\small \texttt{exp\_706}: IDH binary classification, $\beta$ = 0.50, $d$=32, reconstruction = MSE, $r_{S/R}=2$, $\texttt{thr} = 0.60$.}
\label{tab:ctrl_exp_706}
\end{table}

\begin{table}[H]
\centering
\scriptsize

\begin{tabular*}{\textwidth}{@{\extracolsep{\fill}}llccccccc}
\toprule
\emph{Model} & \emph{Regimen} & \emph{Accuracy} & \emph{$\Delta$} & \emph{95\% CI} & \emph{Bal.\ Acc} & \emph{Sensitivity} & \emph{Specificity} & \emph{AUC}\\
\midrule
\multirow{3}{*}{RF + SVM} & TR/TR & $0.763$ & -- & -- & $0.751$ & $0.848$ & $0.654$ & $0.745$\\
 & TS/TR & $0.695$ & $-0.068$ & $[-0.193, +0.045]$ & $0.682$ & $0.788$ & $0.577$ & $0.667$\\
 & TSR/TR & $0.831$ & $+0.068$* & $[+0.016, +0.136]$ & $0.812$ & $0.970$ & $0.654$ & $0.791$\\
\midrule
\multirow{3}{*}{RF} & TR/TR & $0.814$ & -- & -- & $0.797$ & $0.939$ & $0.654$ & $0.790$\\
 & TS/TR & $0.831$ & $+0.017$ & $[-0.038, +0.077]$ & $0.812$ & $0.970$ & $0.654$ & $0.755$\\
 & TSR/TR & $0.847$ & $+0.034$ & $[+0.000, +0.088]$ & $0.831$ & $0.970$ & $0.692$ & $0.784$\\
\midrule
\multirow{3}{*}{XGBoost} & TR/TR & $0.814$ & -- & -- & $0.797$ & $0.939$ & $0.654$ & $0.780$\\
 & TS/TR & $0.831$ & $+0.017$ & $[-0.037, +0.074]$ & $0.812$ & $0.970$ & $0.654$ & $0.798$\\
 & TSR/TR & $0.814$ & $+0.000$ & $[+0.000, +0.000]$ & $0.797$ & $0.939$ & $0.654$ & $0.788$\\
\midrule
\multirow{3}{*}{1D-CNN} & TR/TR & $0.712$ & -- & -- & $0.706$ & $0.758$ & $0.654$ & $0.790$\\
 & TS/TR & $0.678$ & $-0.034$ & $[-0.167, +0.086]$ & $0.675$ & $0.697$ & $0.654$ & $0.768$\\
 & TSR/TR & $0.780$ & $+0.068$ & $[-0.048, +0.196]$ & $0.758$ & $0.939$ & $0.577$ & $0.795$\\
\midrule
CbR & -- & $0.780$ & -- & $[+0.655, +0.891]$ & $0.774$ & $0.818$ & $0.731$ & $0.760$\\
\bottomrule
\end{tabular*}
\caption{\small \texttt{exp\_707}: IDH binary classification, $\beta$ = 0.50, $d$=32, reconstruction = MSE, $r_{S/R} = 3$, $\texttt{thr} = 0.60$.}
\label{tab:ctrl_exp_707}
\end{table}

\subsection{Synthetic pool size for TS/TR}
\label{app:synth-pool}
We double the synthetic pool used for TS/TR from 40,000 spectra (\texttt{exp\_608}) to 80,000 spectra, generated with the same CVAE decoders as \texttt{exp\_608}; the paired comparison of the two pools is in Table~\ref{tab:ctrl_boot_pool}.

\begin{table}[H]
\centering
\scriptsize

\begin{tabular*}{\textwidth}{@{\extracolsep{\fill}}llccccccc}
\toprule
\emph{Model} & \emph{Regimen} & \emph{Accuracy} & \emph{$\Delta$} & \emph{95\% CI} & \emph{Bal.\ Acc} & \emph{Sensitivity} & \emph{Specificity} & \emph{AUC}\\
\midrule
\multirow{2}{*}{RF + SVM} & TR/TR & $0.763$ & -- & -- & $0.751$ & $0.848$ & $0.654$ & $0.745$\\
 & TS/TR & $0.712$ & $-0.051$ & $[-0.167, +0.053]$ & $0.693$ & $0.848$ & $0.538$ & $0.685$\\
\midrule
\multirow{2}{*}{RF} & TR/TR & $0.814$ & -- & -- & $0.797$ & $0.939$ & $0.654$ & $0.790$\\
 & TS/TR & $0.814$ & $+0.000$ & $[-0.051, +0.050]$ & $0.793$ & $0.970$ & $0.615$ & $0.752$\\
\midrule
\multirow{2}{*}{XGBoost} & TR/TR & $0.814$ & -- & -- & $0.797$ & $0.939$ & $0.654$ & $0.780$\\
 & TS/TR & $0.831$ & $+0.017$ & $[-0.054, +0.094]$ & $0.812$ & $0.970$ & $0.654$ & $0.793$\\
\midrule
\multirow{2}{*}{1D-CNN} & TR/TR & $0.712$ & -- & -- & $0.706$ & $0.758$ & $0.654$ & $0.790$\\
 & TS/TR & $0.780$ & $+0.068$ & $[-0.028, +0.172]$ & $0.762$ & $0.909$ & $0.615$ & $0.777$\\
\midrule
CbR & -- & $0.780$ & -- & $[+0.655, +0.891]$ & $0.774$ & $0.818$ & $0.731$ & $0.760$\\
\bottomrule
\end{tabular*}
\caption{\small \texttt{exp\_708}: IDH binary classification, $\beta = 0.50$, $d=32$, reconstruction = MSE, synthetic pool: 80,000 spectra for TS/TR (TSR/TR not run), $\texttt{thr} = 0.60$.}
\label{tab:ctrl_exp_708}
\end{table}

\subsection{Maximum intensity threshold}\label{subsec:max_int}
We test the spectra selection criteria with thresholds $\text{thr} \in \{0.40, 0.60, 0.80\}$. The adaptive rule selects at most the top 90\% of the spectra by maximum intensity conditioned by a floor of $0.80$. We refer to it as $0.80$, although it's not a fixed threshold.
\begin{table}[H]
\centering
\scriptsize

\begin{tabular*}{\textwidth}{@{\extracolsep{\fill}}llccccccc}
\toprule
\emph{Model} & \emph{Regimen} & \emph{Accuracy} & \emph{$\Delta$} & \emph{95\% CI} & \emph{Bal.\ Acc} & \emph{Sensitivity} & \emph{Specificity} & \emph{AUC}\\
\midrule
\multirow{3}{*}{RF + SVM} & TR/TR & $0.746$ & -- & -- & $0.732$ & $0.848$ & $0.615$ & $0.735$\\
 & TS/TR & $0.661$ & $-0.085$ & $[-0.250, +0.075]$ & $0.619$ & $0.970$ & $0.269$ & $0.724$\\
 & TSR/TR & $0.814$ & $+0.068$ & $[+0.000, +0.154]$ & $0.797$ & $0.939$ & $0.654$ & $0.797$\\
\midrule
\multirow{3}{*}{RF} & TR/TR & $0.814$ & -- & -- & $0.797$ & $0.939$ & $0.654$ & $0.770$\\
 & TS/TR & $0.847$ & $+0.034$ & $[+0.000, +0.088]$ & $0.831$ & $0.970$ & $0.692$ & $0.755$\\
 & TSR/TR & $0.831$ & $+0.017$ & $[+0.000, +0.055]$ & $0.812$ & $0.970$ & $0.654$ & $0.784$\\
\midrule
\multirow{3}{*}{XGBoost} & TR/TR & $0.814$ & -- & -- & $0.797$ & $0.939$ & $0.654$ & $0.776$\\
 & TS/TR & $0.831$ & $+0.017$ & $[-0.037, +0.074]$ & $0.812$ & $0.970$ & $0.654$ & $0.789$\\
 & TSR/TR & $0.814$ & $+0.000$ & $[+0.000, +0.000]$ & $0.797$ & $0.939$ & $0.654$ & $0.787$\\
\midrule
\multirow{3}{*}{1D-CNN} & TR/TR & $0.678$ & -- & -- & $0.700$ & $0.515$ & $0.885$ & $0.759$\\
 & TS/TR & $0.712$ & $+0.034$ & $[-0.178, +0.250]$ & $0.718$ & $0.667$ & $0.769$ & $0.759$\\
 & TSR/TR & $0.712$ & $+0.034$ & $[-0.156, +0.224]$ & $0.685$ & $0.909$ & $0.462$ & $0.807$\\
\midrule
CbR & -- & $0.797$ & -- & $[+0.684, +0.914]$ & $0.794$ & $0.818$ & $0.769$ & $0.776$\\
\bottomrule
\end{tabular*}
\caption{\small \texttt{exp\_4.608}: IDH binary classification, $\beta$ = 0.50, $d$=32, reconstruction = MSE, $r_{S/R}=1$, $\texttt{thr} = 0.40$.}
\label{tab:ctrl_exp_4.608}
\end{table}

\begin{table}[H]
\centering
\scriptsize

\begin{tabular*}{\textwidth}{@{\extracolsep{\fill}}llccccccc}
\toprule
\emph{Model} & \emph{Regimen} & \emph{Accuracy} & \emph{$\Delta$} & \emph{95\% CI} & \emph{Bal.\ Acc} & \emph{Sensitivity} & \emph{Specificity} & \emph{AUC}\\
\midrule
\multirow{3}{*}{RF + SVM} & TR/TR & $0.763$ & -- & -- & $0.751$ & $0.848$ & $0.654$ & $0.745$\\
 & TS/TR & $0.695$ & $-0.068$ & $[-0.193, +0.045]$ & $0.682$ & $0.788$ & $0.577$ & $0.667$\\
 & TSR/TR & $0.831$ & $+0.068$* & $[+0.016, +0.138]$ & $0.816$ & $0.939$ & $0.692$ & $0.803$\\
\midrule
\multirow{3}{*}{RF} & TR/TR & $0.814$ & -- & -- & $0.797$ & $0.939$ & $0.654$ & $0.790$\\
 & TS/TR & $0.831$ & $+0.017$ & $[-0.038, +0.077]$ & $0.812$ & $0.970$ & $0.654$ & $0.755$\\
 & TSR/TR & $0.831$ & $+0.017$ & $[+0.000, +0.055]$ & $0.812$ & $0.970$ & $0.654$ & $0.783$\\
\midrule
\multirow{3}{*}{XGBoost} & TR/TR & $0.814$ & -- & -- & $0.797$ & $0.939$ & $0.654$ & $0.780$\\
 & TS/TR & $0.831$ & $+0.017$ & $[-0.037, +0.074]$ & $0.812$ & $0.970$ & $0.654$ & $0.798$\\
 & TSR/TR & $0.814$ & $+0.000$ & $[+0.000, +0.000]$ & $0.797$ & $0.939$ & $0.654$ & $0.782$\\
\midrule
\multirow{3}{*}{1D-CNN} & TR/TR & $0.712$ & -- & -- & $0.706$ & $0.758$ & $0.654$ & $0.790$\\
 & TS/TR & $0.678$ & $-0.034$ & $[-0.167, +0.086]$ & $0.675$ & $0.697$ & $0.654$ & $0.768$\\
 & TSR/TR & $0.746$ & $+0.034$ & $[-0.113, +0.186]$ & $0.724$ & $0.909$ & $0.538$ & $0.786$\\
\midrule
CbR & -- & $0.780$ & -- & $[+0.655, +0.891]$ & $0.774$ & $0.818$ & $0.731$ & $0.760$\\
\bottomrule
\end{tabular*}
\caption{\small \texttt{exp\_608}: IDH binary classification, $\beta$ = 0.50, $d$=32, reconstruction = MSE, $r_{S/R}=1$, $\texttt{thr} = 0.60$.}
\label{tab:ctrl_exp_608_thr}
\end{table}

\begin{table}[H]
\centering
\scriptsize

\begin{tabular*}{\textwidth}{@{\extracolsep{\fill}}llccccccc}
\toprule
\emph{Model} & \emph{Regimen} & \emph{Accuracy} & \emph{$\Delta$} & \emph{95\% CI} & \emph{Bal.\ Acc} & \emph{Sensitivity} & \emph{Specificity} & \emph{AUC}\\
\midrule
\multirow{3}{*}{RF + SVM} & TR/TR & $0.746$ & -- & -- & $0.732$ & $0.848$ & $0.615$ & $0.726$\\
 & TS/TR & $0.559$ & $-0.186$* & $[-0.322, -0.041]$ & $0.573$ & $0.455$ & $0.692$ & $0.608$\\
 & TSR/TR & $0.814$ & $+0.068$ & $[+0.000, +0.154]$ & $0.797$ & $0.939$ & $0.654$ & $0.809$\\
\midrule
\multirow{3}{*}{RF} & TR/TR & $0.780$ & -- & -- & $0.762$ & $0.909$ & $0.615$ & $0.775$\\
 & TS/TR & $0.780$ & $+0.000$ & $[-0.078, +0.075]$ & $0.754$ & $0.970$ & $0.538$ & $0.742$\\
 & TSR/TR & $0.831$ & $+0.051$ & $[+0.000, +0.111]$ & $0.812$ & $0.970$ & $0.654$ & $0.793$\\
\midrule
\multirow{3}{*}{XGBoost} & TR/TR & $0.797$ & -- & -- & $0.777$ & $0.939$ & $0.615$ & $0.790$\\
 & TS/TR & $0.763$ & $-0.034$ & $[-0.134, +0.067]$ & $0.735$ & $0.970$ & $0.500$ & $0.783$\\
 & TSR/TR & $0.814$ & $+0.017$ & $[+0.000, +0.054]$ & $0.797$ & $0.939$ & $0.654$ & $0.789$\\
\midrule
\multirow{3}{*}{1D-CNN} & TR/TR & $0.661$ & -- & -- & $0.681$ & $0.515$ & $0.846$ & $0.756$\\
 & TS/TR & $0.610$ & $-0.051$ & $[-0.184, +0.063]$ & $0.627$ & $0.485$ & $0.769$ & $0.737$\\
 & TSR/TR & $0.746$ & $+0.085$ & $[-0.111, +0.296]$ & $0.716$ & $0.970$ & $0.462$ & $0.823$\\
\midrule
CbR & -- & $0.763$ & -- & $[+0.625, +0.883]$ & $0.759$ & $0.788$ & $0.731$ & $0.755$\\
\bottomrule
\end{tabular*}
\caption{\small \texttt{exp\_2.608}: IDH binary classification, $\beta$ = 0.50, $d$=32, reconstruction = MSE, $r_{S/R}=1$, \texttt{thr} = adaptive, top-90\%-by-rank per slide floored at $0.80$ (v2 preprocessing; see note below).}
\label{tab:ctrl_exp_2.608}
\end{table}

We note that the preprocessing in \texttt{exp\_2.608} is not a pure fixed threshold despite the $0.80$ figure above: it keeps each slide's own top-90\% of pixels by max-intensity rank (a per-slide 10th-percentile cutoff), with $\texttt{thr}>0.80$ applied only as an absolute floor within that selection. Configurations \texttt{exp\_608} and \texttt{exp\_4.608} use a single fixed threshold directly on max-intensity, with no rank-based component.

\subsection{Results for $\beta \in \{0.25, 0.75\}$ at the adaptive 0.80 threshold}
We test two more values for beta for the 0.80 spectra selection threshold.
\begin{table}[H]
\centering
\scriptsize
\begin{tabular*}{\textwidth}{@{\extracolsep{\fill}}llccccccc}
\toprule
\emph{Model} & \emph{Regimen} & \emph{Accuracy} & \emph{$\Delta$} & \emph{95\% CI} & \emph{Bal.\ Acc} & \emph{Sensitivity} & \emph{Specificity} & \emph{AUC}\\
\midrule
\multirow{3}{*}{RF + SVM} & TR/TR & $0.746$ & -- & -- & $0.732$ & $0.848$ & $0.615$ & $0.726$\\
 & TS/TR & $0.644$ & $-0.102$ & $[-0.254, +0.037]$ & $0.608$ & $0.909$ & $0.308$ & $0.589$\\
 & TSR/TR & $0.831$ & $+0.085$* & $[+0.018, +0.161]$ & $0.816$ & $0.939$ & $0.692$ & $0.796$\\
\midrule
\multirow{3}{*}{RF} & TR/TR & $0.780$ & -- & -- & $0.762$ & $0.909$ & $0.615$ & $0.775$\\
 & TS/TR & $0.797$ & $+0.017$ & $[-0.055, +0.089]$ & $0.773$ & $0.970$ & $0.577$ & $0.780$\\
 & TSR/TR & $0.831$ & $+0.051$ & $[+0.000, +0.111]$ & $0.812$ & $0.970$ & $0.654$ & $0.794$\\
\midrule
\multirow{3}{*}{XGBoost} & TR/TR & $0.797$ & -- & -- & $0.777$ & $0.939$ & $0.615$ & $0.790$\\
 & TS/TR & $0.814$ & $+0.017$ & $[-0.055, +0.091]$ & $0.788$ & $1.000$ & $0.577$ & $0.826$\\
 & TSR/TR & $0.814$ & $+0.017$ & $[+0.000, +0.054]$ & $0.797$ & $0.939$ & $0.654$ & $0.787$\\
\midrule
\multirow{3}{*}{1D-CNN} & TR/TR & $0.678$ & -- & -- & $0.700$ & $0.515$ & $0.885$ & $0.719$\\
 & TS/TR & $0.695$ & $+0.017$ & $[-0.159, +0.214]$ & $0.691$ & $0.727$ & $0.654$ & $0.815$\\
 & TSR/TR & $0.814$ & $+0.136$ & $[-0.035, +0.300]$ & $0.797$ & $0.939$ & $0.654$ & $0.748$\\
\midrule
CbR & -- & $0.780$ & -- & $[+0.655, +0.893]$ & $0.770$ & $0.848$ & $0.692$ & $0.809$\\
\bottomrule
\end{tabular*}
\caption{\small \texttt{exp\_2.607}: IDH binary classification, $\beta$ = 0.25, $d$=32, reconstruction = MSE, $r_{S/R}=1$, \texttt{thr} = adaptive, top-90\%-by-rank per slide floored at $0.80$ (v2 preprocessing).}
\label{tab:ctrl_exp_2.607}
\end{table}

\texttt{exp\_2.608}: shared configuration, see its table above.

\begin{table}[H]
\centering
\scriptsize
\begin{tabular*}{\textwidth}{@{\extracolsep{\fill}}llccccccc}
\toprule
\emph{Model} & \emph{Regimen} & \emph{Accuracy} & \emph{$\Delta$} & \emph{95\% CI} & \emph{Bal.\ Acc} & \emph{Sensitivity} & \emph{Specificity} & \emph{AUC}\\
\midrule
\multirow{3}{*}{RF + SVM} & TR/TR & $0.746$ & -- & -- & $0.732$ & $0.848$ & $0.615$ & $0.726$\\
 & TS/TR & $0.525$ & $-0.220$* & $[-0.357, -0.078]$ & $0.543$ & $0.394$ & $0.692$ & $0.580$\\
 & TSR/TR & $0.814$ & $+0.068$* & $[+0.015, +0.140]$ & $0.801$ & $0.909$ & $0.692$ & $0.821$\\
\midrule
\multirow{3}{*}{RF} & TR/TR & $0.780$ & -- & -- & $0.762$ & $0.909$ & $0.615$ & $0.775$\\
 & TS/TR & $0.729$ & $-0.051$ & $[-0.160, +0.048]$ & $0.700$ & $0.939$ & $0.462$ & $0.758$\\
 & TSR/TR & $0.814$ & $+0.034$ & $[+0.000, +0.089]$ & $0.793$ & $0.970$ & $0.615$ & $0.790$\\
\midrule
\multirow{3}{*}{XGBoost} & TR/TR & $0.797$ & -- & -- & $0.777$ & $0.939$ & $0.615$ & $0.790$\\
 & TS/TR & $0.695$ & $-0.102$ & $[-0.207, +0.000]$ & $0.662$ & $0.939$ & $0.385$ & $0.768$\\
 & TSR/TR & $0.814$ & $+0.017$ & $[+0.000, +0.054]$ & $0.797$ & $0.939$ & $0.654$ & $0.782$\\
\midrule
\multirow{3}{*}{1D-CNN} & TR/TR & $0.610$ & -- & -- & $0.619$ & $0.545$ & $0.692$ & $0.652$\\
 & TS/TR & $0.475$ & $-0.136$ & $[-0.268, +0.000]$ & $0.510$ & $0.212$ & $0.808$ & $0.754$\\
 & TSR/TR & $0.695$ & $+0.085$ & $[-0.140, +0.328]$ & $0.670$ & $0.879$ & $0.462$ & $0.772$\\
\midrule
CbR & -- & $0.746$ & -- & $[+0.607, +0.879]$ & $0.748$ & $0.727$ & $0.769$ & $0.669$\\
\bottomrule
\end{tabular*}
\caption{\small \texttt{exp\_2.609}: IDH binary classification, $\beta$ = 0.75, $d$=32, reconstruction = MSE, $r_{S/R}=1$, \texttt{thr} = adaptive, top-90\%-by-rank per slide floored at $0.80$ (v2 preprocessing).}
\label{tab:ctrl_exp_2.609}
\end{table}

%% file: appendix_bootstrap_tables.tex
\begin{table}[H]
\centering
\scriptsize
\begin{tabular*}{\textwidth}{@{\extracolsep{\fill}}llcccc}
\toprule
\multirow{2}{*}{\emph{Model}} & \multirow{2}{*}{\emph{$d$}} & \multicolumn{2}{c}{\emph{TS/TR}} & \multicolumn{2}{c}{\emph{TSR/TR}} \\
\cmidrule(lr){3-4} \cmidrule(lr){5-6}
 & & \emph{$\Delta$} & \emph{95\% CI} & \emph{$\Delta$} & \emph{95\% CI} \\
\midrule
\multirow{2}{*}{RF + SVM} & $16 \to 32$ & $+0.068$ & $[-0.020, +0.177]$ & $+0.000$ & $[+0.000, +0.000]$ \\
 & $32 \to 64$ & $-0.085$ & $[-0.246, +0.071]$ & $+0.000$ & $[+0.000, +0.000]$ \\
\midrule
\multirow{2}{*}{RF} & $16 \to 32$ & $+0.034$ & $[-0.031, +0.107]$ & $+0.000$ & $[+0.000, +0.000]$ \\
 & $32 \to 64$ & $-0.034$ & $[-0.103, +0.033]$ & $+0.000$ & $[+0.000, +0.000]$ \\
\midrule
\multirow{2}{*}{XGBoost} & $16 \to 32$ & $+0.068$* & $[+0.017, +0.138]$ & $+0.000$ & $[+0.000, +0.000]$ \\
 & $32 \to 64$ & $-0.051$ & $[-0.113, +0.000]$ & $+0.000$ & $[+0.000, +0.000]$ \\
\midrule
\multirow{2}{*}{1D-CNN} & $16 \to 32$ & $-0.051$ & $[-0.170, +0.051]$ & $-0.034$ & $[-0.143, +0.061]$ \\
 & $32 \to 64$ & $+0.068$ & $[-0.016, +0.161]$ & $+0.051$ & $[-0.056, +0.164]$ \\
\bottomrule
\end{tabular*}
\caption{\small Pairwise bag-level bootstrap between CVAE latent dimensions ($\beta=0.50$, $d\in\{16,32,64\}$). $\Delta$ is the bag accuracy at the larger $d$ minus at the smaller $d$, with its 95\% CI from the class-stratified bootstrap over the 46 patients ($n=59$ paired bag decisions; 1,000 iterations, paired); (*) marks a CI excluding zero.}
\label{tab:ctrl_boot_latent}
\note{16 tests (4 models $\times$ 2 comparisons $\times$ 2 regimens), no multiplicity correction. A zero-width interval ($\Delta=0$, $[+0.000,+0.000]$) means the two models make identical decisions on every bag, thus the resampled difference is exactly zero; it is not a degenerate bootstrap.}
\end{table}

\begin{table}[H]
\centering
\scriptsize

\begin{tabular*}{\textwidth}{@{\extracolsep{\fill}}lccccc}
\toprule
\emph{Model} & \emph{40k} & \emph{80k} & \emph{$\Delta$} & \emph{95\% CI} & \emph{$p$} \\
\midrule
RF + SVM & $0.695$ & $0.712$ & $+0.017$ & $[-0.051, +0.098]$ & $0.790$ \\
RF & $0.831$ & $0.814$ & $-0.017$ & $[-0.075, +0.038]$ & $0.790$ \\
XGBoost & $0.831$ & $0.831$ & $+0.000$ & $[-0.048, +0.049]$ & $1.000$ \\
1D-CNN & $0.678$ & $0.780$ & $+0.102$ & $[+0.000, +0.228]$ & $0.112$ \\
\bottomrule
\end{tabular*}
\caption{\small Pairwise bag-level bootstrap between synthetic pools of 40,000 (\texttt{exp\_608}) and 80,000 (\texttt{exp\_708}) spectra per fold, TS/TR ($\beta=0.50$; same CVAE decoders, models and validation folds). $\Delta$ is the bag accuracy with the 80k pool minus with the 40k pool; 95\% CI as in Table~\ref{tab:ctrl_boot_latent}; (*) marks a CI excluding zero.}
\label{tab:ctrl_boot_pool}
\end{table}

%% file: results_mil_appendix_tables.tex
\subsubsection*{\texttt{exp\_607} ($\beta=0.25$, $\texttt{thr}=0.60$)}

\begin{table}[H]
\centering
\scriptsize
\captionsetup{position=bottom,font=small}
\begin{tabular*}{\textwidth}{@{\extracolsep{\fill}}lccccc}
\toprule
\emph{Model} & \emph{Accuracy} & \emph{Bal.\ Acc} & \emph{Sensitivity} & \emph{Specificity} & \emph{AUC} \\
\midrule
Naive MIL (RF) & $0.814$ & $0.797$ & $0.939$ & $0.654$ & $0.790$ \\
\midrule
mi-SVM & $0.763$ & $0.772$ & $0.697$ & $0.846$ & $0.773$ \\
Attention MIL & $0.525$ & $0.510$ & $0.636$ & $0.385$ & $0.566$ \\
CVAE MIL (reconstruction) & $0.729$ & $0.717$ & $0.818$ & $0.615$ & $0.828$ \\
CVAE MIL (encoder head) & $0.763$ & $0.751$ & $0.848$ & $0.654$ & $0.777$ \\
\bottomrule
\end{tabular*}
\caption{\small MIL tier bag-level results, \texttt{exp\_607} ($\beta=0.25$, $\texttt{thr}=0.6$). Binary IDH classification, $n=59$ slides, against the Naive MIL benchmark (a random forest trained on real data, every instance inheriting its bag's label, one majority-vote decision per bag).}
\label{tab:ctrl_mil_607}
\end{table}

\begin{table}[H]
\centering
\scriptsize
\captionsetup{position=bottom,font=small}
\setlength{\tabcolsep}{1.5pt}
\begin{tabular*}{\textwidth}{@{\extracolsep{\fill}}lcccccc}
\toprule
 & \multicolumn{3}{c}{\emph{Moran\textquotesingle{}s $I$}} & \multicolumn{3}{c}{\emph{Var-stand. semivariogram}} \\
\cmidrule(lr){2-4} \cmidrule(lr){5-7}
\emph{Model} & \emph{real} & \emph{shuf.} & \emph{$\Delta$ [95\% CI]} & \emph{real} & \emph{shuf.} & \emph{$\Delta$ [95\% CI]} \\
\midrule
mi-SVM & $0.005$ & $0.017$ & $-0.013$ $[-0.025,-0.004]$ & $0.009$ & $0.052$ & $-0.043$ $[-0.062,-0.025]$ \\
Attention MIL & $0.074$ & $0.065$ & $+0.009$ $[-0.007,+0.027]$ & $0.080$ & $0.071$ & $+0.009$ $[-0.007,+0.027]$ \\
CVAE MIL (reconstruction) & $0.144$ & $0.150$ & $-0.006$ $[-0.035,+0.023]$ & $0.269$ & $0.270$ & $-0.001$ $[-0.030,+0.027]$ \\
CVAE MIL (encoder head) & $0.056$ & $0.056$ & $-0.000$ $[-0.030,+0.025]$ & $0.071$ & $0.064$ & $+0.006$ $[-0.024,+0.033]$ \\
\bottomrule
\end{tabular*}
\caption{\small Spatial coherence of the MIL posterior maps, \texttt{exp\_607} ($\beta=0.25$, $\texttt{thr}=0.6$): Moran's $I$ and the variance-standardized semivariogram, reported as $1-\hat\gamma(3)/\hat\sigma^2$, at Chebyshev distance $\delta=3$ (higher is more coherent), real fit vs.\ shuffled-bag-label refit, averaged over the slides where both are defined. $\Delta$ is the mean per-slide difference (real $-$ shuffled) with its 95\% class-stratified bootstrap CI over patients; (*) marks a CI above zero. The real fit is more coherent than its control only if both statistics are starred.}
\label{tab:ctrl_spatial_607}
\end{table}

\begin{table}[H]
\centering
\scriptsize
\captionsetup{position=bottom,font=small}
\begin{tabular*}{\textwidth}{@{\extracolsep{\fill}}lcc}
\toprule
\emph{Model} & \emph{Accuracy} & \emph{$\Delta$ [95\% CI]} \\
\midrule
Naive MIL (RF) & $0.814$ & \\
\midrule
mi-SVM & $0.763$ & $-0.051$ $[-0.125,+0.017]$ \\
Attention MIL & $0.831$ & $+0.017$ $[+0.000,+0.056]$ \\
CVAE MIL (reconstruction) & $0.763$ & $-0.051$ $[-0.127,+0.018]$ \\
CVAE MIL (encoder head) & $0.797$ & $-0.017$ $[-0.056,+0.000]$ \\
\bottomrule
\end{tabular*}
\caption{\small Key-instance utility, \texttt{exp\_607} ($\beta=0.25$, $\texttt{thr}=0.6$): bag accuracy of a random forest trained only on the instances a tier selects (median split on non-diagnostic mass) and voting with the selected validation instances, against the Naive MIL benchmark. $\Delta$ is the paired per-slide accuracy difference with its 95\% CI from a class-stratified bootstrap over patients (10,000 resamples); (*) marks a CI above zero, i.e.\ the tier is better than the Naive benchmark.}
\label{tab:ctrl_keyinst_607}
\end{table}

\subsubsection*{\texttt{exp\_608} ($\beta=0.50$, $\texttt{thr}=0.6$)}

\begin{table}[H]
\centering
\scriptsize
\captionsetup{position=bottom,font=small}
\begin{tabular*}{\textwidth}{@{\extracolsep{\fill}}lccccc}
\toprule
\emph{Model} & \emph{Accuracy} & \emph{Bal.\ Acc} & \emph{Sensitivity} & \emph{Specificity} & \emph{AUC} \\
\midrule
Naive MIL (RF) & $0.814$ & $0.797$ & $0.939$ & $0.654$ & $0.790$ \\
\midrule
mi-SVM & $0.763$ & $0.772$ & $0.697$ & $0.846$ & $0.774$ \\
Attention MIL & $0.559$ & $0.529$ & $0.788$ & $0.269$ & $0.521$ \\
CVAE MIL (reconstruction) & $0.729$ & $0.709$ & $0.879$ & $0.538$ & $0.758$ \\
CVAE MIL (encoder head) & $0.797$ & $0.786$ & $0.879$ & $0.692$ & $0.784$ \\
\bottomrule
\end{tabular*}
\caption{\small MIL tier bag-level results, \texttt{exp\_608} ($\beta=0.50$, $\texttt{thr}=0.6$). Binary IDH classification, $n=59$ slides, against the Naive MIL benchmark (a random forest trained on real data, every instance inheriting its bag's label, one majority-vote decision per bag). }
\label{tab:ctrl_mil_608}
\end{table}

\begin{table}[H]
\centering
\scriptsize
\captionsetup{position=bottom,font=small}
\setlength{\tabcolsep}{1.5pt}
\begin{tabular*}{\textwidth}{@{\extracolsep{\fill}}lcccccc}
\toprule
 & \multicolumn{3}{c}{\emph{Moran\textquotesingle{}s $I$}} & \multicolumn{3}{c}{\emph{Var-stand. semivariogram}} \\
\cmidrule(lr){2-4} \cmidrule(lr){5-7}
\emph{Model} & \emph{real} & \emph{shuf.} & \emph{$\Delta$ [95\% CI]} & \emph{real} & \emph{shuf.} & \emph{$\Delta$ [95\% CI]} \\
\midrule
mi-SVM & $0.005$ & $0.017$ & $-0.013$ $[-0.025,-0.004]$ & $0.010$ & $0.052$ & $-0.042$ $[-0.061,-0.024]$ \\
Attention MIL & $0.054$ & $0.066$ & $-0.012$ $[-0.026,+0.001]$ & $0.069$ & $0.077$ & $-0.008$ $[-0.026,+0.008]$ \\
CVAE MIL (reconstruction) & $0.157$ & $0.171$ & $-0.014$ $[-0.042,+0.013]$ & $0.279$ & $0.285$ & $-0.006$ $[-0.039,+0.026]$ \\
CVAE MIL (encoder head) & $0.059$ & $0.057$ & $+0.003$ $[-0.027,+0.027]$ & $0.076$ & $0.066$ & $+0.010$ $[-0.022,+0.038]$ \\
\bottomrule
\end{tabular*}
\caption{\small Spatial coherence of the MIL posterior maps, \texttt{exp\_608} ($\beta=0.50$, $\texttt{thr}=0.6$): Moran's $I$ and the variance-standardized semivariogram, reported as $1-\hat\gamma(3)/\hat\sigma^2$, at Chebyshev distance $\delta=3$ (higher is more coherent), real fit vs.\ shuffled-bag-label refit, averaged over the slides where both are defined. $\Delta$ is the mean per-slide difference (real $-$ shuffled) with its 95\% class-stratified bootstrap CI over patients; (*) marks a CI above zero. The real fit is more coherent than its control only if both statistics are starred.}
\label{tab:ctrl_spatial_608}
\end{table}

\begin{table}[H]
\centering
\scriptsize
\captionsetup{position=bottom,font=small}
\begin{tabular*}{\textwidth}{@{\extracolsep{\fill}}lcc}
\toprule
\emph{Model} & \emph{Accuracy} & \emph{$\Delta$ [95\% CI]} \\
\midrule
Naive MIL (RF) & $0.814$ & \\
\midrule
mi-SVM & $0.763$ & $-0.051$ $[-0.125,+0.017]$ \\
Attention MIL & $0.797$ & $-0.017$ $[-0.091,+0.056]$ \\
CVAE MIL (reconstruction) & $0.763$ & $-0.051$ $[-0.117,+0.000]$ \\
CVAE MIL (encoder head) & $0.814$ & $+0.000$ $[+0.000,+0.000]$ \\
\bottomrule
\end{tabular*}
\caption{\small Key-instance utility, \texttt{exp\_608} ($\beta=0.50$, $\texttt{thr}=0.6$): bag accuracy of a random forest trained only on the instances a tier selects (median split on non-diagnostic mass) and voting with the selected validation instances, against the Naive MIL benchmark. $\Delta$ is the paired per-slide accuracy difference (tier $-$ Naive) with its 95\% CI from a class-stratified bootstrap over patients (10,000 resamples); (*) marks a CI above zero, i.e.\ the tier is better than the Naive benchmark.}
\label{tab:ctrl_keyinst_608}
\end{table}

\subsubsection*{\texttt{exp\_609} ($\beta=0.75$, $\texttt{thr}=0.6$)}

\begin{table}[H]
\centering
\scriptsize
\captionsetup{position=bottom,font=small}
\begin{tabular*}{\textwidth}{@{\extracolsep{\fill}}lccccc}
\toprule
\emph{Model} & \emph{Accuracy} & \emph{Bal.\ Acc} & \emph{Sensitivity} & \emph{Specificity} & \emph{AUC} \\
\midrule
Naive MIL (RF) & $0.814$ & $0.797$ & $0.939$ & $0.654$ & $0.790$ \\
\midrule
mi-SVM & $0.763$ & $0.772$ & $0.697$ & $0.846$ & $0.774$ \\
Attention MIL & $0.610$ & $0.578$ & $0.848$ & $0.308$ & $0.622$ \\
CVAE MIL (reconstruction) & $0.678$ & $0.667$ & $0.758$ & $0.577$ & $0.659$ \\
CVAE MIL (encoder head) & $0.780$ & $0.766$ & $0.879$ & $0.654$ & $0.772$ \\
\bottomrule
\end{tabular*}
\caption{\small MIL tier bag-level results, \texttt{exp\_609} ($\beta=0.75$, $\texttt{thr}=0.6$). Binary IDH classification, $n=59$ slides, against the Naive MIL benchmark (a random forest trained on real data, every instance inheriting its bag's label, one majority-vote decision per bag).}
\label{tab:ctrl_mil_609}
\end{table}

\begin{table}[H]
\centering
\scriptsize
\captionsetup{position=bottom,font=small}
\setlength{\tabcolsep}{1.5pt}
\begin{tabular*}{\textwidth}{@{\extracolsep{\fill}}lcccccc}
\toprule
 & \multicolumn{3}{c}{\emph{Moran\textquotesingle{}s $I$}} & \multicolumn{3}{c}{\emph{Var-stand. semivariogram}} \\
\cmidrule(lr){2-4} \cmidrule(lr){5-7}
\emph{Model} & \emph{real} & \emph{shuf.} & \emph{$\Delta$ [95\% CI]} & \emph{real} & \emph{shuf.} & \emph{$\Delta$ [95\% CI]} \\
\midrule
mi-SVM & $0.005$ & $0.017$ & $-0.013$ $[-0.025,-0.004]$ & $0.010$ & $0.052$ & $-0.042$ $[-0.061,-0.024]$ \\
Attention MIL & $0.043$ & $0.061$ & $-0.018$ $[-0.032,-0.006]$ & $0.051$ & $0.081$ & $-0.031$ $[-0.051,-0.012]$ \\
CVAE MIL (reconstruction) & $0.158$ & $0.164$ & $-0.006$ $[-0.037,+0.022]$ & $0.269$ & $0.252$ & $+0.017$ $[-0.018,+0.049]$ \\
CVAE MIL (encoder head) & $0.061$ & $0.058$ & $+0.003$ $[-0.026,+0.028]$ & $0.078$ & $0.065$ & $+0.012$ $[-0.018,+0.040]$ \\
\bottomrule
\end{tabular*}
\caption{\small Spatial coherence of the MIL posterior maps, \texttt{exp\_609} ($\beta=0.75$, $\texttt{thr}=0.6$): Moran's $I$ and the variance-standardized semivariogram, reported as $1-\hat\gamma(3)/\hat\sigma^2$, at Chebyshev distance $\delta=3$ (higher is more coherent), real fit vs.\ shuffled-bag-label refit, averaged over the slides where both are defined. $\Delta$ is the mean per-slide difference (real $-$ shuffled) with its 95\% class-stratified bootstrap CI over patients; (*) marks a CI above zero. The real fit is more coherent than its control only if both statistics are starred.}
\label{tab:ctrl_spatial_609}
\end{table}

\begin{table}[H]
\centering
\scriptsize
\captionsetup{position=bottom,font=small}
\begin{tabular*}{\textwidth}{@{\extracolsep{\fill}}lcc}
\toprule
\emph{Model} & \emph{Accuracy} & \emph{$\Delta$ [95\% CI]} \\
\midrule
Naive MIL (RF) & $0.814$ & \\
\midrule
mi-SVM & $0.763$ & $-0.051$ $[-0.125,+0.017]$ \\
Attention MIL & $0.780$ & $-0.034$ $[-0.089,+0.000]$ \\
CVAE MIL (reconstruction) & $0.746$ & $-0.068$ $[-0.137,-0.016]$ \\
CVAE MIL (encoder head) & $0.814$ & $+0.000$ $[+0.000,+0.000]$ \\
\bottomrule
\end{tabular*}
\caption{\small Key-instance utility, \texttt{exp\_609} ($\beta=0.75$, $\texttt{thr}=0.6$): bag accuracy of a random forest trained only on the instances a tier selects (median split on non-diagnostic mass) and voting with the selected validation instances, against the Naive MIL benchmark. $\Delta$ is the paired per-slide accuracy difference (tier $-$ Naive) with its 95\% CI from a class-stratified bootstrap over patients (10,000 resamples); (*) marks a CI above zero, i.e.\ the tier is better than the Naive benchmark.}
\label{tab:ctrl_keyinst_609}
\end{table}

\subsubsection*{\texttt{exp\_2.607} ($\beta=0.25$, $\texttt{thr}=0.8^*$)}
The threshold $\texttt{thr} = 0.8^*$ is an adaptive rule that retains the top 90\% of the slide with an additional condition of selecting spectra with maximum intensity of $0.80$.
\begin{table}[H]
\centering
\scriptsize
\captionsetup{position=bottom,font=small}
\begin{tabular*}{\textwidth}{@{\extracolsep{\fill}}lccccc}
\toprule
\emph{Model} & \emph{Accuracy} & \emph{Bal.\ Acc} & \emph{Sensitivity} & \emph{Specificity} & \emph{AUC} \\
\midrule
Naive MIL (RF) & $0.780$ & $0.762$ & $0.909$ & $0.615$ & $0.775$ \\
\midrule
mi-SVM & $0.627$ & $0.626$ & $0.636$ & $0.615$ & $0.691$ \\
Attention MIL & $0.593$ & $0.567$ & $0.788$ & $0.346$ & $0.624$ \\
CVAE MIL (reconstruction) & $0.746$ & $0.732$ & $0.848$ & $0.615$ & $0.807$ \\
CVAE MIL (encoder head) & $0.780$ & $0.770$ & $0.848$ & $0.692$ & $0.800$ \\
\bottomrule
\end{tabular*}
\caption{\small MIL tier bag-level results, \texttt{exp\_2.607} ($\beta=0.25$, $\texttt{thr}=0.8^*$). Binary IDH classification, $n=59$ slides, against the Naive MIL benchmark (a random forest trained on real data, every instance inheriting its bag's label, one majority-vote decision per bag).}
\label{tab:ctrl_mil_2607}
\end{table}

\begin{table}[H]
\centering
\scriptsize
\captionsetup{position=bottom,font=small}
\setlength{\tabcolsep}{1.5pt}
\begin{tabular*}{\textwidth}{@{\extracolsep{\fill}}lcccccc}
\toprule
 & \multicolumn{3}{c}{\emph{Moran\textquotesingle{}s $I$}} & \multicolumn{3}{c}{\emph{Var-stand. semivariogram}} \\
\cmidrule(lr){2-4} \cmidrule(lr){5-7}
\emph{Model} & \emph{real} & \emph{shuf.} & \emph{$\Delta$ [95\% CI]} & \emph{real} & \emph{shuf.} & \emph{$\Delta$ [95\% CI]} \\
\midrule
mi-SVM & $0.006$ & $0.012$ & $-0.006$ $[-0.011,-0.002]$ & $0.006$ & $0.042$ & $-0.035$ $[-0.057,-0.014]$ \\
Attention MIL & $0.058$ & $0.063$ & $-0.005$ $[-0.016,+0.007]$ & $0.062$ & $0.067$ & $-0.005$ $[-0.017,+0.008]$ \\
CVAE MIL (reconstruction) & $0.136$ & $0.135$ & $+0.001$ $[-0.024,+0.028]$ & $0.210$ & $0.226$ & $-0.016$ $[-0.043,+0.011]$ \\
CVAE MIL (encoder head) & $0.051$ & $0.041$ & $+0.010$ $[-0.016,+0.029]$ & $0.062$ & $0.048$ & $+0.014$ $[-0.013,+0.036]$ \\
\bottomrule
\end{tabular*}
\caption{\small Spatial coherence of the MIL posterior maps, \texttt{exp\_2.607} ($\beta=0.25$, $\texttt{thr}=0.8^*$): Moran's $I$ and the variance-standardized semivariogram, reported as $1-\hat\gamma(3)/\hat\sigma^2$, at Chebyshev distance $\delta=3$ (higher is more coherent), real fit vs.\ shuffled-bag-label refit, averaged over the slides where both are defined. $\Delta$ is the mean per-slide difference (real $-$ shuffled) with its 95\% class-stratified bootstrap CI over patients; (*) marks a CI above zero. The real fit is more coherent than its control only if both statistics are starred.}
\label{tab:ctrl_spatial_2607}
\end{table}

\begin{table}[H]
\centering
\scriptsize
\captionsetup{position=bottom,font=small}
\begin{tabular*}{\textwidth}{@{\extracolsep{\fill}}lcc}
\toprule
\emph{Model} & \emph{Accuracy} & \emph{$\Delta$ [95\% CI]} \\
\midrule
Naive MIL (RF) & $0.780$ & \\
\midrule
mi-SVM & $0.763$ & $-0.017$ $[-0.057,+0.000]$ \\
Attention MIL & $0.797$ & $+0.017$ $[+0.000,+0.056]$ \\
CVAE MIL (reconstruction) & $0.780$ & $+0.000$ $[-0.049,+0.050]$ \\
CVAE MIL (encoder head) & $0.831$ & $+0.051$ $[+0.000,+0.115]$ \\
\bottomrule
\end{tabular*}
\caption{\small Key-instance utility, \texttt{exp\_2.607} ($\beta=0.25$, $\texttt{thr}=0.8^*$): bag accuracy of a random forest trained only on the instances a tier selects (median split on non-diagnostic mass) and voting with the selected validation instances, against the Naive MIL benchmark. $\Delta$ is the paired per-slide accuracy difference (tier $-$ Naive) with its 95\% CI from a class-stratified bootstrap over patients (10,000 resamples); (*) marks a CI above zero, i.e.\ the tier is better than the Naive benchmark.}
\label{tab:ctrl_keyinst_2607}
\end{table}

\subsubsection*{\texttt{exp\_2.608} ($\beta=0.50$, $\texttt{thr}=0.8^*$)}

\begin{table}[H]
\centering
\scriptsize
\captionsetup{position=bottom,font=small}
\begin{tabular*}{\textwidth}{@{\extracolsep{\fill}}lccccc}
\toprule
\emph{Model} & \emph{Accuracy} & \emph{Bal.\ Acc} & \emph{Sensitivity} & \emph{Specificity} & \emph{AUC} \\
\midrule
Naive MIL (RF) & $0.780$ & $0.762$ & $0.909$ & $0.615$ & $0.775$ \\
\midrule
mi-SVM & $0.627$ & $0.626$ & $0.636$ & $0.615$ & $0.691$ \\
Attention MIL & $0.593$ & $0.551$ & $0.909$ & $0.192$ & $0.550$ \\
CVAE MIL (reconstruction) & $0.729$ & $0.721$ & $0.788$ & $0.654$ & $0.746$ \\
CVAE MIL (encoder head) & $0.797$ & $0.786$ & $0.879$ & $0.692$ & $0.775$ \\
\bottomrule
\end{tabular*}
\caption{\small MIL tier bag-level results, \texttt{exp\_2.608} ($\beta=0.50$, $\texttt{thr}=0.8^*$). Binary IDH classification, $n=59$ slides, against the Naive MIL benchmark (a random forest trained on real data, every instance inheriting its bag's label, one majority-vote decision per bag).}
\label{tab:ctrl_mil_2608}
\end{table}

\begin{table}[H]
\centering
\scriptsize
\captionsetup{position=bottom,font=small}
\setlength{\tabcolsep}{1.5pt}
\begin{tabular*}{\textwidth}{@{\extracolsep{\fill}}lcccccc}
\toprule
 & \multicolumn{3}{c}{\emph{Moran\textquotesingle{}s $I$}} & \multicolumn{3}{c}{\emph{Var-stand. semivariogram}} \\
\cmidrule(lr){2-4} \cmidrule(lr){5-7}
\emph{Model} & \emph{real} & \emph{shuf.} & \emph{$\Delta$ [95\% CI]} & \emph{real} & \emph{shuf.} & \emph{$\Delta$ [95\% CI]} \\
\midrule
mi-SVM & $0.006$ & $0.012$ & $-0.006$ $[-0.011,-0.002]$ & $0.006$ & $0.042$ & $-0.035$ $[-0.057,-0.014]$ \\
Attention MIL & $0.052$ & $0.065$ & $-0.012$ $[-0.028,+0.004]$ & $0.054$ & $0.083$ & $-0.029$ $[-0.046,-0.011]$ \\
CVAE MIL (reconstruction) & $0.152$ & $0.151$ & $+0.001$ $[-0.028,+0.026]$ & $0.239$ & $0.237$ & $+0.002$ $[-0.028,+0.030]$ \\
CVAE MIL (encoder head) & $0.050$ & $0.038$ & $+0.012$ $[-0.017,+0.036]$ & $0.063$ & $0.046$ & $+0.017$ $[-0.014,+0.043]$ \\
\bottomrule
\end{tabular*}
\caption{\small Spatial coherence of the MIL posterior maps, \texttt{exp\_2.608} ($\beta=0.50$, $\texttt{thr}=0.8^*$): Moran's $I$ and the variance-standardized semivariogram, reported as $1-\hat\gamma(3)/\hat\sigma^2$, at Chebyshev distance $\delta=3$ (higher is more coherent), real fit vs.\ shuffled-bag-label refit, averaged over the slides where both are defined. $\Delta$ is the mean per-slide difference (real $-$ shuffled) with its 95\% class-stratified bootstrap CI over patients; (*) marks a CI above zero. The real fit is more coherent than its control only if both statistics are starred.}
\label{tab:ctrl_spatial_2608}
\end{table}

\begin{table}[H]
\centering
\scriptsize
\captionsetup{position=bottom,font=small}
\begin{tabular*}{\textwidth}{@{\extracolsep{\fill}}lcc}
\toprule
\emph{Model} & \emph{Accuracy} & \emph{$\Delta$ [95\% CI]} \\
\midrule
Naive MIL (RF) & $0.780$ & \\
\midrule
mi-SVM & $0.763$ & $-0.017$ $[-0.057,+0.000]$ \\
Attention MIL & $0.797$ & $+0.017$ $[+0.000,+0.056]$ \\
CVAE MIL (reconstruction) & $0.797$ & $+0.017$ $[-0.036,+0.074]$ \\
CVAE MIL (encoder head) & $0.814$ & $+0.034$ $[+0.000,+0.088]$ \\
\bottomrule
\end{tabular*}
\caption{\small Key-instance utility, \texttt{exp\_2.608} ($\beta=0.50$, $\texttt{thr}=0.8^*$): bag accuracy of a random forest trained only on the instances a tier selects (median split on non-diagnostic mass) and voting with the selected validation instances, against the Naive MIL benchmark. $\Delta$ is the paired per-slide accuracy difference (tier $-$ Naive) with its 95\% CI from a class-stratified bootstrap over patients (10,000 resamples); (*) marks a CI above zero, i.e.\ the tier is better than the Naive benchmark.}
\label{tab:ctrl_keyinst_2608}
\end{table}

\subsubsection*{\texttt{exp\_4.608} ($\beta=0.50$, $\texttt{thr}=0.40$)}

\begin{table}[H]
\centering
\scriptsize
\captionsetup{position=bottom,font=small}
\begin{tabular*}{\textwidth}{@{\extracolsep{\fill}}lccccc}
\toprule
\emph{Model} & \emph{Accuracy} & \emph{Bal.\ Acc} & \emph{Sensitivity} & \emph{Specificity} & \emph{AUC} \\
\midrule
Naive MIL (RF) & $0.814$ & $0.797$ & $0.939$ & $0.654$ & $0.770$ \\
\midrule
mi-SVM & $0.627$ & $0.618$ & $0.697$ & $0.538$ & $0.667$ \\
Attention MIL & $0.610$ & $0.598$ & $0.697$ & $0.500$ & $0.562$ \\
CVAE MIL (reconstruction) & $0.712$ & $0.698$ & $0.818$ & $0.577$ & $0.798$ \\
CVAE MIL (encoder head) & $0.814$ & $0.801$ & $0.909$ & $0.692$ & $0.794$ \\
\bottomrule
\end{tabular*}
\caption{\small MIL tier bag-level results, \texttt{exp\_4.608} ($\beta=0.50$, $\text{thr}=0.40$). Binary IDH classification, $n=59$ slides, against the Naive MIL benchmark (a random forest trained on real data, every instance inheriting its bag's label, one majority-vote decision per bag).}
\label{tab:ctrl_mil_4608}
\end{table}

\begin{table}[H]
\centering
\scriptsize
\captionsetup{position=bottom,font=small}
\setlength{\tabcolsep}{1.5pt}
\begin{tabular*}{\textwidth}{@{\extracolsep{\fill}}lcccccc}
\toprule
 & \multicolumn{3}{c}{\emph{Moran\textquotesingle{}s $I$}} & \multicolumn{3}{c}{\emph{Var-stand. semivariogram}} \\
\cmidrule(lr){2-4} \cmidrule(lr){5-7}
\emph{Model} & \emph{real} & \emph{shuf.} & \emph{$\Delta$ [95\% CI]} & \emph{real} & \emph{shuf.} & \emph{$\Delta$ [95\% CI]} \\
\midrule
mi-SVM & $0.009$ & $0.013$ & $-0.004$ $[-0.013,+0.006]$ & $0.012$ & $0.041$ & $-0.029$ $[-0.049,-0.009]$ \\
Attention MIL & $0.074$ & $0.070$ & $+0.004$ $[-0.014,+0.022]$ & $0.098$ & $0.086$ & $+0.012$ $[-0.008,+0.031]$ \\
CVAE MIL (reconstruction) & $0.155$ & $0.184$ & $-0.029$ $[-0.059,-0.000]$ & $0.292$ & $0.298$ & $-0.006$ $[-0.037,+0.024]$ \\
CVAE MIL (encoder head) & $0.066$ & $0.060$ & $+0.006$ $[-0.024,+0.031]$ & $0.086$ & $0.068$ & $+0.018$ $[-0.012,+0.044]$ \\
\bottomrule
\end{tabular*}
\caption{\small Spatial coherence of the MIL posterior maps, \texttt{exp\_4.608} ($\beta=0.50$, $\text{thr}=0.40$): Moran's $I$ and the variance-standardized semivariogram, reported as $1-\hat\gamma(3)/\hat\sigma^2$, at Chebyshev distance $\delta=3$ (higher is more coherent), real fit vs.\ shuffled-bag-label refit, averaged over the slides where both are defined. $\Delta$ is the mean per-slide difference (real $-$ shuffled) with its 95\% class-stratified bootstrap CI over patients; (*) marks a CI above zero. The real fit is more coherent than its control only if both statistics are starred.}
\label{tab:ctrl_spatial_4608}
\end{table}

\begin{table}[H]
\centering
\scriptsize
\captionsetup{position=bottom,font=small}
\begin{tabular*}{\textwidth}{@{\extracolsep{\fill}}lcc}
\toprule
\emph{Model} & \emph{Accuracy} & \emph{$\Delta$ [95\% CI]} \\
\midrule
Naive MIL (RF) & $0.814$ & \\
\midrule
mi-SVM & $0.780$ & $-0.034$ $[-0.089,+0.000]$ \\
Attention MIL & $0.780$ & $-0.034$ $[-0.089,+0.000]$ \\
CVAE MIL (reconstruction) & $0.780$ & $-0.034$ $[-0.105,+0.030]$ \\
CVAE MIL (encoder head) & $0.831$ & $+0.017$ $[+0.000,+0.057]$ \\
\bottomrule
\end{tabular*}
\caption{\small Key-instance utility, \texttt{exp\_4.608} ($\beta=0.50$, $\text{thr}=0.40$): bag accuracy of a random forest trained only on the instances a tier selects (median split on non-diagnostic mass) and voting with the selected validation instances, against the Naive MIL benchmark. $\Delta$ is the paired per-slide accuracy difference (tier $-$ Naive) with its 95\% CI from a class-stratified bootstrap over patients (10,000 resamples); (*) marks a CI above zero, i.e.\ the tier is better than the Naive benchmark.}
\label{tab:ctrl_keyinst_4608}
\end{table}

%% file: appendix_mil.tex
\section{Multiple Instance Learning}
\label{app:supporting_methods}
\setcounter{equation}{0}
\renewcommand{\theequation}{\Alph{section}.\arabic{equation}}

This appendix documents the Multiple Instance Learning (MIL) paradigm. We compare four MIL classification models, which we describe below. Moreover, we present in detail the evaluation criteria for the MIL models. 

\subsection{MIL formulations}\label{app:mil_formulations}

Applying the general MIL formulation to our case, we treat each biopsy slide as a bag of Raman spectra with one IDH label, since diagnostic labels for individual spectra were unavailable. For the set of slides $\mathcal{S}$, we consider a slide $b\in\mathcal{S}$ with $n_b$ preprocessed spectra and label $y_b$ as:
\begin{equation}
    b=\bigl((x_j)_{j=1}^{n_b},y_b\bigr),\qquad x_j\in\mathbb{R}^{N_\lambda}.
\end{equation}

In these formulations, we define the biopsy labels as $y_b\in\{1,2\}$ for binary classification ($1$ for IDH-Mutant and $2$ for IDH-Wildtype) and $y_b\in\{1,\ldots,6\}$ for the six-profile problem.

We define $z_j\in\{0,1\}$ as the unknown diagnostic status of spectrum $x_j$, with $1$ denoting diagnostic tissue. The true fraction of diagnostic spectra on slide $b$ is
\begin{equation}
    \pi_b=\frac{1}{n_b}\sum_{j=1}^{n_b}\mathbbm{1}_{[z_j=1]}.
\end{equation}
Because neither $z_j$ nor $\pi_b$ was observed, the models selected candidate key instances from their predictions without establishing whether these spectra came from diagnostic tissue. We used $\pi_{\mathrm{min}}=0.02$ as a minimum selection fraction to avoid empty selections, not as an estimate of the true diagnostic fraction.

We compared four MIL formulations: Naive MIL, mi-SVM, Attention MIL and CVAE MIL.

\subsubsection*{Naive MIL}
The naive benchmark assigned the slide label to every spectrum, which allowed us to train an instance classifier $\mathrm{cls}(\cdot)$ without instance annotations. At inference, the slide received the majority label of its instance predictions:
\begin{equation}
    \mathrm{naiveMIL}(b)=\arg\max_c\sum_{j=1}^{n_b}\mathbbm{1}_{[\mathrm{cls}(x_j)=c]}.
\end{equation}
This method used all spectra for training and voting without estimating which spectra were diagnostic.

\subsubsection*{mi-SVM}
The mi-SVM formulation updated instance labels iteratively, using classifier scores to select the spectra that retained each bag's class label. We used a variant of mi-SVM~\cite{andrews2002misvm}, with three changes (Algorithm~\ref{alg:misvm}):
\begin{enumerate}[label=(\roman*)]
    \item The base classifier was RF+SVM.
 \item The fraction selected for relabelling was $\pi=0.02$, with at least one instance selected per bag. This fraction is distinct from the minimum selection fraction $\pi_{\mathrm{min}}$.
    \item We sampled $500$ instances per bag for training, because fitting the SVM on all spectra is computationally expensive.
\end{enumerate}
We used at most $10$ iterations.

\begin{algorithm}[H]
\caption{mi-SVM}
\label{alg:misvm}
\begin{algorithmic}[1]
\Require
    $\{x_i\}_{i=1}^n$ set of all instances;
    $\mathcal{S}$ set of all bags;
    $B(i)$ bag containing instance $i$;
    $\{y_b\}_{b\in\mathcal{S}}$ bag labels;
    $\pi$ fraction selected for relabelling;
    \texttt{max\_iter} maximum number of iterations;
    base learner $\mathrm{cls}(\cdot)$
\Ensure normalized instance scores $P$
\For{$k\in\{1,\ldots,C\}$}
\State $\hat y_i\gets\mathbbm{1}_{[y_{B(i)}=k]}$ for all $i$ \Comment{inherit the bag label}
\Do
    \State $f_k\gets\text{fit cls on }(x_i,\hat y_i)_{i=1}^n$
    \State $p_{i,k}\gets f_k(x_i)$ for all $i$ \Comment{values in $[0,1]$}
    \For{each bag $b$ with $y_b=k$}
        \State $\mathrm{Top}_b\gets$ indices of the top $\pi$ fraction of $p_{i,k}$ values in $b$
        \State $\hat y_i\gets\mathbbm{1}_{[i\in\mathrm{Top}_b]}$ for instances in $b$
    \EndFor
\While{ labels have changed and iteration $<$ \texttt{max\_iter}}
\EndFor
\State \label{mi-svm:non-diagnostic} $p_{i,0}\gets1-\max_k p_{i,k}$ for all $i$ \Comment{non-diagnostic score}
\State \label{mi-svm:norm} $P_{i,:}\gets(p_{i,0},p_{i,1},\ldots,p_{i,C})\,/\,\sum_{k=0}^C p_{i,k}$
\State \Return $P$
\end{algorithmic}
\end{algorithm}

In lines~\ref{mi-svm:non-diagnostic}--\ref{mi-svm:norm}, we added a class-$0$ score to the $C$ class scores and normalized the vector. We called this component the non-diagnostic score, although it was derived from confidence in the slide classes and did not represent a learned probability of non-diagnostic tissue.

\subsubsection*{Attention MIL}
Attention MIL assigned a weight to each instance such that spectra could contribute differently to the slide prediction. We attached a gated attention head~\cite{ilse2018attention} to the frozen encoder's latent mean vector $\mu_\phi(x)$ and also used the CVAE's instance predictions. For training, we sampled batches of size 228 of spectra (instances) in each slide (bag).

We trained on subsets of spectra, called sub-bags, because full bags were too large to fit in memory. We used layer normalization and a fully connected instance head with L2 regularization and dropout $0.30$. The latent vectors were centered per bag, during training and testing as well. Label smoothing was applied with $\epsilon = 0.01$. Attention weights and instance scores are model outputs, not ground-truth annotations of diagnostic tissue.

\subsubsection*{CVAE MIL}
The CVAE formulation reused the trained generative model to score individual spectra for MIL, with two alternatives based on classification and reconstruction:
\begin{enumerate}
    \item The encoder's classification head gave a softmax over the $C$ slide classes (Section~\ref{sec:cvae_architecture}).
 \item We converted the class-specific CbR reconstruction errors (Section~\ref{sec:cbr}) to scores with $\mathrm{softmax}(-\mathrm{MSE/T_{}})$, where $T$ is a temperature parameter set here to $T = 0.001$.
\end{enumerate}
We added the class-$0$ component using the same convention as in Algorithm~\ref{alg:misvm}. For a binary softmax,
\begin{equation}
    p_{i,0}=1-\max(p_{i,1},p_{i,2})
    =\min(p_{i,1},p_{i,2})\leq\max(p_{i,1},p_{i,2}).
\end{equation}
Normalization preserves this ordering, thus the added component cannot be the unique largest score. It does not define a separate tissue class through an argmax decision. We used it to rank and select instances, as described below.

\subsection{Evaluation of MIL models}\label{app:mil_evaluation}

We evaluated slide classification and instance selection separately, using bag-level accuracy, balanced accuracy, sensitivity, specificity and AUC for classification. Because annotations of diagnostic regions were unavailable, we evaluated instance selection indirectly through spatial coherence and key-instance utility.

\subsubsection*{Spatial coherence}
To test spatial correlation of the model's selection outputs, we used the pixel coordinates $p_j=(h_j,w_j)$ of each spectrum $x_j$ and defined pairs at fixed Chebyshev distance $\delta$ as
\begin{equation}
    P_\delta=\{(i,j):\lVert p_i-p_j\rVert_\infty=\delta\}.
\end{equation}
Let $\hat z_j$ denote the model-derived selection value at pixel $p_j$, and let $\bar z$ be its mean over the slide. This output is distinct from the unknown tissue label $z_j$ in the MIL formulation above (Section~\ref{app:mil_formulations}). We computed Moran's $I$~\cite{moran1950,rittel2026cohort}:
\begin{equation}
    I_\delta=\frac{n_b}{|P_\delta|}
    \frac{\sum_{(i,j)\in P_\delta}(\hat z_i-\bar z)(\hat z_j-\bar z)}
    {\sum_{i=1}^{n_b}(\hat z_i-\bar z)^2}.
\end{equation}
We also computed the empirical semivariogram~\cite{veronesi2025machine}:
\begin{equation}
    \hat\gamma(\delta)=\frac{1}{2|P_\delta|}
    \sum_{(i,j)\in P_\delta}(\hat z_i-\hat z_j)^2.
\end{equation}
We used $\delta=3$, taking the 3$\times$3 median filter into account. The results reported the variance-standardized form: $1-\hat\gamma(3)/\hat\sigma^2$, where $\hat\sigma^2$ is the variance of the selection values on the slide. A larger value indicates greater spatial coherence at the chosen distance.

For each model, we computed paired per-slide differences in spatial statistics between a fit with true bag labels and a refit with shuffled labels. A model passed this test only if the 95\% CIs were above zero for both statistics (Table~\ref{tab:mil_tests_status}). Spatial coherence is an indirect criterion because tissue structure and preprocessing can produce it even when selection is unrelated to IDH status.

\subsubsection*{Key-instance utility}
We trained RF on instances selected by each MIL model using a class-$0$ score below the median of the training fold (Appendix~\ref{app:controlled_experiments}). Each held-out slide was classified by majority vote over its selected instances. We compared this result with RF trained on all instances under Naive MIL and with RF trained on a size-matched random subset. This test evaluated selection during both training and slide classification (Table~\ref{tab:mil_tests_keyinst}).

An improvement over the random subset would support the predictive utility of selection without establishing that the selected spectra came from diagnostic tissue. Neither test replaces validation against histopathological annotations.

%% file: references.bib
@article{gbm_review_2025,
  author = {Królikowska, K. and Błaszczak, K. and Ławicki, S. and Zajkowska, M. and Gudowska-Sawczuk, M.},
  title = {Glioblastoma -- A Contemporary Overview of Epidemiology, Classification, Pathogenesis, Diagnosis, and Treatment: A Review Article},
  journal = {International Journal of Molecular Sciences},
  volume = {26},
  number = {24},
  year = {2025},
  publisher = {MDPI AG},
  doi = {10.3390/ijms262412162}
}

@article{stupak2025raman,
  author = {Stupak, E. V. and Glotov, V. M. and Askandaryan, A. S. and Clancy, S. E. and Hiana, J. C. and Cherkasova, O. P. and Stupak, V. V.},
  title = {Raman Spectroscopy in the Diagnosis of Brain Gliomas: A Literature Review},
  journal = {Cureus},
  volume = {17},
  number = {2},
  year = {2025},
  doi = {10.7759/cureus.79165},
  pmid = {40109807},
  pmcid = {PMC11921993}
}

@article{berisha2021overfitting,
  author = {Visar Berisha and Chelsea Krantsevich and P. Richard Hahn and Shira Hahn and Gautam Dasarathy and Pavan Turaga and Julie Liss},
  title = {Digital medicine and the curse of dimensionality},
  journal = {npj Digital Medicine},
  volume = {4},
  number = {1},
  pages = {153},
  year = {2021},
  publisher = {Nature Publishing Group},
  doi = {10.1038/s41746-021-00521-5}
}

@article{subramanian2013overfitting,
  author = {Jyothi Subramanian and Richard Simon},
  title = {Overfitting in prediction models -- is it a problem only in high dimensions?},
  journal = {Contemporary Clinical Trials},
  volume = {36},
  number = {2},
  pages = {636-641},
  year = {2013},
  publisher = {Elsevier},
  issn = {1551-7144},
  doi = {https://doi.org/10.1016/j.cct.2013.06.011}
}

@article{wu2021raman,
  author = {Wu, M. and Wang, S. and Pan, S. and Terentis, A. C. and Strasswimmer, J. and Zhu, X.},
  title = {Deep learning data augmentation for Raman spectroscopy cancer tissue classification},
  journal = {Scientific Reports},
  volume = {11},
  number = {1},
  pages = {23842},
  year = {2021},
  doi = {10.1038/s41598-021-02687-0},
  pmid = {34903743},
  pmcid = {PMC8668947}
}

@article{zhao2024raman,
  author = {Zhao, J. and Lui, H. and Kalia, S. and Lee, T. K. and Zeng, H.},
  title = {Improving skin cancer detection by Raman spectroscopy using convolutional neural networks and data augmentation},
  journal = {Frontiers in Oncology},
  volume = {14},
  pages = {1320220},
  year = {2024},
  doi = {10.3389/fonc.2024.1320220},
  pmid = {38962264},
  pmcid = {PMC11219827}
}

@article{jordon2022synthetic,
  author = {Jordon, James and Szpruch, Lukasz and Houssiau, Florimond and Bottarelli, Mirko and Cherubin, Giovanni and Maple, Carsten and Cohen, Samuel N. and Weller, Adrian},
  title = {Synthetic Data -- what, why and how?},
  year = {2022},
  publisher = {arXiv},
  doi = {10.48550/ARXIV.2205.03257},
  copyright = {arXiv.org perpetual, non-exclusive license}
}

@misc{esteban2017real,
  title={Real-valued (Medical) Time Series Generation with Recurrent Conditional GANs}, 
  author={Cristóbal Esteban and Stephanie L. Hyland and Gunnar Rätsch},
  year={2017}, 
  eprint={1706.02633},
  archivePrefix={arXiv},
  primaryClass={stat.ML},
  url={https://arxiv.org/abs/1706.02633}, 
}

@inproceedings{higgins2017betavae,
  author = {Higgins, Irina and Matthey, Loic and Pal, Arka and Burgess, Christopher and Glorot, Xavier and Botvinick, Matthew and Mohamed, Shakir and Lerchner, Alexander},
  title = {beta-{VAE}: Learning Basic Visual Concepts with a Constrained Variational Framework},
  booktitle = {International Conference on Learning Representations},
  year = {2017}
}

@article{Asperti2020,
  author = {Asperti, Andrea and Trentin, Matteo},
  title = {Balancing reconstruction error and Kullback-Leibler divergence in variational autoencoders},
  journal = {IEEE Access},
  volume = {8},
  pages = {199440--199448},
  year = {2020},
  publisher = {IEEE},
  doi = {https://dx.doi.org/10.1109/ACCESS.2020.3034828}
}

@article{Ceccarelli2016,
  author = {Ceccarelli, Michele and Barthel, Floris P. and Malta, Tathiane M. and Sabedot, Thais S. and Salama, Sofie R. and Murray, Bradley A. and Morozova, Olena and Newton, Yulia and Radenbaugh, Amie and Pagnotta, Stefano M. and Anjum, Samreen and Wang, Jiguang and Manyam, Ganiraju and Zoppoli, Pietro and Ling, Shiyun and Rao, Arjun A. and Grifford, Mia and Cherniack, Andrew D. and Zhang, Hailei and Poisson, Laila and Carlotti, Carlos Gilberto and Tirapelli, Daniela Pretti da Cunha and Rao, Arvind and Mikkelsen, Tom and Lau, Ching C. and Yung, W.K. Alfred and Rabadan, Raul and Huse, Jason and Brat, Daniel J. and Lehman, Norman L. and Barnholtz-Sloan, Jill S. and Zheng, Siyuan and Hess, Kenneth and Rao, Ganesh and Meyerson, Matthew and Beroukhim, Rameen and Cooper, Lee and Akbani, Rehan and Wrensch, Margaret and Haussler, David and Aldape, Kenneth D. and Laird, Peter W. and Gutmann, David H. and Noushmehr, Houtan and Iavarone, Antonio and Verhaak, Roel G.W. and Anjum, Samreen and Arachchi, Harindra and Auman, J. Todd and Balasundaram, Miruna and Balu, Saianand and Barnett, Gene and Baylin, Stephen and Bell, Sue and Benz, Christopher and Bir, Natalie and Black, Keith L. and Bodenheimer, Tom and Boice, Lori and Bootwalla, Moiz S. and Bowen, Jay and Bristow, Christopher A. and Butterfield, Yaron S.N. and Chen, Qing-Rong and Chin, Lynda and Cho, Juok and Chuah, Eric and Chudamani, Sudha and Coetzee, Simon G. and Cohen, Mark L. and Colman, Howard and Couce, Marta and D'Angelo, Fulvio and Davidsen, Tanja and Davis, Amy and Demchok, John A. and Devine, Karen and Ding, Li and Duell, Rebecca and Elder, J. Bradley and Eschbacher, Jennifer M. and Fehrenbach, Ashley and Ferguson, Martin and Frazer, Scott and Fuller, Gregory and Fulop, Jordonna and Gabriel, Stacey B. and Garofano, Luciano and Gastier-Foster, Julie M. and Gehlenborg, Nils and Gerken, Mark and Getz, Gad and Giannini, Caterina and Gibson, William J. and Hadjipanayis, Angela and Hayes, D. Neil and Heiman, David I. and Hermes, Beth and Hilty, Joe and Hoadley, Katherine A. and Hoyle, Alan P. and Huang, Mei and Jefferys, Stuart R. and Jones, Corbin D. and Jones, Steven J.M. and Ju, Zhenlin and Kastl, Alison and Kendler, Ady and Kim, Jaegil and Kucherlapati, Raju and Lai, Phillip H. and Lawrence, Michael S. and Lee, Semin and Leraas, Kristen M. and Lichtenberg, Tara M. and Lin, Pei and Liu, Yuexin and Liu, Jia and Ljubimova, Julia Y. and Lu, Yiling and Ma, Yussanne and Maglinte, Dennis T. and Mahadeshwar, Harshad S. and Marra, Marco A. and McGraw, Mary and McPherson, Christopher and Meng, Shaowu and Mieczkowski, Piotr A. and Miller, C. Ryan and Mills, Gordon B. and Moore, Richard A. and Mose, Lisle E. and Mungall, Andrew J. and Naresh, Rashi and Naska, Theresa and Neder, Luciano and Noble, Michael S. and Noss, Ardene and O'Neill, Brian Patrick and Ostrom, Quinn T. and Palmer, Cheryl and Pantazi, Angeliki and Parfenov, Michael and Park, Peter J. and Parker, Joel S. and Perou, Charles M. and Pierson, Christopher R. and Pihl, Todd and Protopopov, Alexei and Radenbaugh, Amie and Ramirez, Nilsa C. and Rathmell, W. Kimryn and Ren, Xiaojia and Roach, Jeffrey and Robertson, A. Gordon and Saksena, Gordon and Schein, Jacqueline E. and Schumacher, Steven E. and Seidman, Jonathan and Senecal, Kelly and Seth, Sahil and Shen, Hui and Shi, Yan and Shih, Juliann and Shimmel, Kristen and Sicotte, Hugues and Sifri, Suzanne and Silva, Tiago and Simons, Janae V. and Singh, Rosy and Skelly, Tara and Sloan, Andrew E. and Sofia, Heidi J. and Soloway, Matthew G. and Song, Xingzhi and Sougnez, Carrie and Souza, Camila and Staugaitis, Susan M. and Sun, Huandong and Sun, Charlie and Tan, Donghui and Tang, Jiabin and Tang, Yufang and Thorne, Leigh and Trevisan, Felipe Amstalden and Triche, Timothy and Van Den Berg, David J. and Veluvolu, Umadevi and Voet, Doug and Wan, Yunhu and Wang, Zhining and Warnick, Ronald and Weinstein, John N. and Weisenberger, Daniel J. and Wilkerson, Matthew D. and Williams, Felicia and Wise, Lisa and Wolinsky, Yingli and Wu, Junyuan and Xu, Andrew W. and Yang, Lixing and Yang, Liming and Zack, Travis I. and Zenklusen, Jean C. and Zhang, Jianhua and Zhang, Wei and Zhang, Jiashan and Zmuda, Erik},
  title = {Molecular Profiling Reveals Biologically Discrete Subsets and Pathways of Progression in Diffuse Glioma},
  journal = {Cell},
  volume = {164},
  number = {3},
  pages = {550–563},
  year = {2016},
  month = {01},
  publisher = {Elsevier BV},
  issn = {0092-8674},
  doi = {10.1016/j.cell.2015.12.028}
}

@article{apollonov2024raman,
  author = {Lita, Adrian and Sjöberg, Joel and Păcioianu, David and Siminea, Nicoleta and Celiku, Orieta and Dowdy, Tyrone and Paun, Andrei and Gilbert, Mark and Noushmehr, Houtan and Petre, Ion and Larion, Mioara},
  title = {Raman-based machine learning platform reveals unique metabolic differences between IDHmut and IDHwt glioma},
  journal = {Neuro-oncology},
  volume = {26},
  pages = {},
  year = {2024},
  month = {06},
  doi = {10.1093/neuonc/noae101}
}

@article{krafft2009raman,
  author = {Krafft, Christoph and Steiner, Gerwin and Salzer, Reiner},
  title = {Biomedical applications of Raman and infrared spectroscopy to diagnose tissues},
  journal = {Journal of Physics D: Applied Physics},
  volume = {42},
  number = {18},
  pages = {183001},
  year = {2009},
  doi = {10.1088/0022-3727/42/18/183001}
}

@article{bocklitz2016raman,
  author = {Bocklitz, Thomas and Salah, Faris S. and Vogler, Nadine and Heuke, Sarah and Chernavskaia, Olga and Schmidt, Christina and Waldner, Michael and Greten, Florian R. and Br{\"a}utigam, Katrin and Schmitt, Michael and Popp, J{\"u}rgen},
  title = {Raman based molecular imaging and analytics: a magic bullet for biomedical applications?},
  journal = {Analyst},
  volume = {141},
  number = {2},
  pages = {505--515},
  year = {2016},
  doi = {10.1039/C5AN01806G}
}

@article{Harris2023,
  author = {Harris, Georgia and Stickland, Clarissa A. and Lim, Matthias and Goldberg Oppenheimer, Pola},
  title = {Raman Spectroscopy Spectral Fingerprints of Biomarkers of Traumatic Brain Injury},
  journal = {Cells},
  volume = {12},
  number = {22},
  pages = {2589},
  year = {2023},
  month = {11},
  publisher = {MDPI AG},
  issn = {2073-4409},
  doi = {10.3390/cells12222589},
  url = {http://dx.doi.org/10.3390/cells12222589}
}

@inproceedings{Kizilov2025,
  author = {Kizilov, Mykyta and Cheburkanov, Vsevolod and Harrington, Joseph and Yakovlev, Vladislav},
  title = {Advanced preprocessing and analysis techniques for enhanced Raman spectroscopy data interpretation},
  booktitle = {Optical Biopsy XXIII: Toward Real-Time Spectroscopic Imaging and Diagnosis},
  editor = {Alfano, Robert R. and Seddon, Angela B. and Shi, Lingyan and Wu, Binlin},
  publisher = {SPIE},
  year = {2025},
  month = {03},
  pages = {51},
  doi = {10.1117/12.3048830},
  url = {http://dx.doi.org/10.1117/12.3048830}
}

@inproceedings{ilse2018attention,
  author = {Ilse, Maximilian and Tomczak, Jakub and Welling, Max},
  title = {Attention-based deep multiple instance learning},
  booktitle = {International conference on machine learning},
  organization = {Pmlr},
  year = {2018},
  pages = {2127--2136}
}

@article{craven1978smoothing,
  author = {Craven, Peter and Wahba, Grace},
  title = {Smoothing noisy data with spline functions: estimating the correct degree of smoothing by the method of generalized cross-validation},
  journal = {Numerische mathematik},
  volume = {31},
  number = {4},
  pages = {377--403},
  year = {1978},
  publisher = {Springer}
}

@article{hansen1992analysis,
  author = {Hansen, Per Christian},
  title = {Analysis of discrete ill-posed problems by means of the L-curve},
  journal = {SIAM review},
  volume = {34},
  number = {4},
  pages = {561--580},
  year = {1992},
  publisher = {SIAM}
}

@article{ye2020baseline,
  author = {Ye, Jianfeng and Tian, Ziyang and Wei, Haoyun and Li, Yan},
  title = {Baseline correction method based on improved asymmetrically reweighted penalized least squares for the Raman spectrum},
  journal = {Applied Optics},
  volume = {59},
  number = {34},
  pages = {10933--10943},
  year = {2020},
  publisher = {Optical Society of America}
}

@article{Kizilov2026,
  title={Modular and automated workflow for streamlined Raman signal analysis},
  author={Kizilov, Mykyta and Cheburkanov, Vsevolod and Harrington, Joseph and Yakovlev, Vladislav V},
  journal={Journal of Raman Spectroscopy},
  volume={57},
  number={2},
  pages={360--377},
  year={2026},
  publisher={Wiley Online Library}
}

@article{waqas2024exploring,
  title={Exploring multiple instance learning (MIL): A brief survey},
  author={Waqas, Muhammad and Ahmed, Syed Umaid and Tahir, Muhammad Atif and Wu, Jia and Qureshi, Rizwan},
  journal={Expert Systems with Applications},
  volume={250},
  pages={123893},
  year={2024},
  publisher={Elsevier}
}

@article{dietterich1997solving,
  title={Solving the multiple instance problem with axis-parallel rectangles},
  author={Dietterich, Thomas G and Lathrop, Richard H and Lozano-P{\'e}rez, Tom{\'a}s},
  journal={Artificial intelligence},
  volume={89},
  number={1-2},
  pages={31--71},
  year={1997},
  publisher={Elsevier}
}

@article{chawla2002smote,
  title={SMOTE: synthetic minority over-sampling technique},
  author={Chawla, Nitesh V and Bowyer, Kevin W and Hall, Lawrence O and Kegelmeyer, W Philip},
  journal={Journal of artificial intelligence research},
  volume={16},
  pages={321--357},
  year={2002}
}

@article{song2022rapid,
  title={Rapid identification of papillary thyroid carcinoma and papillary microcarcinoma based on serum Raman spectroscopy combined with machine learning models},
  author={Song, Haitao and Dong, Chao and Zhang, Xudan and Wu, Wei and Chen, Cheng and Ma, Binlin and Chen, Fangfang and Chen, Chen and Lv, Xiaoyi},
  journal={Photodiagnosis and Photodynamic Therapy},
  volume={37},
  pages={102647},
  year={2022},
  publisher={Elsevier}
}

@article{zhou2022deeply,
    author = {Zhou, Wei and Tang, Yujun and Qian, Ziheng and Wang, Junwei and Guo, Hanming},
    title = {Deeply-recursive convolutional neural network for Raman spectra identification},
    journal = {RSC Advances},
    volume = {12},
    number = {8},
    pages = {5053-5061},
    year = {2022},
    month = {02},
    issn = {2046-2069},
    doi = {10.1039/d1ra08804a},
    url = {https://doi.org/10.1039/d1ra08804a},
    eprint = {https://pubs.rsc.org/ra/article-pdf/12/8/5053/8137654/d1ra08804a.pdf},
}

@article{kiyasseh2020pleth,
  author={Kiyasseh, Dani and Tadesse, Girmaw Abebe and Nhan, Le Nguyen Thanh and Van Tan, Le and Thwaites, Louise and Zhu, Tingting and Clifton, David},
  journal={IEEE Journal of Biomedical and Health Informatics}, 
  title={PlethAugment: GAN-Based PPG Augmentation for Medical Diagnosis in Low-Resource Settings}, 
  year={2020},
  volume={24},
  number={11},
  pages={3226-3235},
  doi={10.1109/JBHI.2020.2979608}}

@article{kingma2013auto,
  title={Auto-encoding variational bayes},
  author={Kingma, Diederik P and Welling, Max},
  journal={arXiv preprint arXiv:1312.6114},
  year={2013}
}

@article{sohn2015learning,
  title={Learning structured output representation using deep conditional generative models},
  author={Sohn, Kihyuk and Lee, Honglak and Yan, Xinchen},
  journal={Advances in neural information processing systems},
  volume={28},
  year={2015}
}

@article{flanagan2025exploring,
  title={Exploring generative artificial intelligence and data augmentation techniques for spectroscopy analysis},
  author={Flanagan, Aaron R and Dalal, Dhairya and Glavin, Frank G},
  journal={Chemical Reviews},
  volume={125},
  number={13},
  pages={6130},
  year={2025}
}

@article{louis2021who,
    author = {Louis, David N and Perry, Arie and Wesseling, Pieter and Brat, Daniel J and Cree, Ian A and Figarella-Branger, Dominique and Hawkins, Cynthia and Ng, H K and Pfister, Stefan M and Reifenberger, Guido and Soffietti, Riccardo and von Deimling, Andreas and Ellison, David W},
    title = {The 2021 WHO Classification of Tumors of the Central Nervous System: a summary},
    journal = {Neuro-Oncology},
    volume = {23},
    number = {8},
    pages = {1231-1251},
    year = {2021},
    month = {08},
    issn = {1522-8517},
    doi = {10.1093/neuonc/noab106},
    url = {https://doi.org/10.1093/neuonc/noab106}
}

@misc{mozaffari2020review,
      title={A Review of 1D Convolutional Neural Networks toward Unknown Substance Identification in Portable Raman Spectrometer}, 
      author={M. Hamed Mozaffari and Li-Lin Tay},
      year={2020},
      eprint={2006.10575},
      archivePrefix={arXiv},
      primaryClass={eess.SP},
      url={https://arxiv.org/abs/2006.10575}, 
}

@article{caruana1997multitask,
  author    = {Caruana, Rich},
  title     = {Multitask Learning},
  journal   = {Machine Learning},
  volume    = {28},
  number    = {1},
  pages     = {41--75},
  year      = {1997},
  publisher = {Springer},
  doi       = {10.1023/A:1007379606734}
}

@article{gadermayr2024multiple,
  title={Multiple instance learning for digital pathology: A review of the state-of-the-art, limitations \& future potential},
  author={Gadermayr, Michael and Tschuchnig, Maximilian},
  journal={Computerized Medical Imaging and Graphics},
  volume={112},
  pages={102337},
  year={2024},
  publisher={Elsevier}
}

@article{andrews2002misvm,
author = {Andrews, Stuart and Tsochantaridis, Ioannis and Hofmann, Thomas},
year = {2002},
month = {01},
pages = {561-568},
title = {Support Vector Machines for Multiple-Instance Learning},
volume = {15},
journal = {Advances in Neural Information Processing Systems}
}

@article{guo2021chemometric,
  author  = {Guo, Shuxia and Popp, J{\"u}rgen and Bocklitz, Thomas},
  title   = {Chemometric analysis in Raman spectroscopy from experimental design to machine learning-based modeling},
  journal = {Nature Protocols},
  volume  = {16},
  pages   = {5426--5459},
  year    = {2021},
  doi     = {10.1038/s41596-021-00620-3}
}

@inproceedings{zheng2023revisiting,
  title={Revisiting discriminative vs. generative classifiers: Theory and implications},
  author={Zheng, Chenyu and Wu, Guoqiang and Bao, Fan and Cao, Yue and Li, Chongxuan and Zhu, Jun},
  booktitle={International conference on machine learning},
  pages={42420--42477},
  year={2023},
  organization={PMLR}
}

@article{ng2001discriminative,
  title={On discriminative vs. generative classifiers: A comparison of logistic regression and naive bayes},
  author={Ng, Andrew and Jordan, Michael},
  journal={Advances in neural information processing systems},
  volume={14},
  year={2001}
}

@inproceedings{li2023your,
  title={Your diffusion model is secretly a zero-shot classifier},
  author={Li, Alexander C and Prabhudesai, Mihir and Duggal, Shivam and Brown, Ellis and Pathak, Deepak},
  booktitle={2023 IEEE/CVF International Conference on Computer Vision (ICCV)},
  pages={2206--2217},
  year={2023},
  organization={IEEE}
}

@article{cancer2015comprehensive,
  title={Comprehensive, integrative genomic analysis of diffuse lower-grade gliomas},
  author={Cancer Genome Atlas Research Network},
  journal={New England Journal of Medicine},
  volume={372},
  number={26},
  pages={2481--2498},
  year={2015},
  publisher={Mass Medical Soc}
}

@article{eckel2015glioma,
  title={Glioma groups based on 1p/19q, IDH, and TERT promoter mutations in tumors},
  author={Eckel-Passow, Jeanette E and Lachance, Daniel H and Molinaro, Annette M and Walsh, Kyle M and Decker, Paul A and Sicotte, Hugues and Pekmezci, Melike and Rice, Terri and Kosel, Matt L and Smirnov, Ivan V and others},
  journal={New England Journal of Medicine},
  volume={372},
  number={26},
  pages={2499--2508},
  year={2015},
  publisher={Mass Medical Soc}
}

@article{waseem2025review,
  title={Review of generative AI for synthetic data generation: a healthcare perspective},
  author={Waseem, Hafiz Muhammad and Islam, Saif Ul and Matragkas, Nikolaos and Epiphaniou, Gregory and Arvanitis, Theodoros N and Maple, Carsten},
  journal={Artificial Intelligence Review},
  volume={59},
  number={2},
  pages={55},
  year={2025},
  publisher={Springer}
}

@article{zhang2010baseline,
  title={Baseline correction using adaptive iteratively reweighted penalized least squares},
  author={Zhang, Zhi-Min and Chen, Shan and Liang, Yi-Zeng},
  journal={Analyst},
  volume={135},
  number={5},
  pages={1138--1146},
  year={2010},
  publisher={The Royal Society of Chemistry}
}

@article{moran1950,
  title = {NOTES ON CONTINUOUS STOCHASTIC PHENOMENA},
  volume = {37},
  ISSN = {1464-3510},
  url = {http://dx.doi.org/10.1093/biomet/37.1-2.17},
  DOI = {10.1093/biomet/37.1-2.17},
  number = {1-2},
  journal = {Biometrika},
  publisher = {Oxford University Press (OUP)},
  author = {MORAN,  P. A. P.},
  year = {1950},
  pages = {17–23}
}

@article{rittel2026cohort,
author = {Rittel, Miriam F. and Ebert, Nikolas and Abu Sammour, Denis and Graf, Sebastian and Fröhlich, Björn C. and Birgin, Emrullah and Mohammed, Shad A. and Rahbari, Nuh N. and Wellmann, Axel and Wasenmüller, Oliver and Weis, Cleo-Aron and Schmidt, Stefan and Hopf, Carsten},
title = {Cohort-Scale Spatial Autocorrelation for Tumor Prediction in Mid-Infrared Pathology and Spatial Biomarker Discovery Using MALDI Imaging Lipidomics},
journal = {Advanced Science},
volume = {13},
number = {24},
pages = {e16847},
doi = {https://doi.org/10.1002/advs.202516847},
url = {https://advanced.onlinelibrary.wiley.com/doi/abs/10.1002/advs.202516847},
eprint = {https://advanced.onlinelibrary.wiley.com/doi/pdf/10.1002/advs.202516847},
year = {2026}
}

@article{veronesi2025machine,
  title={Machine learning to detect melanoma exploiting nuclei morphology and Spatial organization},
  author={Veronesi, Giulia and Curti, Nico and Gardini, Aldo and Querzoli, Giulia and Castellani, Gastone and Dika, Emi},
  journal={Scientific Reports},
  volume={15},
  number={1},
  pages={21594},
  year={2025},
  publisher={Nature Publishing Group UK, London}
}

@article{zhang2017mixup,
  title={mixup: Beyond empirical risk minimization},
  author={Zhang, Hongyi and Cisse, Moustapha and Dauphin, Yann N and Lopez-Paz, David},
  journal={arXiv preprint arXiv:1710.09412},
  year={2017}
}

@article{sciortino2021raman,
author = {Sciortino, Tommaso and Secoli, Riccardo and d’Amico, Ester and Moccia, Sara and Conti Nibali, Marco and Gay, Lorenzo and Rossi, Marco and Pecco, Nicolò and Castellano, Antonella and De Momi, Elena and Fernandes, Bethania and Riva, Marco and Bello, Lorenzo},
title = {Raman Spectroscopy and Machine Learning for IDH Genotyping of Unprocessed Glioma Biopsies},
journal = {Cancers},
volume = {13},
year = {2021},
number = {16},
doi = {10.3390/cancers13164196}
}

@article{yang2023histological,
  author  = {Yang, Yifan and Liu, Zhijie and Huang, Jing and Sun, Xiangjie and Ao, Jianpeng and Zheng, Bin and Chen, Wanyuan and Shao, Zhiming and Hu, Hao and Yang, Yinlong and Ji, Minbiao},
  title   = {Histological diagnosis of unprocessed breast core-needle biopsy via stimulated Raman scattering microscopy and multi-instance learning},
  journal = {Theranostics},
  year    = {2023},
  volume  = {13},
  number  = {4},
  pages   = {1342--1354},
  doi     = {10.7150/thno.81784}
}
